\documentclass[journal]{IEEEtran}
\usepackage{amsmath,amsfonts}
\usepackage{algorithm}
\usepackage{algorithmic}
\usepackage{array}
\usepackage{subfigure}
\usepackage{bbm}
\usepackage{textcomp}
\usepackage{stfloats}
\usepackage{url}
\usepackage{verbatim}
\usepackage{graphicx}
\usepackage{cite}
\usepackage{bm}
\usepackage{amsfonts,amssymb,amsmath}
\usepackage{amsthm}
\usepackage{pifont}

\usepackage{color}
\usepackage{listings}
\usepackage{booktabs}
\usepackage{booktabs}
\usepackage{multirow}
\usepackage{makecell}
\usepackage{colortbl}  %彩色表格需要加载的宏包
\usepackage{xcolor}
\definecolor{dkgreen}{rgb}{0,0.6,0}
\definecolor{gray}{rgb}{0.5,0.5,0.5}
\definecolor{mauve}{rgb}{0.58,0,0.82}
\definecolor{lemon}{RGB}{254,255,174}
\definecolor{lemongreen}{RGB}{223,247,155}
\definecolor{lightblue}{RGB}{223,247,255}
\definecolor{lightgreen}{RGB}{241,255,200}
\definecolor{lightgray}{RGB}{227,227,227}
\definecolor{lightlightgray}{RGB}{240,240,240}
\definecolor{lightorange}{RGB}{255,237,224}
\definecolor{lightpink}{RGB}{255,235,246}
\definecolor{midgray}{RGB}{0,0,0}
\definecolor{midgray2}{RGB}{0,128,0}
\definecolor{red}{RGB}{220,20,60}
\definecolor{newblue}{RGB}{0,0,205}

\usepackage{array}   %对表列和表格线的设置需要用到array宏包

\usepackage[
pdfauthor={derajan},
pdftitle={How to do this},
pdfstartview=XYZ,
bookmarks=true,
colorlinks=true,
linkcolor=blue,
urlcolor=blue,
citecolor=blue,
pdftex,
bookmarks=true,
linktocpage=true,   % makes the page number as hyperlink in table of content
hyperindex=true
]{hyperref}

\begin{document}

\title{A Dual-Space Framework for General Knowledge Distillation of Large Language Models}

\author{Xue Zhang\textsuperscript{*}\thanks{* This work was done during internship at Pattern Recognition Center, WeChat AI, Tencent Inc, China.}, Songming Zhang$^\dag$\thanks{$\dag$ {Songming Zhang is the corresponding author.}}, Yunlong Liang, Fandong Meng, Yufeng Chen, Jinan Xu and Jie Zhou
\thanks{Xue Zhang, Songming Zhang, Yufeng Chen, and Jinan Xu are with the School of Computer Science and Technology, Beijing Key Laboratory of Traffic Data Mining and Embodied Intelligence, Beijing Jiaotong University, Beijing 100044, China (e-mail: \{\text{zhang\_xue}; smzhang22; chenyf; jaxu\}@bjtu.edu.cn).}
\thanks{Yunlong Liang, Fandong Meng, and Jie Zhou are with Pattern Recognition Center, WeChat AI, Tencent Inc, China (e-mail: \{yunlonliang; fandongmeng; withtomzhou\}@tencent.com).}
\thanks{This paper is an extended version of our prior work published at the conference EMNLP-2024 (\url{https://aclanthology.org/2024.emnlp-main.1010/}). The key differences are listed in Appendix \ref{sec:appendix-difference}. Our code is publicly available at \url{https://github.com/songmzhang/DSKDv2}.}
}

% \markboth{Journal of \LaTeX\ Class Files,~Vol.~x, No.~x, xxx~2025}%

% \IEEEpubid{0000--0000/00\$00.00~\copyright~2025 IEEE}

\maketitle

\begin{abstract}
Knowledge distillation (KD) is a promising solution to compress large language models (LLMs) by transferring their knowledge to smaller models.
During this process, white-box KD methods usually minimize the distance between the output distributions of the teacher model and the student model to transfer more information.
However, we reveal that the current white-box KD framework exhibits two limitations: a) bridging probability distributions from different output spaces will limit the similarity between the teacher model and the student model; b) this framework cannot be applied to LLMs with different vocabularies.
One of the root causes for these limitations is that the distributions from the teacher and the student for KD are output by different prediction heads, which yield distributions in different output spaces and dimensions.
Therefore, in this paper, we propose a dual-space knowledge distillation (DSKD) framework that unifies the prediction heads of the teacher and the student models for KD.
Specifically, we first introduce two projectors with ideal initialization to project the teacher/student hidden states into the student/teacher representation spaces.
After this, the hidden states from different models can share the same head and unify the output spaces of the distributions.
Furthermore, we develop an exact token alignment (ETA) algorithm to align the same tokens in two differently-tokenized sequences.
Based on the above, our DSKD framework is a general KD framework that supports both off-policy and on-policy KD, and KD between any two LLMs regardless of their vocabularies.
Extensive experiments on instruction-following, mathematical reasoning, and code generation benchmarks show that DSKD significantly outperforms existing methods based on the current white-box KD framework and surpasses other cross-tokenizer KD methods for LLMs with different vocabularies.

\end{abstract}


\section{Introduction}
% 如果篇幅不够可以在第一页右上角加一个图：DSKD和普通white-box-KD支持的蒸馏类型的区别
\IEEEPARstart{E}{xisting} large language models (LLMs) have exhibited strong generalization abilities on various tasks due to their huge model capacities \cite{chowdhery23palm,touvron23llama,openai23gpt4}. 
With faith in the scaling law \cite{kaplan20scaling}, the number of parameters in current LLMs is expanded steadily to achieve higher intelligence.
However, the increasing parameters also bring high deployment costs in real scenarios.
For this problem, knowledge distillation (KD) \cite{hinton15kd} is one of the promising solutions to compress large models with acceptable performance sacrifice.
During the process of KD, the large model typically serves as the teacher and provides supervision signals for a small model (known as the student), and thus the knowledge and the abilities of the teacher can be transferred to the lightweight student. 

Currently, KD algorithms for LLMs are usually under two frameworks, \emph{i.e.}, black-box KD and white-box KD.
Black-box KD uses the teacher's decoding sequences as the training data of the student and directly optimizes the cross-entropy loss on the one-hot target \cite{kim16seqkd,fu23cotkd,li23symbolickd}.
By contrast, white-box KD methods generally minimize the distance (\emph{e.g.}, KL divergence) between the output distributions of the teacher and the student, which theoretically transfer more information and usually perform better than black-box KD \cite{wen23fdiv,gu23minillm,ko24distillm}.
Although the framework of white-box KD has shown its superiority, the distributions of the student and the teacher in this framework are from different output spaces since they are produced by different prediction heads.
At the beginning of this work, we first reveal two inherent limitations of this white-box KD framework due to the discrepancy of prediction heads:
\begin{itemize}
    \item \textbf{Low Teacher-Student Similarity (Section \ref{sec:low_sim}):} The current framework usually yields limited similarity between the teacher and the student on both representation and distribution levels, which limits the distillation performance.
    % 如果篇幅不够这里可以再多解释一句representation 和 distribution 是什么，低相似度导致的结果是什么？
    \item \textbf{Dependency on the Same Vocabulary (Section \ref{sec:depend_same_vocab}):} A key requirement for current white-box KD is that the two LLMs should share the same vocabulary, which, however, is rarely satisfied for various LLMs in this era.
    This largely restricts the practical application of white-box KD for current LLMs.
    % 这里还需要再加例子以进一步说明吗？
    % 可以再加上导致的结果是什么？即只能在相同词表的模型下做蒸馏，非常受限；
\end{itemize}

\IEEEpubidadjcol

Aiming at these limitations, we propose a new framework for white-box KD, named dual-space knowledge distillation (DSKD), which works similarly to the current white-box KD framework but is more general.
Specifically, DSKD unifies the output spaces of the two models by projecting the output hidden states\footnote{In this paper, ``output hidden states'' means the hidden states output by the last layer of the model.} of the teacher/student to the representation spaces of the student/teacher with the ideal initialized projectors.
Then, we can use the shared prediction heads to produce the teacher and student distributions in the same output spaces and conduct the distillation process on the teacher and student spaces, respectively.
Like the current framework, DSKD is also compatible with existing divergences for distribution matching, including KL divergence, reverse KL divergence, and so on.
% which enhances the similarity of distributions of the teacher and student.
% In particular, for models with different vocabularies, we further develop a cross-model attention (CMA) mechanism to automatically align the tokens in two differently tokenized sequences.
In particular, for models with different vocabularies, we further develop an exact token alignment algorithm (ETA) for DSKD to align the same tokens in two differently tokenized sequences.
% for adapting our DSKD to the different-vocabulary distillation scenario. 
By sharing the prediction heads and aligning tokens with the ETA algorithm, the distributions of the two LLMs for divergence calculation are exactly in the same shape, which makes DSKD more general and can be applied to any two LLMs regardless of their vocabularies.
Furthermore, our DSKD can also be extended to the on-policy KD scenario to mitigate the training-inference mismatch problem in off-policy KD.
% Moreover, as our framework only focuses on unifying the representation space of the two models and has no dependency on certain objectives, it is naturally compatible with all the current objectives for token-level KD, such as KL divergence, reverse KL divergence and the more advanced ones \cite{ko24distillm,wu2024rethinking}.

We first evaluate our framework on instruction-following benchmarks under both settings that the two LLMs have the same/different vocabularies. 
Experimental results show that for LLMs with the same vocabulary, our DSKD framework significantly outperforms the current white-box KD framework on various divergences, showcasing the effectiveness of unifying the output space for KD.
Moreover, DSKD surpasses all existing cross-tokenizer KD methods for LLMs with different vocabularies.
Surprisingly, the performance of DSKD on the cross-tokenizer distillation scenario is comparable and even better than DSKD on the same-vocabulary scenario, indicating the potential and possibility of using stronger teacher models with different vocabularies to guide better students under our DSKD framework.
Additionally, we demonstrate that the performance of DSKD can be further improved in the on-policy KD scenario and exceeds existing on-policy KD methods.
Furthermore, we also evaluate DSKD on other tasks, such as LLM alignment, math, and code generation, all of which demonstrate the effectiveness of our framework.

To sum up, the contributions of this paper are as follows:
\begin{itemize}
    \item We empirically reveal that the current white-box KD framework limits the lower similarity between the student and the teacher and relies on a shared vocabulary between the two models. 
    \item As a solution, we propose a new framework for white-box KD, named dual-space knowledge distillation (DSKD), which unifies the output spaces of the distributions from the teacher and the student via sharing prediction heads.
    \item On this basis, we further develop an exact token alignment (ETA) algorithm to support KD between LLMs with different vocabularies, which renders DSKD a more general white-KD framework that can be applied to any two LLMs regardless of their vocabularies.
    \item Extensive experiments on instruction-following, math, and code generation show that DSKD significantly outperforms existing methods under the current white-box KD framework under off/on-policy KD and cross-tokenizer KD, demonstrating the effectiveness and high versatility of our framework. 
\end{itemize}

\begin{figure*}
	\centering
	\subfigure[KL: Before KD]{
		\begin{minipage}[t]{0.23\linewidth}
			\centering
			\includegraphics[width=\linewidth]{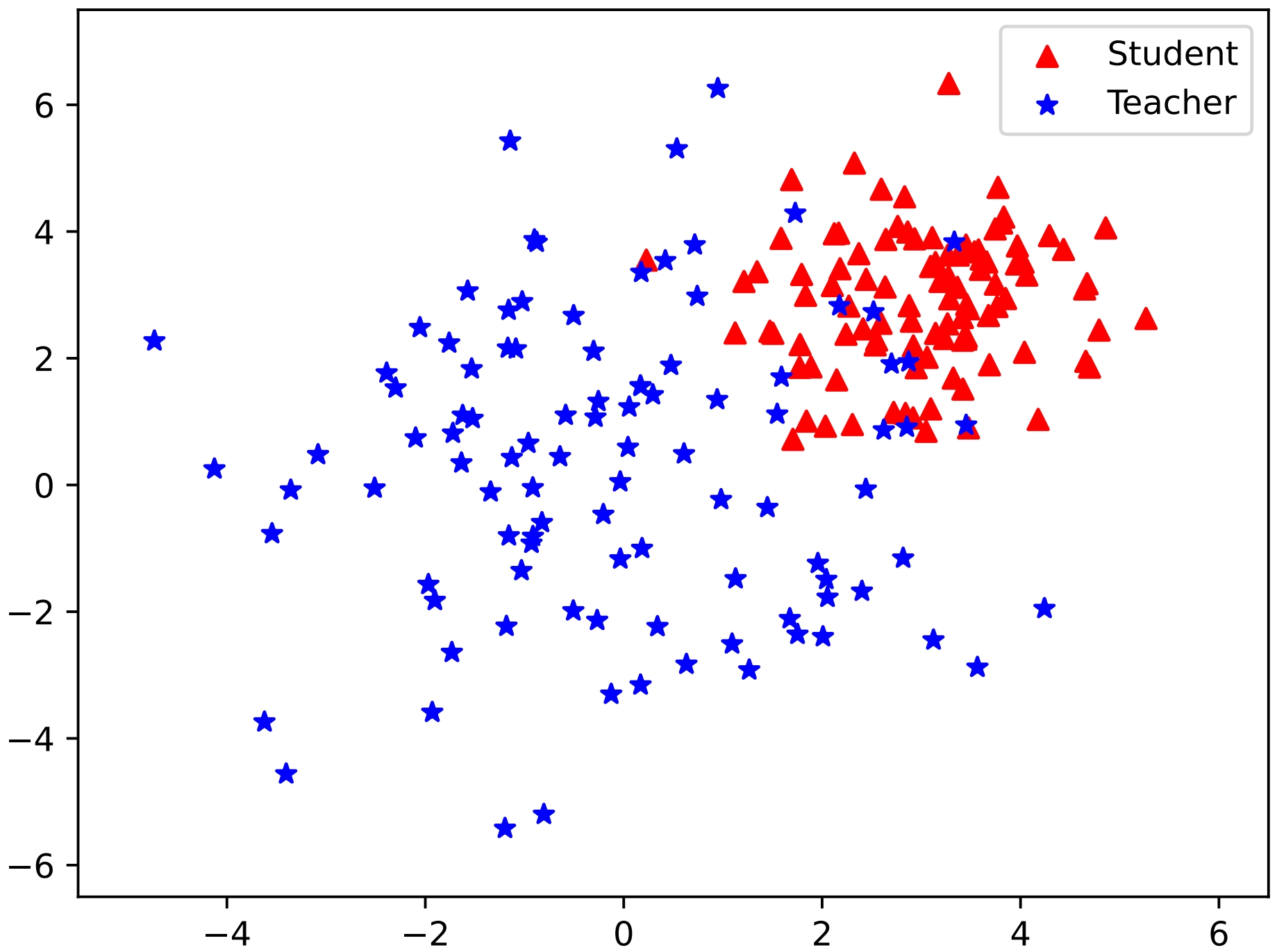}
		\end{minipage}
	}%
	\subfigure[KL: After KD (Different heads)]{
		\begin{minipage}[t]{0.23\linewidth}
			\centering
			\includegraphics[width=\linewidth]{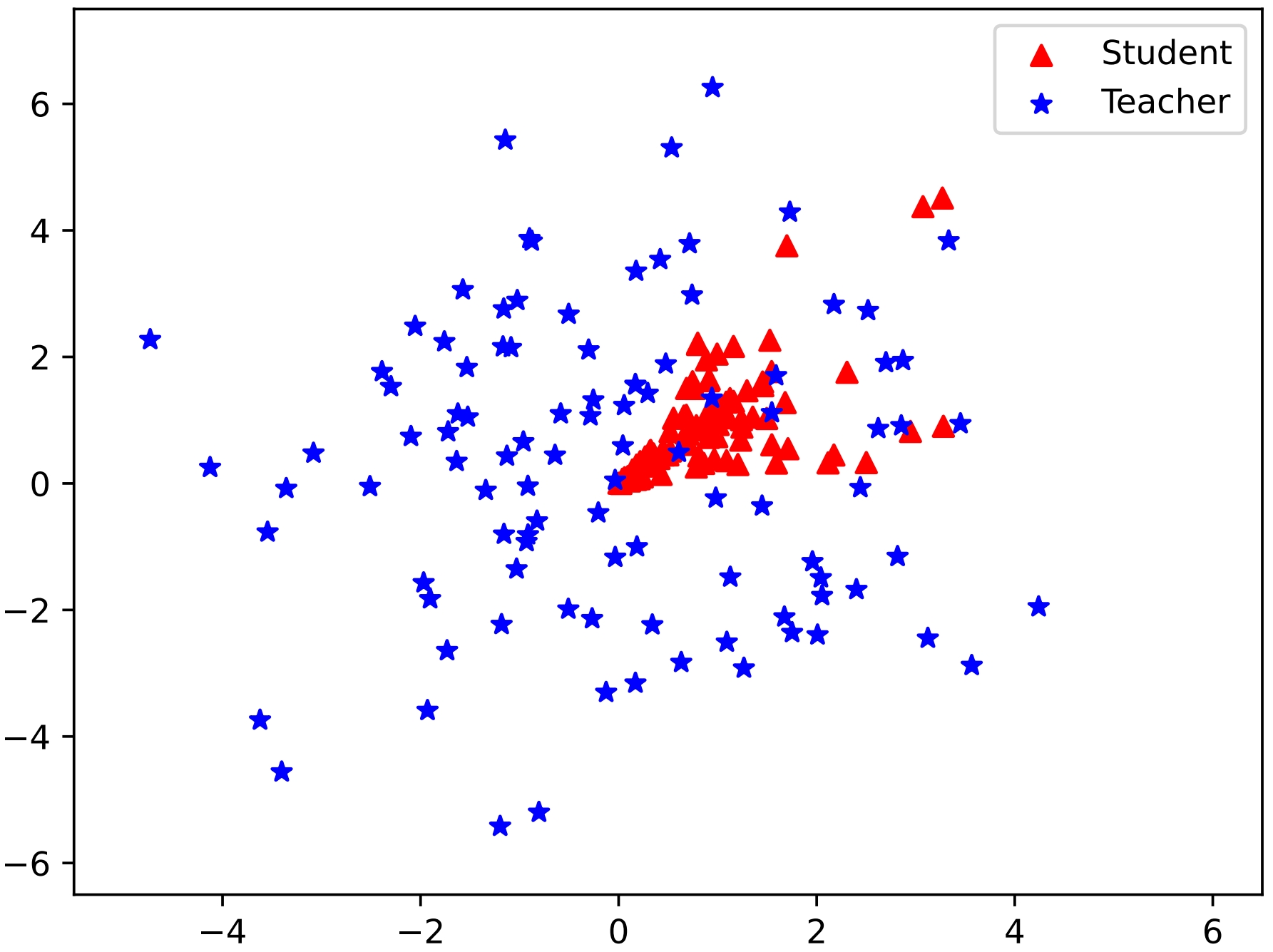}
		\end{minipage}
	}% 
	\subfigure[KL: After KD (Shared head)]{
		\begin{minipage}[t]{0.23\linewidth}
			\centering
			\includegraphics[width=\linewidth]{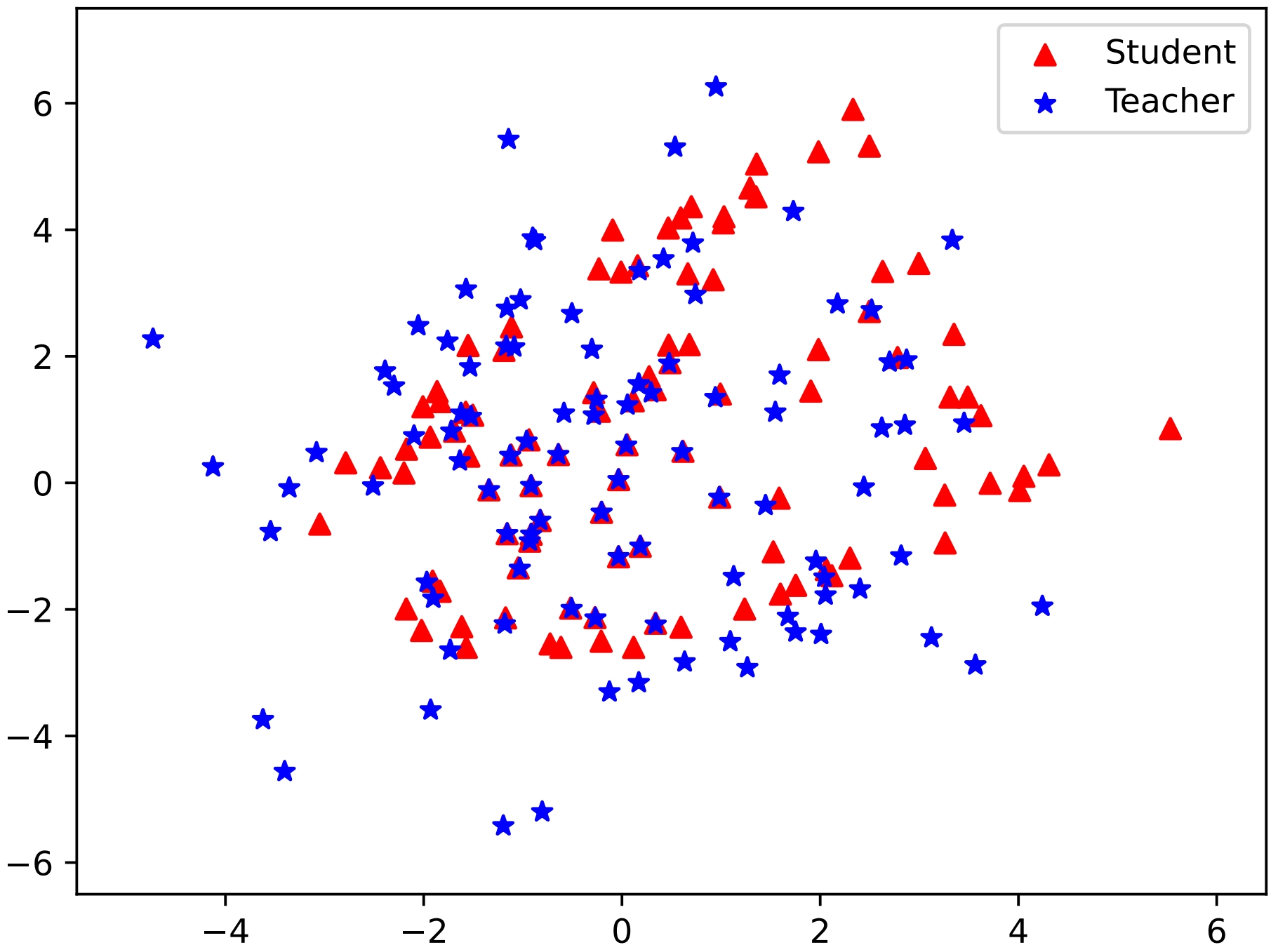}
		\end{minipage}
	}%
	\subfigure[KL: Loss curves of KD]{
		\begin{minipage}[t]{0.23\linewidth}
			\centering
			\includegraphics[width=\linewidth]{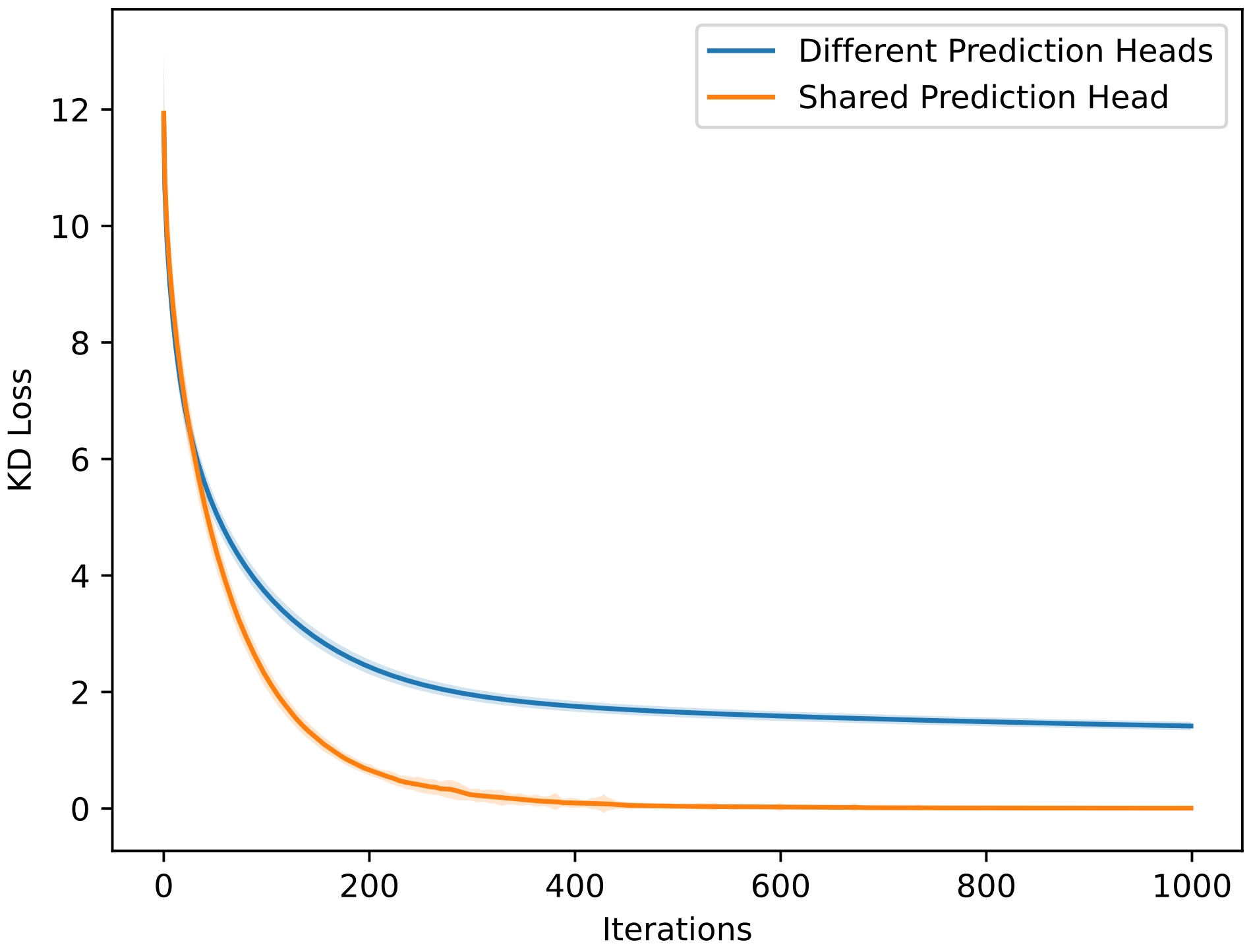}
		\end{minipage}
	}%

        \subfigure[RKL: Before KD]{
		\begin{minipage}[t]{0.23\linewidth}
			\centering
			\includegraphics[width=\linewidth]{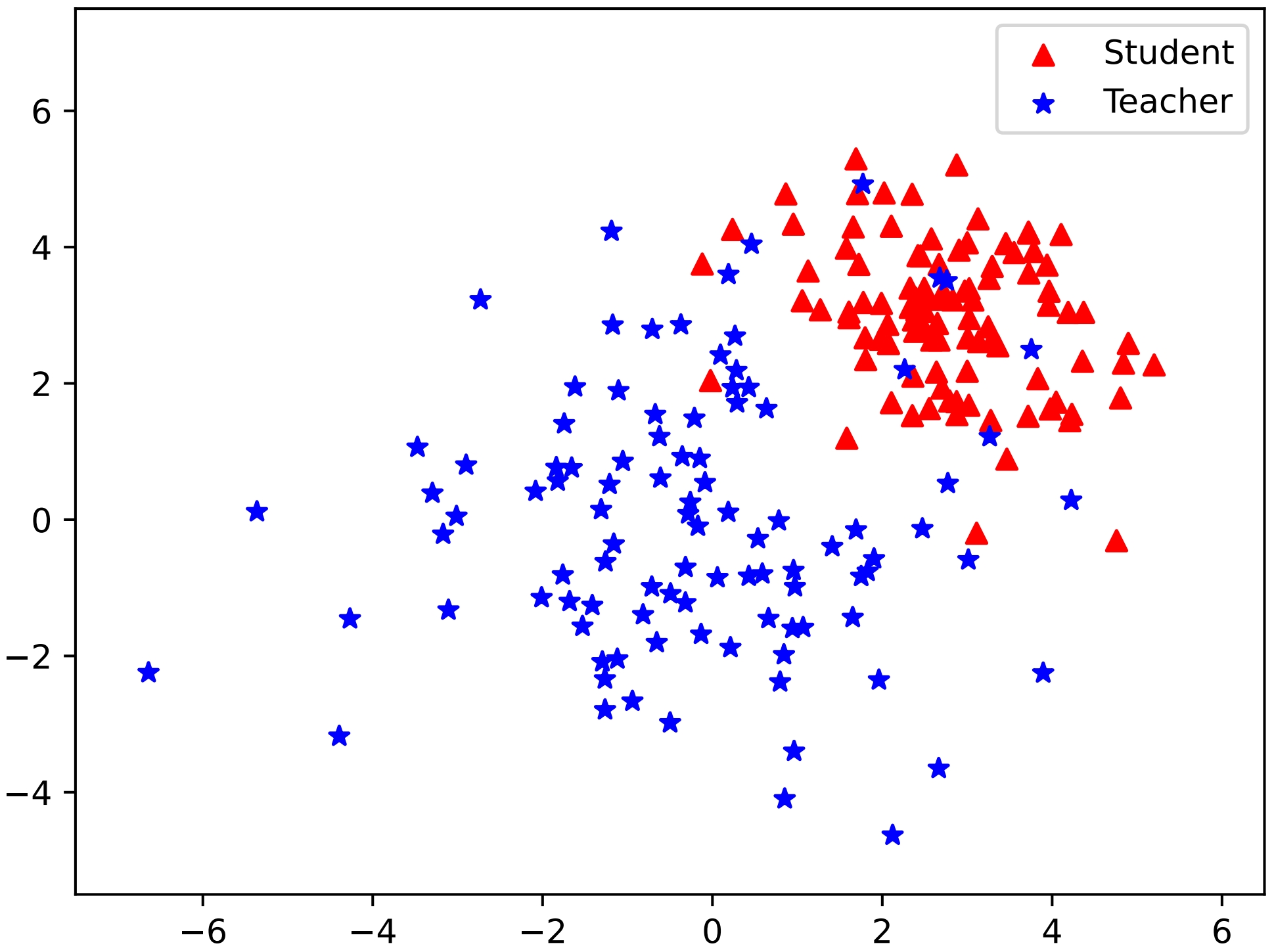}
		\end{minipage}
	}%
	\subfigure[RKL: After KD (Different heads)]{
		\begin{minipage}[t]{0.23\linewidth}
			\centering
			\includegraphics[width=\linewidth]{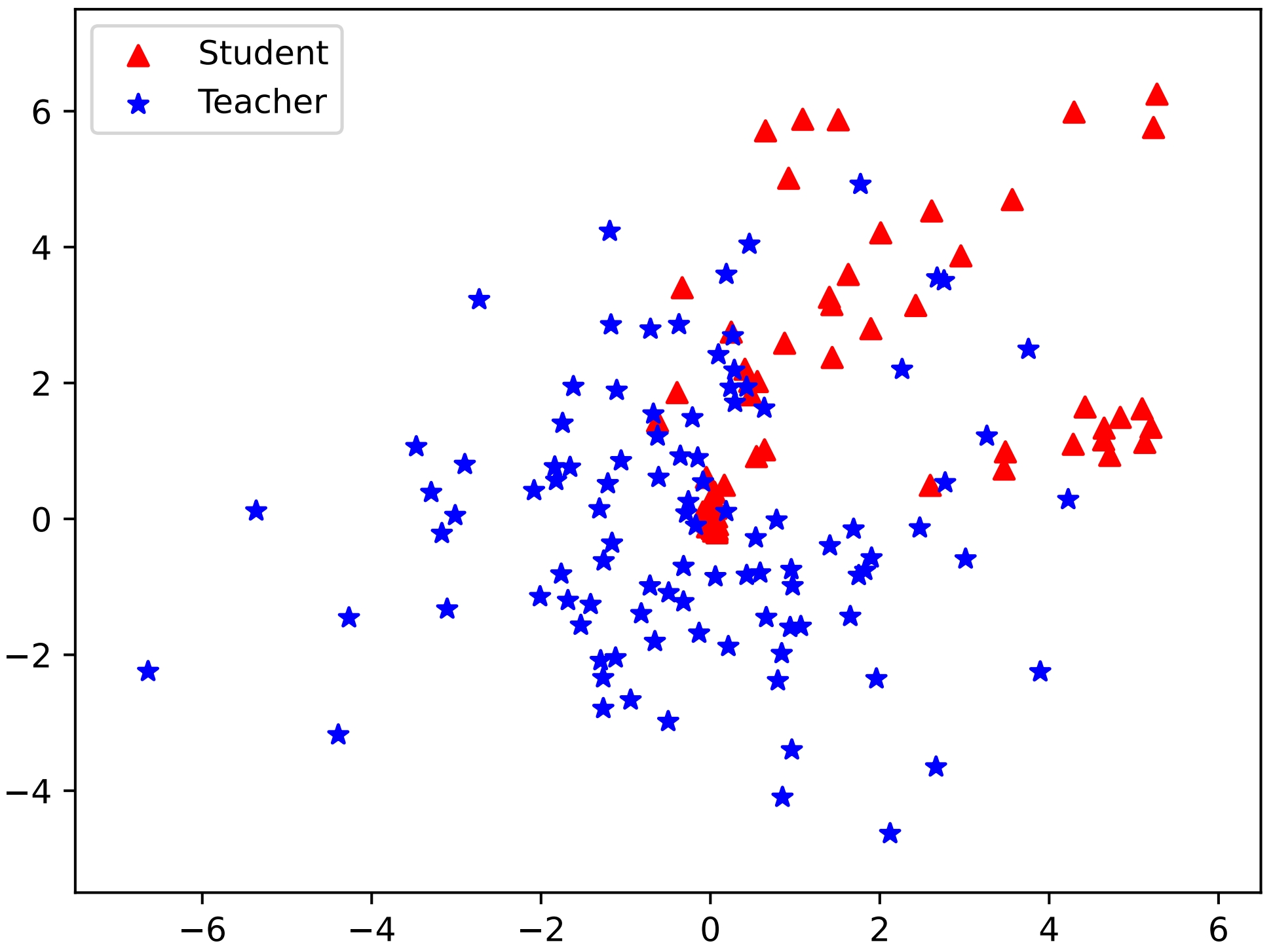}
		\end{minipage}
	}% 
	\subfigure[RKL: After KD (Shared head)]{
		\begin{minipage}[t]{0.23\linewidth}
			\centering
			\includegraphics[width=\linewidth]{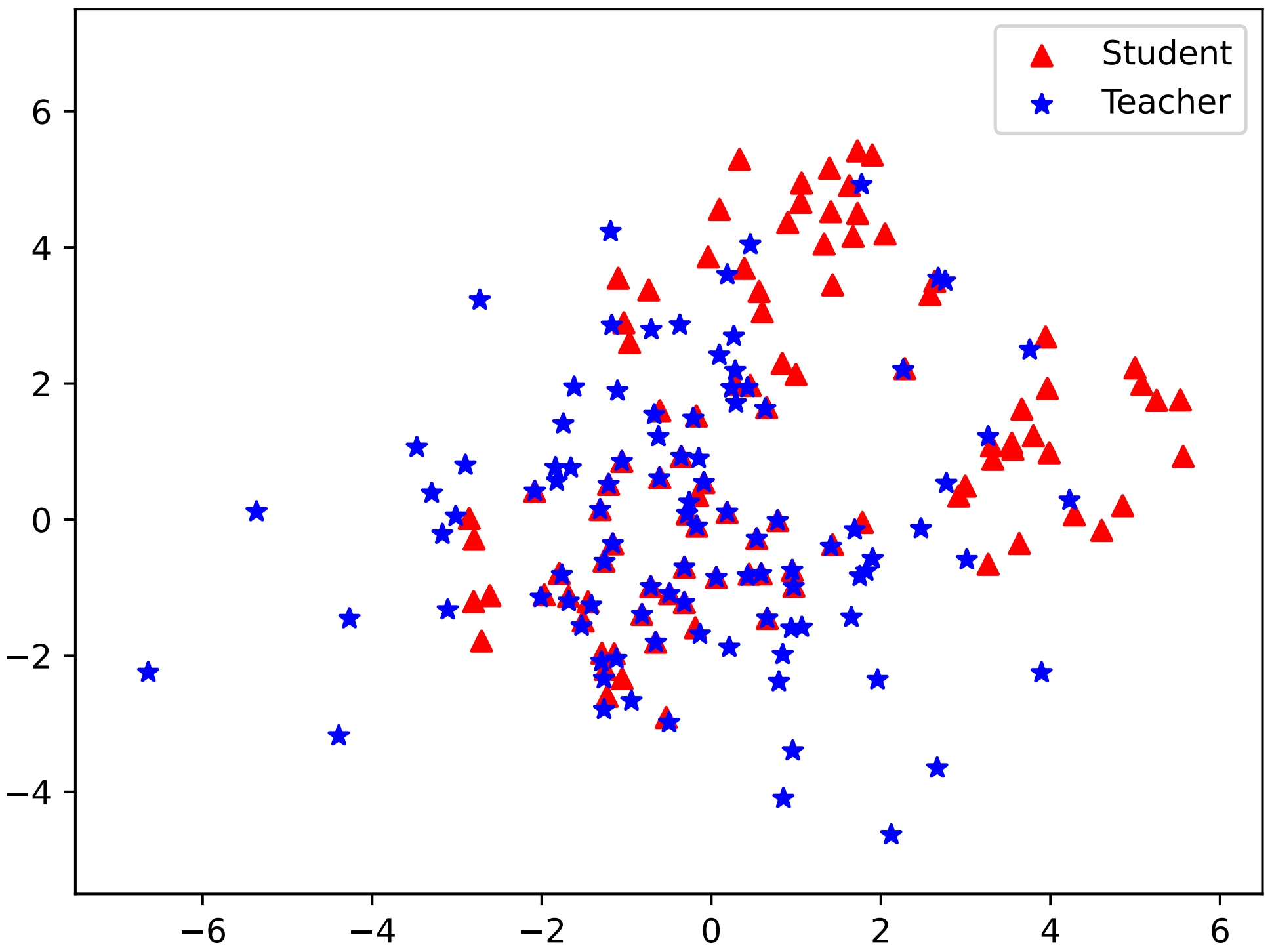}
		\end{minipage}
	}%
	\subfigure[RKL: Loss curves of KD]{
		\begin{minipage}[t]{0.23\linewidth}
			\centering
			\includegraphics[width=\linewidth]{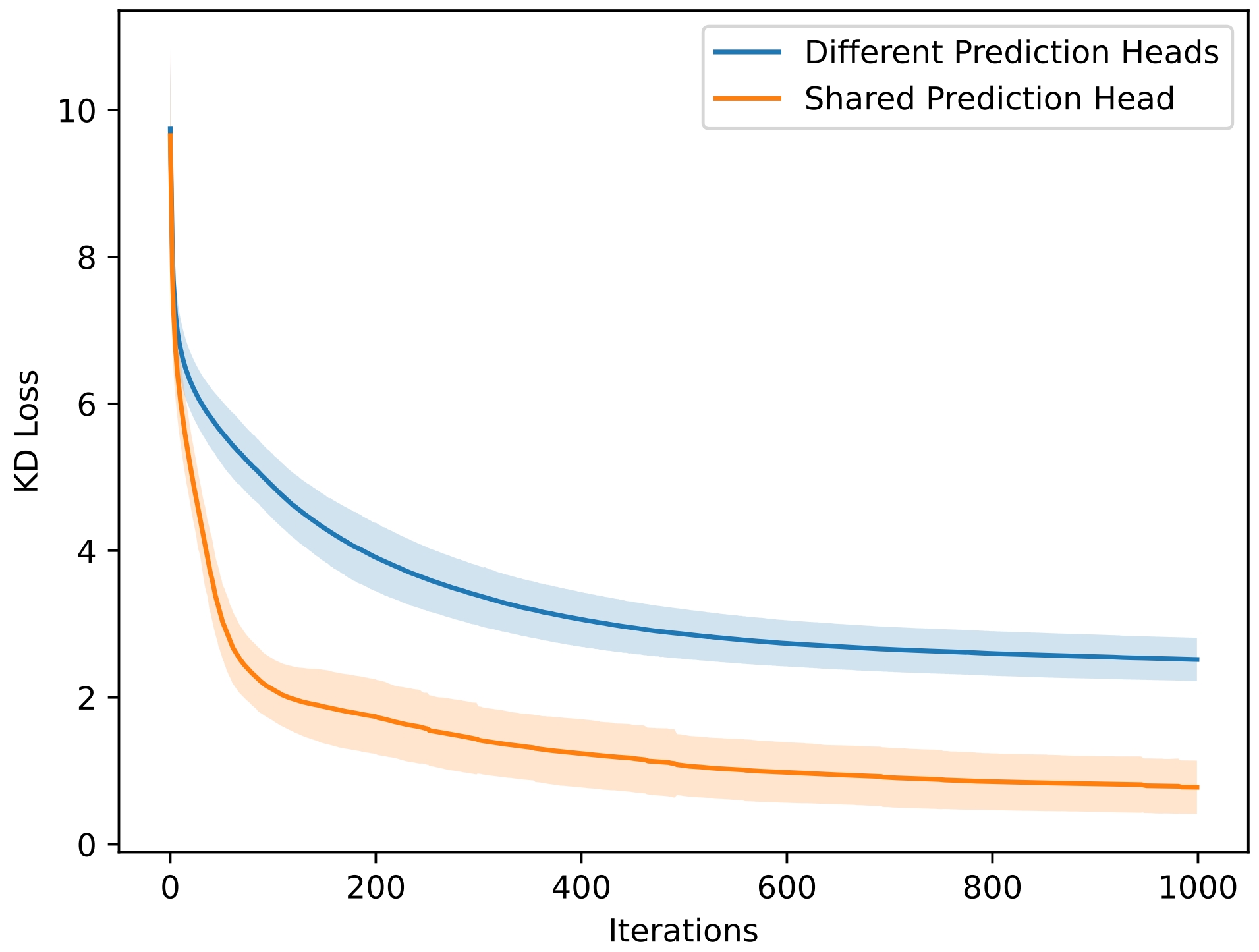}
		\end{minipage}
	}%
    
	\centering
	\caption{Simulation results with KL and RKL divergence as the divergence $\mathcal{D}(\cdot||\cdot)$. (a), (b), (c), (e), (f), and (g) plot the {\color{red} student's hidden states} and the {\color{blue} teacher's hidden states} before and after the two KD processes. ``Different heads'' means using the teacher and student head respectively during KD, while ``Shared head'' means only using the student head as the shared head to obtain the distributions during KD. (d) and (h) show the convergence curves of $\mathcal{L}_{kd}$ in the two KD processes.}
	\label{fig:kl_simulation}
\end{figure*}

\section{Background and Preliminary Study}
\subsection{Current Framework for White-Box KD} \label{sec:background}
Given a prompt $\mathbf{x}$ and a corresponding response $\mathbf{y}$, current LLMs generally learn the causal language modeling objective at each token position $i$ of $\mathbf{y}$ via the cross-entropy loss:
\begin{equation} \label{eq:ce_loss}
    \mathcal{L}_{ce} = - \frac{1}{|\mathbf{y}|} \sum_{i=1}^{|\mathbf{y}|} \log q_{\theta}(y_i=y^{*}_i|\mathbf{y}_{<i},\mathbf{x}), 
\end{equation}
where $q_{\theta}(y_i=y^{*}_i|\mathbf{y}_{<i},\mathbf{x})$ denotes the probability of the student model on the target token $y^{*}_i$ conditioning on the response prefix $\mathbf{y}_{<i}$ and the prompt $\mathbf{x}$.
On this basis, the current white-box KD framework first feeds the prompt and the response into the teacher model to obtain its token-level probability distributions $p(y_i|\mathbf{y}_{<i},\mathbf{x})$.
Then, the following loss is minimized to push the student distribution $q_{\theta}(y_i|\mathbf{y}_{<i},\mathbf{x})$ to the teacher distribution $p(y_i|\mathbf{y}_{<i},\mathbf{x})$:
\begin{equation} \label{eq:kd_loss}
    \mathcal{L}_{kd}= \frac{1}{|\mathbf{y}|} \sum_{i=1}^{|\mathbf{y}|} \mathcal{D}(p(y_i|\mathbf{y}_{<i},\mathbf{x};\tau) || q_{\theta}(y_i|\mathbf{y}_{<i},\mathbf{x};\tau)),
\end{equation}
where $\mathcal{D}(\cdot || \cdot)$ is a divergence function that measures the distance between the two distributions (\emph{e.g.}, KL divergence) and $\tau$ is the temperature coefficient for KD \cite{hinton15kd} to control the sharpness of the distributions (we will omit $\tau$ in the following formulations for simplicity).
% The final loss is usually the interpolation of the two losses controlled by a hyper-parameter $\alpha$:
% \begin{equation} \label{eq:ce_kd_loss}
%     \mathcal{L} = (1 - \alpha) \mathcal{L}_{ce} + \alpha \mathcal{L}_{kd}.
% \end{equation}

On the choice of the divergence $\mathcal{D}(\cdot || \cdot)$ in Eq. (\ref{eq:kd_loss}), there have been several explorations (\emph{e.g.}, (reverse) KL divergence, skewed (reverse) KL divergence, adaptive KL divergence) in recent literature \cite{wen23fdiv,agarwal24gkd,ko24distillm,wu2024rethinking} that aim to improve the performance of KD for LLMs.
% Within this white-box KD framework, all these explorations focus on the formulations of the distance calculation given the two probability distributions like Eq. \ref{eq:kd_loss}.
However, in the following section, we will uncover that no matter which divergence is employed, the current white-box KD framework has two inherent limitations since the two distributions $p(y_i|\mathbf{y}_{<i},\mathbf{x};\tau)$ and $q_{\theta}(y_i|\mathbf{y}_{<i},\mathbf{x};\tau)$ are from different output spaces.

\subsection{Limitations of the Current Framework}
\subsubsection{Low Teacher-Student Similarity} \label{sec:low_sim}

In the current white-box KD framework, the two output distributions in Eq. (\ref{eq:kd_loss}) are calculated from different output spaces of two models using their respective prediction heads, \textit{i.e.}, different \texttt{lm\_head} modules.
Then, the student distribution will be optimized toward the teacher distribution by minimizing their distance.
However, we suspect that this practice will limit the final similarity between the student and the teacher from two aspects: \textbf{a) representation:} as the distributions are the results of the output hidden states through the prediction heads, if the prediction heads of the two models are different, even if the distributions are close after optimization, their hidden states are probably unsimilar; \textbf{b) distribution:} If the output hidden states of the student and the teacher are not similar, the practical distance between their distributions is difficult to reach its theoretical minimum during optimization.

We verify the above conjectures by a simulation experiment.
The pseudo-code of this experiment is present in Algorithm \ref{alg:alg2} of Appendix \ref{sec:pseudo_code}.
In this experiment, we randomly initialize two sets of 2-D vectors (one is trainable and the other is frozen) with different mean values and variances to represent the initial output hidden states of the student and the teacher, respectively (as plotted in Fig. \ref{fig:kl_simulation}(a) and \ref{fig:kl_simulation}(e)).
Besides, we set two prediction heads to produce probability distributions of the student and the teacher from these representation vectors.
Then, we select KL divergence and reverse KL (RKL) divergence as the divergence $\mathcal{D}(\cdot||\cdot)$ to simulate the KD process with $\mathcal{L}_{kd}$ in Eq. (\ref{eq:kd_loss}) for 1000 iterations.
After the iterations, we plot the two sets of vectors again and record the loss curve during the whole process in Fig. \ref{fig:kl_simulation}.

Firstly, we simulate the process of the current white-box KD framework, which uses distributions from different output spaces produced by different prediction heads.
The results in Fig. \ref{fig:kl_simulation}(b) and \ref{fig:kl_simulation}(f) show that the student's hidden states optimized by the current KD framework still exhibit distinct structure discrepancy from the teacher's hidden states after KD, reflecting low similarity on the representation.
As a comparison, we then unify the output spaces of the two distributions by sharing the same prediction head for the student and the teacher (\textit{i.e.}, using the student head as the shared head) and conduct the same KD process as above.
As shown in Fig. \ref{fig:kl_simulation}(c) and \ref{fig:kl_simulation}(g), under this setting, the student's hidden states become more similar and closer to the teacher's hidden states.
The significant difference between these two settings indicates that the current KD framework may lead to sub-optimal similarity between the student and the teacher \textbf{on the representation level}.
By contrast, a better alternative is to unify the output spaces for the distributions of the student and the teacher by sharing the prediction head.

Then, we repeat the simulations of the above two settings 100 times and plot their averaged curves of $\mathcal{L}_{kd}$ in Fig. \ref{fig:kl_simulation}(d) and \ref{fig:kl_simulation}(h).
As we suspected, when using different prediction heads, the value of KL divergence still leaves a large margin to its theoretical minimum (\emph{i.e.}, 0) after convergence.
On the contrary, when using a shared prediction head, the value of KL divergence will converge faster and finally be closer to this minimum.
It sufficiently illustrates that the current KD framework also limits the similarity between the two models \textbf{on the distribution level}.
% Besides KL divergence, we also conduct these simulations with other divergences, including Reverse KL divergence (RKL), JS divergence, skewed KL divergence (SKL), skewed RKL divergence (SRKL), and adaptive KL divergence (AKL).
Besides KL and RKL divergence, the corresponding results of skewed KL divergence (SKL), skewed RKL divergence (SRKL), and adaptive KL divergence (AKL) are shown in Fig. \ref{fig:srakl_simulation} of Appendix \ref{sec:other_simulation}, which present the same conclusion as KL and RKL.
% no matter which divergence is used, the student after KD will have low representation similarity with the teacher when using different prediction heads.
Thus, all these results suggest that the current white-KD framework may have inherent flaws in enhancing the similarity between the student model and the teacher model, which may limit the KD performance.
As a solution, unifying the output spaces by sharing the prediction head for the teacher and the student may achieve a more effective KD process.

\begin{figure}[t]
    \centering
    \includegraphics[width=\linewidth]{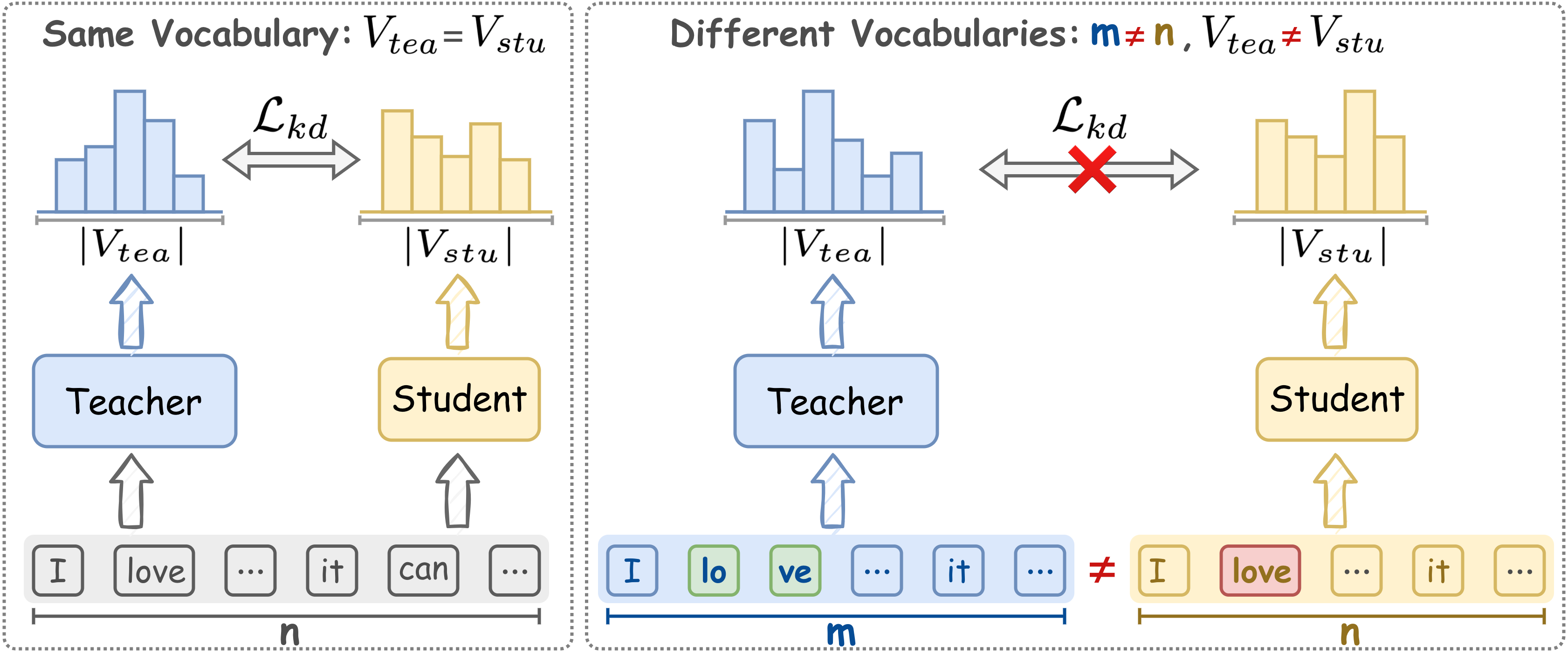}
    \caption{The difference between KD for LLMs with the same vocabulary and different vocabularies.}
    \label{fig:same-dif_pic}
\end{figure}

\subsubsection{Dependency on the Same Vocabulary} \label{sec:depend_same_vocab}
As stated in Section \ref{sec:background}, the current KD framework minimizes the distance between the two distributions at each token position.
However, when the teacher and the student have different vocabularies, the same text may be tokenized into different sequences like $\mathbf{y}^{t}=[y^t_1,y^t_2,...,y^t_m]$ and $\mathbf{y}^{s}=[y^s_1,y^s_2,...,y^s_n]$ as shown in the right part of Fig. \ref{fig:same-dif_pic}.
Under this circumstance, the teacher distribution $p(y^t_i|\mathbf{y}^{t}_{<i};\mathbf{x})$ is probably incorrect for $q_{\theta}(y^s_i|\mathbf{y}^{s}_{<i};\mathbf{x})$.
Additionally, as the output spaces are more different when the prediction heads contain different vocabularies, the produced distributions are even with different dimensions, which is obviously prohibited by Eq. (\ref{eq:kd_loss}).
Therefore, the current white-box KD framework fails to work between LLMs with different vocabularies.

\begin{figure*}[t]
    \centering
    \includegraphics[width=\linewidth]{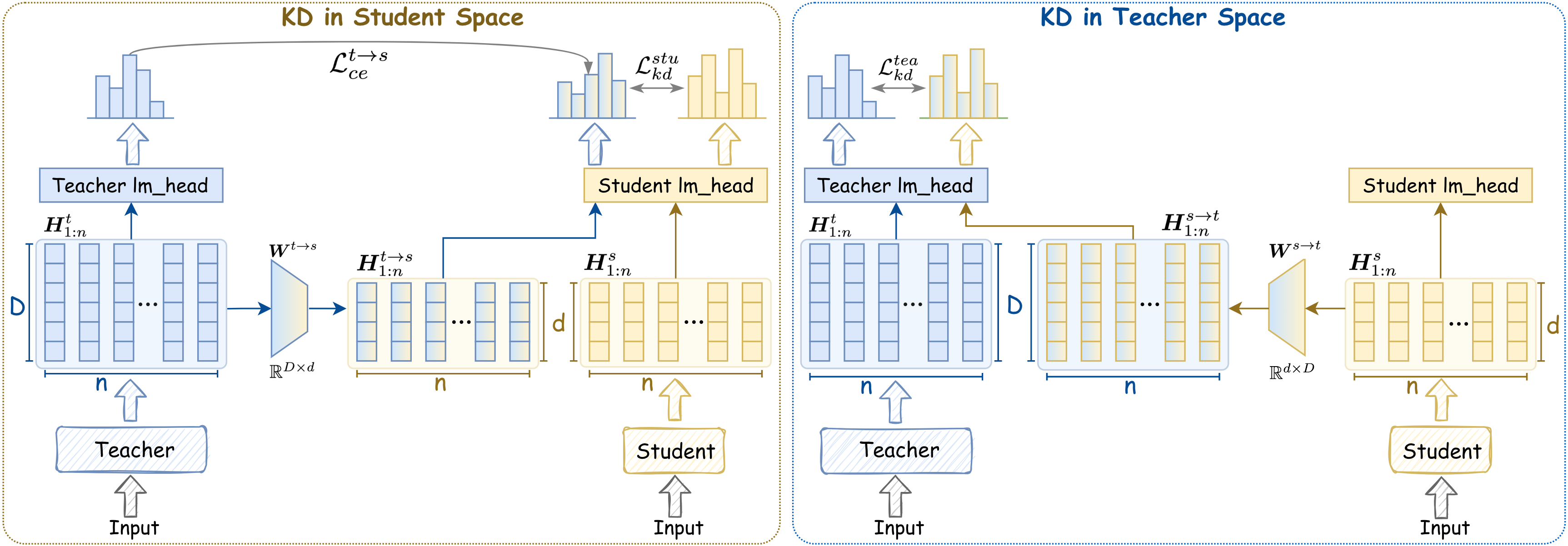}
    \caption{The framework of our DSKD. Our DSKD includes the KD in student space and teacher space. In student space, we use the projector $\bm{W}^{t \rightarrow s}$ to project the teacher hidden states $\bm{H}^{t}_{1:n}$ to the student space as $\bm{H}^{t \rightarrow s}_{1:n}$. Then, we feed $\bm{H}^{t \rightarrow s}_{1:n}$ and $\bm{H}^{s}_{1:n}$ to the student prediction head to obtain the distributions in the same space, which are used to calculate $\mathcal{L}^{stu}_{kd}$ and $\mathcal{L}^{t \rightarrow s}_{ce}$.
    In teacher space, we use the projector $\bm{W}^{s \rightarrow t}$ to project the student hidden states $\bm{H}^{s}_{1:n}$ to the teacher space as $\bm{H}^{s \rightarrow t}_{1:n}$.
    After the teacher prediction head, we calculate $\mathcal{L}^{tea}_{kd}$ with the two distributions in the teacher space.
    The overall loss of DSKD sums the three losses: $\mathcal{L}_{dskd}=\mathcal{L}^{stu}_{kd} + \mathcal{L}^{t \rightarrow s}_{ce} + \mathcal{L}^{tea}_{kd}$.
    }
    \label{fig:main_pic}
\end{figure*}

\section{Methodology}
This section introduces our solutions to the above limitations of the current white-box KD framework.
Firstly, we will introduce our new KD framework, \emph{i.e.}, dual-space knowledge distillation (DSKD) for LLMs with the same vocabulary in Section \ref{sec:method_framework}.
Then, we present the exact token alignment algorithm (ETA) in Section \ref{sec:method_cross_model}, which makes our DSKD support KD for LLMs with different vocabularies.
Finally, we extend our DSKD to the on-policy scenario in Section \ref{sec:on-policy}.

\subsection{Dual-Space Knowledge Distillation Framework} 
\label{sec:method_framework}
Inspired by the observations in Section \ref{sec:low_sim}, we design our dual-space knowledge distillation framework as shown in Fig. \ref{fig:main_pic}.
The core idea is to unify the output spaces of the teacher's and student's distributions in Eq. (\ref{eq:kd_loss}).
To achieve this, we project the output hidden states of the teacher/student model into the representation space of the student/teacher model, after which the subsequent distributions can be output by the same prediction head and thus lie in \textbf{the unified output space}.
Next, we will detail how to conduct the projection and unify KD in the student and teacher space.

\textbf{KD in Student Space:} In the student space, we first use a linear projector $\bm{W}^{t \rightarrow s}$ to transform the hidden states of the teacher model into the representation space of the student model.
Here, we denote the output hidden states of the whole sequence from the teacher model as $\bm{H}^{t}_{1:n}=\{h^t_1, h^t_2, ..., h^t_n\}$.
The projection process can be formulated as follows:
\begin{equation} \label{eq:down_transform}
    \bm{H}^{t \rightarrow s}_{1:n} = \bm{H}^{t}_{1:n}\bm{W}^{t \rightarrow s} \in \mathbb{R}^{n \times d},
\end{equation}
where $\bm{W}^{t \rightarrow s} \in \mathbb{R}^{D \times d}$ is learnable and $d, D$ are the hidden sizes of the student model and the teacher model, respectively.
We can obtain the transformed teacher distribution by feeding the projected hidden states $\bm{H}^{t \rightarrow s}_{1:n}$ into the student's prediction head and thus this new distribution will share the same output space with the student:
\begin{equation} \label{eq:p_t2s}
    p^{t \rightarrow s} = {\rm softmax}(\bm{H}^{t \rightarrow s}_{1:n}{\rm sg}(\bm{W}^s)) \in \mathbb{R}_{+}^{n \times |V|},
\end{equation}
where ${\rm sg}(\bm{W}^s) \in \mathbb{R}^{d \times |V|}$ is the student's prediction head without gradient and $|V|$ is the vocabulary size of the two models.
% \footnote{Note that we stop the gradient of $\bm{W}^s$ in Eq. \ref{eq:p_t2s} to avoid negative effects to the student model.}.
Ideally, the projected teacher distribution should be the same as the original teacher distribution to keep its information. 
More strictly, the projector $\bm{W}^{t \rightarrow s}$ needs to fulfill the following equation to obtain the same logits as the ones in the teacher space:
\begin{equation}\label{eq:init-stu1}
    \forall h^t \in \mathbb{R}^{1 \times D}, h^t\bm{W}^{t \rightarrow s}\bm{W}^{s} = h^t\bm{W}^t,
\end{equation}
where $\bm{W}^t \in \mathbb{R}^{D \times |V|}$ is the teacher's prediction head.
Thus, we have
\begin{equation}\label{eq:init-stu2}
    \bm{W}^{t \rightarrow s}\bm{W}^{s}=\bm{W}^t.
\end{equation}
Based on this equation, we can initialize $\bm{W}^{t \rightarrow s}=\bm{W}^t{\bm{W}^s}^{+}$ at the beginning of distillation, where ${\bm{W}^s}^{+} \in \mathbb{R}^{|V| \times d}$ is the pseudo-inverse of $\bm{W}^s$.
However, as generally $d<D$, $\bm{W}^{t \rightarrow s}\bm{W}^{s}$ is actually a low-rank approximation of $\bm{W}^{t}$ with non-negligible errors.
Therefore, in the distillation process, we further train the projector $\bm{W}^{t \rightarrow s}$ with the cross-entropy loss to predict the top-1 predicted tokens of the teacher to recover the accuracy of $p^{t \rightarrow s}$:
\begin{equation} \label{eq:stuside_ce_loss}
    \mathcal{L}^{t \rightarrow s}_{ce}=- \frac{1}{n} \sum_{i=1}^n \log p^{t \rightarrow s}(y_i=\hat{y}^t_i|\mathbf{y}_{<i},\mathbf{x};\bm{W}^{t \rightarrow s}),
\end{equation}
where $\hat{y}^t_i$ denotes the top-1 predicted token of the teacher model at position $i$.
Meanwhile, the new teacher distribution $p^{t \rightarrow s}$ will guide the student model with the following loss:
\begin{equation} \label{eq:stuside_kd_loss}
    \mathcal{L}^{stu}_{kd}= \frac{1}{n} \sum_{i=1}^n \mathcal{D}({\rm sg}(p^{t \rightarrow s}(y_i|\mathbf{y}_{<i},\mathbf{x})) || q_{\theta}(y_i|\mathbf{y}_{<i},\mathbf{x})),
\end{equation}
where $\mathcal{D}(\cdot||\cdot)$ is as same as the divergence function in Eq. (\ref{eq:kd_loss}).

\textbf{KD in Teacher Space:} Similar to the process in the student space, we also project the hidden states of the student model $\bm{H}^{s}_{1:n}=\{h^s_1, h^s_2, ..., h^s_n\}$ into the teacher's space with another linear projector $\bm{W}^{s \rightarrow t}$:
\begin{equation}
    \bm{H}^{s \rightarrow t}_{1:n} = \bm{H}^{s}_{1:n}\bm{W}^{s \rightarrow t} \in \mathbb{R}^{n \times D}.
\end{equation}
% where $D$ is the hidden size of the teacher model.
Then, the projected distribution of the student model in the teacher's space can be obtained by
\begin{equation} \label{eq:p_s2t}
    q_{\theta}^{s \rightarrow t} = {\rm softmax}(\bm{H}^{s \rightarrow t}\bm{W}^t) \in \mathbb{R}_{+}^{n \times |V|}.
\end{equation}
Also, the projector $\bm{W}^{s \rightarrow t}$ should fulfill the following equation to ensure $q_{\theta}^{s \rightarrow t}\rightarrow q_{\theta}$:
\begin{equation} \label{eq:s2t_condition}
    \forall h^{s} \in \mathbb{R}^{1 \times d}, h^{s}\bm{W}^{s \rightarrow t}\bm{W}^{t}=h^{s}\bm{W}^{s}.
\end{equation}
Note that since $\text{rank}(\bm{W}^{s \rightarrow t}\bm{W}^{t}) \geq \text{rank}(\bm{W}^{s})$, we can initialize $\bm{W}^{s \rightarrow t}=\bm{W}^{s}{\bm{W}^{t}}^{+}$ with much less approximation error than $\bm{W}^{t \rightarrow s}$.
Therefore, different from the projector $\bm{W}^{t \rightarrow s}$, $\bm{W}^{s \rightarrow t}$ does not need training and can be calculated in real-time with $\bm{W}^{s}$ and ${\bm{W}^{t}}^{+}$ during distillation.
% During distillation, we further optimize $\bm{W}^{s \rightarrow t}$ with the following mean squared error:
% \begin{equation}
%     \mathcal{L}^{s \rightarrow t}_{mse}=||\bm{H}^{s}\bm{W}^{s \rightarrow t}\bm{W}^{t}-\bm{H}^{s}\bm{W}^{s}||.
% \end{equation}
On this basis, we only need to calculate the KD loss in the teacher space as
\begin{equation} \label{eq:teaside_kd_loss}
    \mathcal{L}^{tea}_{kd}= \frac{1}{n} \sum_{i=1}^n \mathcal{D}(p(y_i|\mathbf{y}_{<i},\mathbf{x}) || q_{\theta}^{s \rightarrow t}(y_i|\mathbf{y}_{<i},\mathbf{x})).
\end{equation}
% where a difference from Eq. \ref{eq:stuside_kd_loss} is that we directly fix KL divergence as $\mathcal{D}(\cdot||\cdot)$ since we found it more appropriate for KD in the teacher space.

\begin{algorithm}[t]
    \caption{Exact Token Alignment Algorithm}
    \label{alg:token-alignment}
    \renewcommand{\algorithmicrequire}{\textbf{Input:}}
    \renewcommand{\algorithmicensure}{\textbf{Output:}}
    
    \begin{algorithmic}[1]
        \REQUIRE student tokenized sequence $\mathbf{y}^s=[y^s_1,y^s_2,...,y^s_n]$, teacher tokenized sequence $\mathbf{y}^t=[y^t_1,y^t_2,...,y^t_m]$, student vocabulary $V_{stu}$
        \ENSURE aligned token pairs $\mathcal{A}$

        \STATE Initialize $\mathcal{A}=\varnothing$; $i, j=1$.
        \WHILE{$i <= n$ AND $j <= m$}
            \IF{$\mathbf{y}^s_{<i} = \mathbf{y}^t_{<j}$ AND $y^s_i = y^t_j$}
                \IF{$\hat{y}^t_j \in V_{stu}$}
                    \STATE $\mathcal{A} \leftarrow \mathcal{A} \cup \{(i, j)\}$
                \ENDIF
                % \STATE $\mathbf{x}_{<i} \leftarrow \mathbf{x}_{<i} \oplus x_i$, $\mathbf{y}_{<j} \leftarrow \mathbf{y}_{<j} \oplus y_j$
                \STATE $i \leftarrow i + 1$, $j \leftarrow j + 1$
            \ELSIF{$|\mathbf{y}^s_{<i}| > |\mathbf{y}^t_{<j}|$}
                \STATE $j \leftarrow j + 1$
            \ELSIF{$|\mathbf{y}^s_{<i}| < |\mathbf{y}^t_{<j}|$}
                \STATE $i \leftarrow i + 1$
            \ELSE
                % \STATE $\mathbf{x}_{<i} \leftarrow \mathbf{x}_{<i} + x_i$, $\mathbf{y}_{<j} \leftarrow \mathbf{y}_{<j} + y_j$
                \STATE $i \leftarrow i + 1$, $j \leftarrow j + 1$
            \ENDIF
        \ENDWHILE
\RETURN $\mathcal{A}$
    \end{algorithmic}
\end{algorithm}

Finally, the overall loss of DSKD sums up the aforementioned losses in both spaces:
\begin{equation} \label{eq:dskd_loss}
    \mathcal{L}_{dskd}=\mathcal{L}^{stu}_{kd} + \mathcal{L}^{t \rightarrow s}_{ce} + \mathcal{L}^{tea}_{kd}.
\end{equation}

% \begin{theorem}
%     With unified
% \end{theorem}

\subsection{Exact Token Alignment Algorithm}
\label{sec:method_cross_model}
% \subsubsection{Cross-Model Attention}
After introducing the DSKD framework for LLMs with the same vocabulary, in this section, we will illustrate how to apply DSKD on LLMs with different vocabularies, \emph{i.e.}, cross-tokenizer KD.
Benefiting from the head-sharing mechanism in DSKD, the distributions from the student and the teacher always have the same dimensions.
Thus, the remaining requirement for cross-tokenizer KD is just to align the tokens in two differently tokenized sequences.
% \footnote{Here we borrow the notations in Section \ref{sec:depend_same_vocab} and assume that there are $m$ tokens in the teacher's sequence.}.

In practice, although tokenized by different tokenizers, the two sequences from the same sentence tend to overlap on most tokens\footnote{We observe a general phenomenon that over 90\% of tokens can be exactly matched in two differently tokenized sequences.}.
Therefore, we develop an exact token alignment algorithm (ETA) to discover the aligned tokens in the two sequences and only conduct KD on these positions.
Formally, given the two sequences $\mathbf{y}^s=[y^s_1,y^s_2,...,y^s_n]$ and $\mathbf{y}^t=[y^t_1,y^t_2,...,y^t_m]$ tokenized by the student's and the teacher's tokenizers respectively, the goal of ETA is to find aligned pairs of tokens $\mathcal{A}$ that satisfy the following requirements:
\begin{equation}
   \mathcal{A} = \{(i, j) \mid \mathbf{y}^t_{<j} = \mathbf{y}^s_{<i},  y^t_j = y^s_i,  \hat{y}_j^t \in V_{stu}\}.
\end{equation}
where $\hat{y}^t_j \in V_{stu}$ denotes the top-1 token predicted by the teacher model at position $j$.
The complete procedure of the ETA algorithm is present in Algorithm \ref{alg:token-alignment}.
We sequentially traverse the sequences $\mathbf{y}^s$ and $\mathbf{y}^t$ to find all aligned token pairs $\mathcal{A} = \{(i, j)\}_{K}$ according to the above requirements.

\begin{figure}[t]
    \centering
    \includegraphics[width=\linewidth]{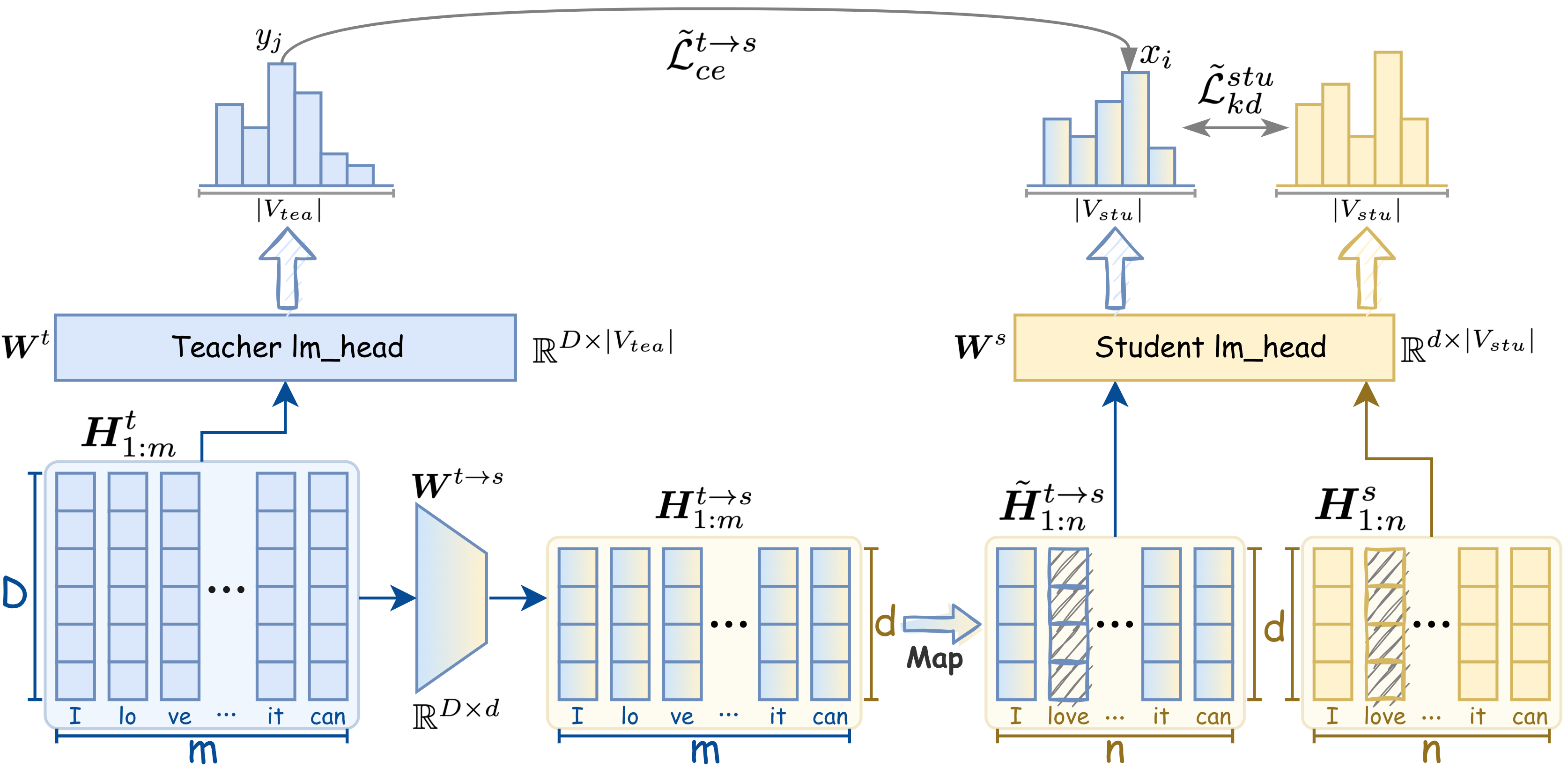}
    \caption{The framework of our DSKD for LLMs with different vocabularies in student space.}
    \label{fig:cma_pic}
\end{figure}

Next, we introduce how to adapt our DSKD to LLMs with different vocabularies utilizing the aligned token pairs.
In the student space (as shown in Fig. \ref{fig:cma_pic}), we first use the projector $W^{t \rightarrow s}$ to project the teacher hidden states $\bm{H}_{1:m}^t \in \mathbb{R}^{m \times D}$ to $\bm{H}_{1:m}^{t \rightarrow s} \in \mathbb{R}^{m \times d}$ following Eq. (\ref{eq:down_transform}).
Afterwards, given the alignment set $\mathcal{A}$, we can select the exactly aligned hidden states from $\bm{H}_{1:m}^{t \rightarrow s}$ and $\bm{H}_{1:n}^{s}$:
\begin{align}
    \bm{H}_{1:K}^{t \rightarrow s} &= \{h^{t \rightarrow s}_{j} \mid \exists \, i, \, (i, j) \in \mathcal{A} \}, \notag \\
    \bm{H}_{1:K}^{s} &= \{h^{s}_{i} \mid \exists \, j, \, (i, j) \in \mathcal{A} \}. \notag
\end{align}
% Following the aligned tokens $\mathcal{A} = \{(i, j)\}_K$, we map $\bm{H}_{1:m}^{t \rightarrow s} \in \mathbb{R}^{m \times d}$ to $\tilde{\bm{H}}_{1:n}^{t \rightarrow s} \in \mathbb{R}^{n \times d}$, which has the same dimension with the student hidden states. 
% Specifically, for the aligned position $i$ in the student sequence and $j$ in the teacher sequence, we set $(\tilde{\bm{H}}_{1:n}^{t \rightarrow s})_i = (\bm{H}_{1:m}^{t \rightarrow s})_j$.
% And for the non-aligned positions, we set the corresponding hidden state vectors to the zero vectors with the dimension $d$.
Hence, the two lists of hidden states $\bm{H}_{1:K}^{t \rightarrow s}$ and $\bm{H}_{1:K}^{s}$ are strictly aligned one-by-one.
Accordingly, we can calculate the projected teacher distribution $\tilde{p}^{t \rightarrow s}$ and the selected student distribution $\tilde{q}^{s}_{\theta}$:
\begin{align}
    \tilde{p}^{t \rightarrow s} &= {\rm softmax}(\tilde{\bm{H}}^{t \rightarrow s}_{1:K}{\rm sg}(\bm{W}^s)) \in \mathbb{R}_{+}^{K \times |V_{stu}|}, \notag \\
    \tilde{q}^{s}_{\theta} &= {\rm softmax}(\tilde{\bm{H}}^{s}_{1:K}\bm{W}^s) \in \mathbb{R}_{+}^{K \times |V_{stu}|} \notag.
\end{align}
On this basis, the cross-entropy loss in Eq. (\ref{eq:stuside_ce_loss}) is re-defined as follows:
\begin{equation} \label{eq:stuside_ce_loss_cv}
    \tilde{\mathcal{L}}^{t \rightarrow s}_{ce}=- \frac{1}{K} \sum_{(i,j) \in \mathcal{A}} \log \hat{p}^{t \rightarrow s}(y^{s}_i=\hat{y}^{t}_j|\mathbf{y}^{s}_{<i},\mathbf{x};\bm{W}^{t \rightarrow s}),
\end{equation}
where $\hat{y}^t_j \in V_{stu}$ denotes the top-1 token predicted by the teacher model at position $j$ in its own sequence, which aligns with the $i$-th token in the student sequence.
Considering that $\tilde{p}^{t \rightarrow s}$ usually has lower accuracy and convergence speed than its counterpart $p^{t \rightarrow s}$ in the same-vocabulary scenario, we add a mask $M$ to the KD loss in the student space:
\begin{equation}
    \tilde{\mathcal{L}}^{stu}_{kd}=\frac{1}{K} \sum_{(i,j) \in \mathcal{A}} M_i \mathcal{D}({\rm sg}(\tilde{p}^{t \rightarrow s}(y^s_i|\mathbf{y}^s_{<i},\mathbf{x})) || \tilde{q}_{\theta}(y^s_i|\mathbf{y}^s_{<i},\mathbf{x})), \notag
\end{equation}
where $M_i=1$ when $\text{argmax} \, \tilde{p}^{t \rightarrow s}(y_i^s|\mathbf{y}^s_{<i},\mathbf{x})=\hat{y}^t_j$ and otherwise $M_i=0$. 

Additionally, due to the different vocabularies between the teacher model and the student model, \textit{i.e.}, $V_{tea} \ne V_{stu}$, we cannot directly initialize $\bm{W}^{t \rightarrow s}=\bm{W}^t{\bm{W}^s}^{+}$. 
To address this, we take the intersection of two vocabularies $V_{tea} \cap V_{stu}$ as $\tilde{V}$, and then initialize $\bm{W}^{t \rightarrow s}$ as follows:
\begin{equation}
    \bm{W}^{t \rightarrow s}=\tilde{\bm{W}}^t\tilde{\bm{W}}^{s^+},
\end{equation}
where $\tilde{\bm{W}}^t \in \mathbb{R}^{D \times |\tilde{V}|}$ is the teacher prediction head at the overlapped positions, and $\tilde{\bm{W}}^{s^+} \in \mathbb{R}^{|\tilde{V}| \times d}$ is the pseudo-inverse of the overlapped student prediction head $\tilde{\bm{W}}^s$.

Similarly in the teacher space, we first project $\bm{H}_{1:n}^s \in \mathbb{R}^{n \times d}$ to ${\bm{H}}_{1:n}^{s \rightarrow t} \in \mathbb{R}^{n \times D}$ by $\bm{W}^{s \rightarrow t} \in \mathbb{R}^{d \times D} $. 
Then we select aligned hidden states from ${\bm{H}}_{1:n}^s \in \mathbb{R}^{n \times D}$ to construct $\tilde{\bm{H}}_{1:K}^s \in \mathbb{R}^{K \times D}$ according the aligned token pairs.
Likewise, we can obtain the teacher distribution $\tilde{p}$ and the projected student distribution $\tilde{q}_{\theta}^{s \rightarrow t}$.
% For example in aligned position $j$, $(\tilde{\bm{H}}_{1:m}^{s \rightarrow t})_j = (\bm{H}_{1:n}^{t \rightarrow s})_i$. 
% And the hidden states on non-aligned positions are also zero vectors with dimension $D$.
Subsequently, the KD loss in the teacher space is defined as follows:
\begin{equation}
    \tilde{\mathcal{L}}^{tea}_{kd}=\frac{1}{K} \sum_{(i,j) \in \mathcal{A}} \mathcal{D}(\tilde{p}(y^t_j|\mathbf{y}^t_{<j},\mathbf{x}) || \tilde{q}_{\theta}^{s \rightarrow t}(y^t_j|\mathbf{y}^t_{<j},\mathbf{x})). \notag
\end{equation}
% $\tilde{\mathcal{L}}^{tea}_{kd} = M^{s \rightarrow t} {\mathcal{L}}^{tea}_{kd}$ in the non-zero vector positions, \textit{i.e.}, the aligned positions.
Additionally, we initialize $\bm{W}^{s \rightarrow t}=\tilde{\bm{W}}^{s}\tilde{\bm{W}}^{t^+}$ with similar setting in the student space.

In summary, the overall loss of DSKD for LLMs with different vocabularies sums the losses in both spaces:
\begin{equation} \label{eq:dskd_loss_dv}
    \tilde{\mathcal{L}}_{dskd}=\tilde{\mathcal{L}}^{stu}_{kd} + \tilde{\mathcal{L}}^{t \rightarrow s}_{ce} + \tilde{\mathcal{L}}^{tea}_{kd}.
\end{equation}

\subsection{On-Policy Dual-Space Knowledge Distillation} \label{sec:on-policy}
Previous work has revealed the potential of on-policy knowledge distillation for LLMs with the same vocabulary \cite{lin20imitkd,wen23fdiv,gu23minillm,agarwal24gkd}.
However, similar exploration for cross-tokenizer KD is still lacking.
In the following part, we showcase that our DSKD framework, as a general white-box KD framework, can be extended to the on-policy scenario for any two LLMs with the same or different vocabularies.
% Thus, we further expand our DSKD framework to the on-policy scenario.

On-policy knowledge distillation aims to minimize the expected discrepancy between the student model and the teacher model under the outputs sampled from the student's distribution $q_{\theta}$:
\begin{equation} \label{eq:token_level_op_kd_loss}
    \mathcal{L}^{op}_{kd}=\frac{1}{|\hat{\mathbf{y}}|} \sum_{i=1}^{|\hat{\mathbf{y}}|} \mathcal{D}(p(y_i|\hat{\mathbf{y}}_{<i},\mathbf{x})||q_{\theta}(y_i|\hat{\mathbf{y}}_{<i},\mathbf{x})),
\end{equation}
where $\hat{\mathbf{y}}$ is sampled from $q_{\theta}(\cdot|\mathbf{x})$.
Actually, this objective is fully compatible with DSKD.
Thus, we can easily modify our off-policy DSKD to on-policy DSKD by switching the ground-truth output $\mathbf{y}$ to the sampled output $\hat{\mathbf{y}}$ from the student model.
Thus, the training loss for on-policy DSKD is to substitute the ground-truth prefix $\mathbf{y}_{<i}$ in off-policy DSKD with the model-generated prefix $\hat{\mathbf{y}}_{<i}$.
For example, in the on-policy scenario, the KD loss in the student space becomes:
\begin{equation} \label{eq:op_stuside_kd_loss}
    \mathcal{L}^{stu}_{op\mbox{-}kd}= \frac{1}{|\hat{\mathbf{y}}|} \sum_{i=1}^{|\hat{\mathbf{y}}|} \mathcal{D}({\rm sg}(p^{t \rightarrow s}(y_i|\hat{\mathbf{y}}_{<i},\mathbf{x})) || q_{\theta}(y_i|\hat{\mathbf{y}}_{<i},\mathbf{x})). \notag
\end{equation}
Moreover, for cross-tokenizer KD, the initial sequences $\mathbf{y}^s$ and $\mathbf{y}^t$ in the ETA algorithm should be changed to the student-generated outputs $\hat{\mathbf{y}}^s$ and $\hat{\mathbf{y}}^t$.

\section{Experiments}
\subsection{Experimental Setup}
\textbf{Data:} In the main experiments, we evaluate our DSKD framework on several instruction-following datasets following \cite{gu23minillm}.
Specifically, we choose $\mathtt{databricks}$-$\mathtt{dolly}$-$\mathtt{15k}$ dataset processed by \cite{gu23minillm} to conduct the KD process, which contains about 11k samples for training, 1k for validation, and 500 for testing.
Besides, we also select Self-Instruct (\textbf{SelfInst}) \cite{wang23selfinst}, Vicuna-Evaluation ({\bf VicunaEval}) \cite{wang23selfinst}, Super Natural Instructions ({\bf S-NI}) \cite{wang22sni}, and Unnatural Instructions ({\bf UnNI}) \cite{honovich23unni} as the additional test sets for more comprehensive evaluation.
All the test sets in our experiments are processed by \cite{gu23minillm}. 
For all these test sets, Dolly contains 500 samples, SelfInst contains 242 samples, VicunaEval contains 80 samples, S-NI contains 1694 samples with response lengths in $[11, +\infty]$, and UnNI contains 10000 samples with response lengths in $[11, +\infty]$.
Note that we do not use the additional pre-training corpus during distillation as \cite{gu23minillm} did for simplicity.

\textbf{Models:} For student LLMs, we select both GPT2-120M \cite{radford19gpt2} and TinyLLaMA-1.1B \cite{zhang24tinyllama}.
For GPT2-120M, we employ GPT2-1.5B and Qwen2-1.5B \cite{qwen2} respectively as the teacher LLMs that have the same/different vocabularies with the student LLM.
(For GPT2-1.5B, we directly use the checkpoint released by \cite{gu23minillm}.)
For TinyLLaMA-1.1B, we choose LLaMA2-7B \cite{touvron23llama} and Mistral-7B \cite{jiang23mistral} as the teacher LLMs that have the same/different vocabularies with the student LLM.

\begin{table}[ht]
\caption{Detailed training configurations of KD for GPT2 and TinyLLaMA.}\label{tab:train_config}
    \centering
    \resizebox{\linewidth}{!}{
        \begin{tabular}{c|cc|cc}
            \bottomrule
            \multirow{2}{*}{\textbf{Settings}} & \multicolumn{2}{c|}{\textbf{KD for GPT2}} & \multicolumn{2}{c}{\textbf{KD for TinyLLaMA}} \\
            \cline{2-5}
            & GPT2-1.5B & Qwen2-1.5B & LLaMA2-7B & Mistral-7B \\ 
            \hline
            Epoch & 20 & 20 & 10 & 10 \\
            Learning Rate & 5e-4 & 5e-4 & 1e-3 & 1e-3 \\
            Projector Learning Rate & 5e-4 & 5e-4  & 1e-3 & 1e-3 \\
            Batch Size & 64 & 64 & 32 & 32 \\
            LR Scheduler & Cosine & Cosine  & Cosine & Cosine \\
            Fine-Tuning Method & Full & Full & LoRA & LoRA \\
            Lora Rank & N/A & N/A & 256 & 256 \\
            Lora Alpha & N/A & N/A & 8 & 8 \\
            Lora Dropout & N/A & N/A & 0.1 & 0.1 \\
            \toprule
        \end{tabular}
    }
\end{table}

\textbf{Training:} For KD on GPT2, we employ full-finetuning for the teachers and the students (\textit{e.g.}, ``Teacher'' and ``SFT'' in Table \ref{tab:main_results_off_policy_same_vocab}).
For KD on TinyLLaMA, we fine-tune the students and the teachers with LoRA.
In particular, we set the temperature $\tau$ to 2.0 according to the performance on the validation set (The detailed results are plotted in Fig. \ref{fig:temperature} of Appendix \ref{sec:temperature}).
The detailed training configurations are listed in Table \ref{tab:train_config}.
Besides, our DSKD only introduces one additional trainable matrix $\bm{W}^{t \rightarrow s} \in \mathbb{R}^{D \times d}$ beyond the student model, whose parameters are negligible in KD training (\emph{e.g.}, $\approx$1M for DSKD on GPT2). 
Each training requires several hours on 8$\times$32G V100 or 8$\times$40G A100.

Additionally, we combine the original cross-entropy loss on the ground-truth outputs in Eq. (\ref{eq:ce_loss}) and the KD loss in Eq. (\ref{eq:kd_loss}) or Eq. (\ref{eq:dskd_loss}) or Eq. (\ref{eq:dskd_loss_dv}) as the overall training loss $\mathcal{L} = 0.5 * \mathcal{L}_{ce} + 0.5 * \mathcal{L}_{(ds)kd}$ for all the white-box KD methods in our main experiments.

\textbf{Evaluation:} We use the sampling-based decoding strategy to generate the responses for all models.
To be specific, we set both the decoding temperature and top\_p to 1.0.  
Then, we generate the responses with random seeds in [10, 20, 30, 40, 50] and report the averaged Rouge-L \cite{lin04rouge} scores and the standard deviations among five seeds following \cite{gu23minillm}.
The Rouge-L is calculated between the generated responses and the human-labeled ones.

\begin{table*}[t]
\caption{The Rouge-L scores (\%) on several benchmarks on the same-vocabulary \textbf{off-policy} KD with GPT2-120M and TinyLLaMA-1.1B as students. We list the mean values and the standard deviations among 5 random seeds. The average scores (\textbf{Avg.}) on all benchmarks are also listed. ``\emph{w/} DSKD'' denotes our DSKD using the corresponding divergence as $\mathcal{D}(\cdot||\cdot)$ in Eq. (\ref{eq:stuside_kd_loss}) and (\ref{eq:teaside_kd_loss}). ``{\color{midgray2}(+*$\uparrow$)}'' means the improvement of our DSKD than the baseline divergence.}
    \centering
    \resizebox{0.9\linewidth}{!}{
        \begin{tabular}{lccccc|l}
            \bottomrule
            \textbf{Methods} & \textbf{Dolly} & \textbf{SelfInst} & \textbf{VicunaEval} & \textbf{S-NI} & \textbf{UnNI} & \quad \quad \textbf{Avg.} \\
            \hline
            \hline
            % \rowcolor{lightblue}
            % \rowcolor{lightgray}
            \multicolumn{7}{c}{\textbf{GPT2-1.5B $\rightarrow$ GPT2-120M}} \\
            \hline
            \rowcolor{lightgray}
            {\color{midgray} Teacher} & {\color{midgray} 27.19$_{\pm 0.23}$} & {\color{midgray} 14.64$_{\pm 0.64}$} & {\color{midgray} 16.30$_{\pm 0.37}$} & {\color{midgray} 27.55$_{\pm 0.30}$} & {\color{midgray} 31.42$_{\pm 0.11}$} & {\color{midgray} 23.42} \\
            \hline
            SFT &  23.19$_{\pm 	0.32	}$ & 	10.09$_{\pm 	0.31	}$ & 	15.08$_{\pm 	0.62	}$ & 	17.25$_{\pm 	0.24	}$ & 	20.21$_{\pm 	0.13 	}$ & 	17.16				\\
            \hline
            SeqKD &  23.03$_{\pm 	0.33	}$ & 	9.71$_{\pm 	0.37	}$ & 	15.58$_{\pm 	0.32	}$ & 	16.27$_{\pm 	0.12	}$ & 	18.73$_{\pm 	0.14 	}$ & 	16.66				\\
            \hline
            KL	&	24.15$_{\pm 	0.32	}$ & 	10.69$_{\pm 	0.56	}$ & 	16.05$_{\pm 	0.50	}$ & 	19.29$_{\pm 	0.14	}$ & 	22.34$_{\pm 	0.07 	}$ & 	18.50				\\
            \quad \quad \emph{w/} DSKD (ours)	&	24.68$_{\pm 	0.58	}$ & 	10.98$_{\pm 	0.55	}$ & 	15.87$_{\pm 	0.28	}$ & 	21.73$_{\pm 	0.24	}$ & 	24.12$_{\pm 	0.11 	}$ & 	19.48	{\color{midgray2}(+0.98$\uparrow$)}	\\
            \hline
            RKL	&	24.41$_{\pm 	0.22	}$ & 	11.01$_{\pm 	0.40	}$ & 	15.16$_{\pm 	0.46	}$ & 	19.28$_{\pm 	0.32	}$ & 	22.56$_{\pm 	0.15 	}$ & 	18.48				\\
            \quad \quad \emph{w/} DSKD (ours)	&	25.19$_{\pm 	0.30	}$ & 	11.25$_{\pm 	0.36	}$ & 	15.90$_{\pm 	0.44	}$ & 	23.97$_{\pm 	0.20	}$ & 	25.17$_{\pm 	0.08 	}$ & 	20.30	{\color{midgray2}(+\textbf{1.82}$\uparrow$)}	\\
            \hline
            SKL	&	23.66$_{\pm 	0.24	}$ & 	11.62$_{\pm 	0.60	}$ & 	15.17$_{\pm 	0.19	}$ & 	20.37$_{\pm 	0.21	}$ & 	22.27$_{\pm 	0.23 	}$ & 	18.62				\\
            \quad \quad \emph{w/} DSKD (ours)	&	24.78$_{\pm 	0.26	}$ & 	12.10$_{\pm 	0.33	}$ & 	15.45$_{\pm 	0.36	}$ & 	21.25$_{\pm 	0.11	}$ & 	24.58$_{\pm 	0.11 	}$ & 	19.63	{\color{midgray2}(+1.01$\uparrow$)}	\\
            \hline
            SRKL	&	24.19$_{\pm 	0.43	}$ & 	10.71$_{\pm 	0.39	}$ & 	15.38$_{\pm 	0.43	}$ & 	19.18$_{\pm 	0.24	}$ & 	22.27$_{\pm 	0.04 	}$ & 	18.35				\\
            \quad \quad \emph{w/} DSKD (ours)	&	24.90$_{\pm 	0.45	}$ & 	11.13$_{\pm 	0.25	}$ & 	15.49$_{\pm 	0.17	}$ & 	22.15$_{\pm 	0.26	}$ & 	24.38$_{\pm 	0.06 	}$ & 	19.61	{\color{midgray2}(+1.26$\uparrow$)}	\\
            \hline
            AKL	&	24.29$_{\pm 	0.35	}$ & 	11.26$_{\pm 	0.41	}$ & 	15.82$_{\pm 	0.56	}$ & 	19.29$_{\pm 	0.29	}$ & 	22.36$_{\pm 	0.15 	}$ & 	18.60				\\
            \quad \quad \emph{w/} DSKD (ours)	&	25.30$_{\pm 	0.33	}$ & 	11.91$_{\pm 	0.23	}$ & 	16.06$_{\pm 	0.52	}$ & 	22.67$_{\pm 	0.23	}$ & 	25.44$_{\pm 	0.06 	}$ & 	20.28	{\color{midgray2}(+1.68$\uparrow$)}	\\
            \hline
            % \rowcolor{lightgray}
            % \rowcolor{lightpink} 
            % \rowcolor{lightblue}
            % \rowcolor{lightgray}
            \multicolumn{7}{c}{\textbf{LLaMA2-7B $\rightarrow$ TinyLLaMA-1.1B}} \\
            \hline
            \rowcolor{lightgray}
            Teacher & 28.32$_{\pm 0.46}$ & 20.95$_{\pm 0.69}$ & 18.76$_{\pm 0.35}$ & 32.05$_{\pm 0.28}$ & 32.41$_{\pm 0.12}$ & 26.50 \\
            \hline
            SFT & 23.20$_{\pm 0.13}$ & 14.88$_{\pm 0.54}$ & 16.42$_{\pm 0.35}$ & 27.79$_{\pm 0.27}$ & 26.12$_{\pm 0.11}$ & 21.68 \\
            \hline
            SeqKD & 23.21$_{\pm 0.22}$ & 16.46$_{\pm 0.72}$ & 16.58$_{\pm 0.38}$ & 26.33$_{\pm 0.26}$ & 27.69$_{\pm 0.10}$ & 22.05 \\
            \hline
            KL & 25.46$_{\pm 0.63}$ & 17.21$_{\pm 0.25}$ & 16.43$_{\pm 0.53}$ & 29.27$_{\pm 0.29}$ & 29.28$_{\pm 0.09}$ & 23.53 \\
            \quad \quad \emph{w/} DSKD (ours)	&	26.80$_{\pm 	0.57	}$ & 	18.81$_{\pm 	1.17	}$ & 	17.86$_{\pm 	0.11	}$ & 	31.90$_{\pm 	0.17	}$ & 	31.80$_{\pm 	0.08 	}$ & 	25.43	{\color{midgray2}(+1.90$\uparrow$)}	\\
            \hline
            RKL & 24.49$_{\pm 0.41}$ & 17.14$_{\pm 0.61}$ & 16.87$_{\pm 0.26}$ & 29.50$_{\pm 0.28}$ & 29.36$_{\pm 0.08}$ & 23.47 \\
            \quad \quad \emph{w/} DSKD (ours)	&	27.02$_{\pm 	0.60	}$ & 	19.57$_{\pm 	0.94	}$ & 	18.71$_{\pm 	0.35	}$ & 	33.74$_{\pm 	0.21	}$ & 	33.30$_{\pm 	0.06 	}$ & 	26.47	{\color{midgray2}(+\textbf{3.00}$\uparrow$)}	\\
            \hline
            SKL \cite{ko24distillm} & 24.14$_{\pm 0.53}$ & 15.98$_{\pm 0.72}$ & 16.89$_{\pm 0.22}$ & 29.30$_{\pm 0.18}$ & 28.71$_{\pm 0.12}$ & 23.01 \\
            \quad \quad \emph{w/} DSKD (ours)	&	25.50$_{\pm 	0.37	}$ & 	17.87$_{\pm 	1.05	}$ & 	17.40$_{\pm 	0.33	}$ & 	30.23$_{\pm 	0.09	}$ & 	30.60$_{\pm 	0.08 	}$ & 	24.32	{\color{midgray2}(+1.31$\uparrow$)}	\\
            \hline
            SRKL \cite{ko24distillm} & 24.28$_{\pm 0.58}$ & 16.91$_{\pm 0.67}$ & 16.88$_{\pm 0.20}$ & 29.55$_{\pm 0.19}$ & 28.64$_{\pm 0.21}$ & 23.25 \\
            \quad \quad \emph{w/} DSKD (ours)	&	25.83$_{\pm 	0.31	}$ & 	17.95$_{\pm 	0.55	}$ & 	17.18$_{\pm 	0.29	}$ & 	29.54$_{\pm 	0.28	}$ & 	30.54$_{\pm 	0.12 	}$ & 	24.21	{\color{midgray2}(+0.96$\uparrow$)}	\\
            \hline
            AKL \cite{wu2024rethinking} & 24.80$_{\pm 0.70}$ & 16.79$_{\pm 1.09}$ & 16.80$_{\pm 0.44}$ & 29.29$_{\pm 0.35}$ & 28.81$_{\pm 0.09}$ & 23.30 \\
            % DSKD (ours) & & & & & \\
            % \rowcolor{lightlightgray}
            \quad \quad \emph{w/} DSKD (ours)	&	26.73$_{\pm 	0.29	}$ & 	19.08$_{\pm 	0.92	}$ & 	18.30$_{\pm 	0.54	}$ & 	32.59$_{\pm 	0.22	}$ & 	31.55$_{\pm 	0.02 	}$ & 	25.65	{\color{midgray2}(+2.35$\uparrow$)}	\\
            \toprule
        \end{tabular}
    }
    \label{tab:main_results_off_policy_same_vocab}
\end{table*}

\begin{table*}[t]
\caption{The Rouge-L scores (\%) on several benchmarks on the same-vocabulary \textbf{on-policy} KD with GPT2-120M and TinyLLaMA-1.1B as students. ``+ *KL'' denotes our DSKD using the corresponding divergence as $\mathcal{D}(\cdot||\cdot)$ in Eq. (\ref{eq:stuside_kd_loss}) and (\ref{eq:teaside_kd_loss}). The result with $^*$ marked represents it \textbf{surpasses} the performance of the teacher. The \textbf{bold} denotes the best result on each model. ``{\color{newblue}(+*$\uparrow$)}'' means the improvement of our on-policy DSKD than off-policy DSKD in Table \ref{tab:main_results_off_policy_same_vocab}.}
    \centering
    \resizebox{\linewidth}{!}{
        \begin{tabular}{llccccc|l}
            \bottomrule
            \textbf{Methods} &  & \textbf{Dolly} & \textbf{SelfInst} & \textbf{VicunaEval} & \textbf{S-NI} & \textbf{UnNI} &  \textbf{Avg.} \\
            \hline
            \hline
            % \rowcolor{lightblue}
            % \rowcolor{lightgray}
            \multicolumn{8}{c}{\textbf{GPT2-1.5B $\rightarrow$ GPT2-120M}} \\
            \hline
            \rowcolor{lightgray}
            {\color{midgray} Teacher}  &  & {\color{midgray} 27.19$_{\pm 0.23}$} & {\color{midgray} 14.64$_{\pm 0.64}$} & {\color{midgray} 16.30$_{\pm 0.37}$} & {\color{midgray} 27.55$_{\pm 0.30}$} & {\color{midgray} 31.42$_{\pm 0.11}$} & {\color{midgray} 23.42} \\
            \hline
            GKD \cite{agarwal24gkd} & & 25.40$_{\pm 	0.21	}$ & 	11.88$_{\pm 	0.19	}$ & 	17.23$_{\pm 	0.46	}$ & 	20.81$_{\pm 	0.40	}$ & 	23.19$_{\pm 	0.12 	}$ & 	19.70 \\
            MiniLLM \cite{gu23minillm} & & 24.84$_{\pm 	0.18	}$ & 	11.24$_{\pm 	0.40	}$ & 	16.44$_{\pm 	0.61	}$ & 	21.04$_{\pm 	0.40	}$ & 	24.65$_{\pm 	0.24 	}$ & 	19.64 \\
            DistiLLM \cite{ko24distillm} & & 24.71$_{\pm 	0.22	}$ & 	11.19$_{\pm 	0.42	}$ & 	16.08$_{\pm 	0.36	}$ & 	20.75$_{\pm 	0.25	}$ & 	22.57$_{\pm 	0.13 	}$ & 	19.06 \\
            \hline
            \multirow{5}{*}{\makecell{On-Policy DSKD (ours)}} & + KL	&	25.61$_{\pm 	0.31	}$ & 	12.81$_{\pm 	0.21	}$ & 	17.25$_{\pm 	0.24	}$ & 	23.26$_{\pm 	0.36	}$ & 	25.64$_{\pm 	0.10 	}$ & 	20.91	{\color{newblue}(+1.43$\uparrow$)}	\\
            & + RKL	&	25.51$_{\pm 	0.32	}$ & 	11.99$_{\pm 	0.29	}$ & 	17.86$_{\pm 	0.32	}$ & 	23.83$_{\pm 	0.36	}$ & 	26.19$_{\pm 	0.11 	}$ & 	\textbf{21.08} {\color{newblue}(+0.78$\uparrow$)}	\\
            & + SKL	&	25.48$_{\pm 	0.44	}$ & 	12.10$_{\pm 	0.62	}$ & 	16.85$_{\pm 	0.44	}$ & 	23.98$_{\pm 	0.17	}$ & 	26.60$_{\pm 	0.07 	}$ & 	21.00 {\color{newblue}(+1.37$\uparrow$)}	\\
            & + SRKL 	&	25.77$_{\pm 	0.22	}$ & 	12.48$_{\pm 	0.22	}$ & 	17.46$_{\pm 	0.37	}$ & 	23.00$_{\pm 	0.44	}$ & 	25.95$_{\pm 	0.07 	}$ & 	20.93	{\color{newblue}(+1.32$\uparrow$)}	\\
            & + AKL	&	25.80$_{\pm 	0.26	}$ & 	11.89$_{\pm 	0.40	}$ & 	17.51$_{\pm 	0.41	}$ & 	21.69$_{\pm 	0.19	}$ & 	25.35$_{\pm 	0.10 	}$ & 	20.45	{\color{newblue}(+0.17$\uparrow$)} \\
            \hline

            % \rowcolor{lightgray}
            % \rowcolor{lightpink} 
            % \rowcolor{lightblue}
            % \rowcolor{lightgray}
            \multicolumn{8}{c}{\textbf{LLaMA2-7B $\rightarrow$ TinyLLaMA-1.1B}} \\
            \hline
            \rowcolor{lightgray}
            Teacher &   & 28.32$_{\pm 0.46}$ & 20.95$_{\pm 0.69}$ & 18.76$_{\pm 0.35}$ & 32.05$_{\pm 0.28}$ & 32.41$_{\pm 0.12}$ & 26.50 \\
            \hline
            GKD \cite{agarwal24gkd} & & 23.45$_{\pm 	0.60	}$ & 	16.70$_{\pm 	0.30	}$ & 	15.61$_{\pm 	0.51	}$ & 	28.18$_{\pm 	0.18	}$ & 	27.60$_{\pm 	0.23 	}$ & 	22.31 \\
            MiniLLM \cite{gu23minillm} & & 23.28$_{\pm 	0.49	}$ & 	17.70$_{\pm 	0.46	}$ & 	15.34$_{\pm 	0.52	}$ & 	31.57$_{\pm 	0.29	}$ & 	29.40$_{\pm 	0.08 	}$ & 	23.46 \\
            DistiLLM \cite{ko24distillm} & & 24.63$_{\pm 	0.32	}$ & 	17.43$_{\pm 	0.83	}$ & 	16.29$_{\pm 	0.25	}$ & 	31.07$_{\pm 	0.23	}$ & 	29.85$_{\pm 	0.17 	}$ & 	23.85 \\
            \hline
            \multirow{5}{*}{\makecell{On-Policy  DSKD (ours)}} & + KL 	&	27.44$_{\pm 	0.27	}$ & 	20.06$_{\pm 	0.29	}$ & 	18.78$_{\pm 	0.28	}$ & 	33.56$_{\pm 	0.22	}$ & 	34.70$_{\pm 	0.11 	}$ & 	26.91$^*$ {\color{newblue}(+1.48$\uparrow$)}	\\
            & + RKL 	&	28.33$_{\pm 	0.30	}$ & 	19.47$_{\pm 	0.52	}$ & 	19.71$_{\pm 	0.24	}$ & 	33.53$_{\pm 	0.24	}$ & 	35.82$_{\pm 	0.05 	}$ & 	27.37$^*$ {\color{newblue}(+0.90$\uparrow$)}	\\
            & + SKL	&	27.42$_{\pm 	0.34	}$ & 	20.09$_{\pm 	0.71	}$ & 	18.49$_{\pm 	0.43	}$ & 	32.69$_{\pm 	0.32	}$ & 	34.09$_{\pm 	0.13 	}$ & 	26.56$^*$ {\color{newblue}(+2.24$\uparrow$)}	\\
            & + SRKL	&	27.44$_{\pm 	0.38	}$ & 	18.67$_{\pm 	0.80	}$ & 	18.16$_{\pm 	0.26	}$ & 	32.25$_{\pm 	0.43	}$ & 	33.50$_{\pm 	0.17 	}$ & 	26.00\color{white}{$^*$} {\color{newblue}(+1.79$\uparrow$)}	\\
            & + AKL	&	28.77$_{\pm 	0.47	}$ & 	20.22$_{\pm 	0.75	}$ & 	19.66$_{\pm 	0.24	}$ & 	33.67$_{\pm 	0.30	}$ & 	34.76$_{\pm 	0.16 	}$ & 	\textbf{27.42}$^*$ {\color{newblue}(+1.77$\uparrow$)}	\\
            \toprule
        \end{tabular}
    }
    \label{tab:main_results_on_policy_same_vocab}
\end{table*}

\subsection{Baselines}
We compare our framework with existing methods under two settings:

\textbf{KD with the same vocabulary.} In this setting, we compare DSKD with the current white-box KD framework on the following divergences:
\begin{itemize}
    \item \textbf{KL.} The standard KL divergence used in KD is proposed by \cite{hinton15kd}.
    \item \textbf{RKL.} The reverse KL divergence swaps the two distributions in the KL divergence.
    \item \textbf{SKL.} The skewed KL proposed by \cite{ko24distillm}, which skews the student distribution $q_{\theta}$ in KL as $\lambda p + (1-\lambda)q_{\theta}$ (we set $\lambda=0.1$).
    \item \textbf{SRKL.} The skewed RKL proposed by \cite{ko24distillm}, which skews the teacher distribution $p$ in RKL as $\lambda q_{\theta} + (1 - \lambda) p$ (we set $\lambda=0.1$).
    \item \textbf{AKL.} The adaptive fusion of KL and RKL proposed by \cite{wu2024rethinking}.
\end{itemize}

\textbf{KD with different vocabularies.} We also compare DSKD with the ETA algorithm to the cross-tokenizer KD methods:
\begin{itemize}
    \item \textbf{MinCE.} The method proposed by \cite{wan24fusellm} aligns the logits between different models via dynamic programming that minimizes the edit distances of token strings.
    \item \textbf{ULD.} The method proposed by \cite{boizard2024uld}, which replaces the usual KL divergence with a closed-form solution of Wasserstein distance to overcome the limitation on the same tokenizers between the teacher and the student.
    \item \textbf{MultiLevelOT.} \cite{cui2024multilevelot} improves ULD by formulating cross-tokenizer KD as a multi-level optimal transportation problem and proposes MultiLevelOT to solve it.
\end{itemize}

Besides, we also compare our framework with the black-box KD method, \emph{i.e.}, sequence-level KD (\textbf{SeqKD}) \cite{kim16seqkd}, under both settings.
Moreover, we further evaluate on-policy DSKD on both settings and compare our framework with existing on-policy KD methods, such as GKD \cite{agarwal24gkd}, MiniLLM \cite{gu23minillm}, and DistiLLM \cite{ko24distillm}.
For a fair comparison, we reimplement these methods in the same setting as our method, which may be different from the original implementation and are listed as follows:
\begin{itemize}
    \item We mix the cross-entropy loss on the ground-truth outputs and the KD loss as the final training loss, while MiniLLM \cite{gu23minillm} and DistiLLM \cite{ko24distillm} did not use the cross-entropy loss;
    \item The temperature $\tau$ is set to 2, consistent with the off-policy setting.
    \item The pre-training corpus in MiniLLM and DistiLLM is not used as it only has a minor impact on the final performance.
    \item We did not warm up the student model with supervised fine-tuning since it is unnecessary according to \cite{ko24distillm}. 
\end{itemize}
% {\color{red}(Need more details.)}
% 修改：介绍具体工作细节

\begin{table*}[t]
\caption{The Rouge-L scores (\%) on several benchmarks on the same/different-vocabulary \textbf{off/on-policy} KD with GPT2-120M and TinyLLaMA-1.1B as students. ``+ *KL'' denotes our DSKD using the corresponding divergence as $\mathcal{D}(\cdot||\cdot)$ in Eq. (\ref{eq:stuside_kd_loss}) and (\ref{eq:teaside_kd_loss}). The result with $^{\dagger}$ marked represents it surpasses the performance of the same-vocabulary KD in Table \ref{tab:main_results_off_policy_same_vocab} and \ref{tab:main_results_on_policy_same_vocab}.}
    \centering
    \resizebox{0.9\linewidth}{!}{
        \begin{tabular}{llccccc|l}
            \bottomrule
            \multicolumn{2}{l}{{\textbf{Methods}}} & \textbf{Dolly} & \textbf{SelfInst} & \textbf{VicunaEval} & \textbf{S-NI} & \textbf{UnNI} & \textbf{Avg.} \\
            \hline
            \hline
            % \rowcolor{lightblue}
            % \rowcolor{lightgray}
            \multicolumn{8}{c}{\textbf{Qwen2-1.5B $\rightarrow$ GPT2-120M}} \\
            \hline
            \rowcolor{lightgray}
            % 这里重新测试教师模型的结果
            \multicolumn{2}{l}{Teacher} & 28.23$_{\pm 	0.41	}$ & 	20.76$_{\pm 	0.63	}$ & 	20.75$_{\pm 	0.44	}$ & 	37.32$_{\pm 	0.38	}$ & 	39.27$_{\pm 	0.16 	}$ & 	29.26 \\
            \hline
            \multicolumn{2}{l}{SFT} &  23.19$_{\pm 	0.32	}$ & 	10.09$_{\pm 	0.31	}$ & 	15.08$_{\pm 	0.62	}$ & 	17.25$_{\pm 	0.24	}$ & 	20.21$_{\pm 	0.13 	}$ & 	17.16 \\
            \multicolumn{2}{l}{SeqKD \cite{kim16seqkd}} &  23.22$_{\pm 	0.30	}$ & 	10.12$_{\pm 	0.57	}$ & 	15.35$_{\pm 	0.29	}$ & 	16.96$_{\pm 	0.16	}$ & 	19.71$_{\pm 	0.10 	}$ & 	17.07 \\
            \multicolumn{2}{l}{MinED \cite{wan24fusellm}} & 24.31$_{\pm 	0.34	}$ & 	10.79$_{\pm 	0.48	}$ & 	15.46$_{\pm 	0.55	}$ & 	19.77$_{\pm 	0.20	}$ & 	22.70$_{\pm 	0.17 	}$ & 	18.61		\\
            \multicolumn{2}{l}{ULD \cite{boizard2024uld}} \qquad &  23.59$_{\pm 	0.19	}$ & 	10.76$_{\pm 	0.29	}$ & 	15.44$_{\pm 	0.46	}$ & 	18.26$_{\pm 	0.10	}$ & 	20.90$_{\pm 	0.12 	}$ & 	17.79		\\
            \multicolumn{2}{l}{MultiLevelOT \cite{cui2024multilevelot}} \qquad &  24.16$_{\pm 	0.31	}$ & 	10.58$_{\pm 	0.39	}$ & 	14.90$_{\pm 	0.38	}$ & 	18.25$_{\pm 	0.31	}$ & 	22.06$_{\pm 	0.11 	}$ & 	17.99		\\
            \hline

            \multirow{5}{*}{\makecell{\textbf{Off-Policy} DSKD-ETA (ours)}} & + KL 	&	24.90$_{\pm 	0.29	}$ & 	11.63$_{\pm 	0.28	}$ & 	15.84$_{\pm 	0.38	}$ & 	21.01$_{\pm 	0.34	}$ & 	24.04$_{\pm 	0.13 	}$ & 	19.48 		\\ 
            & + RKL &	24.87$_{\pm 	0.35	}$ & 	12.27$_{\pm 	0.48	}$ & 	15.48$_{\pm 	0.22	}$ & 	22.53$_{\pm 	0.24	}$ & 	25.10$_{\pm 	0.15 	}$ & 	\textbf{20.05}		\\
            & + SKL &	25.12$_{\pm 	0.35	}$ & 	11.31$_{\pm 	0.57	}$ & 	15.24$_{\pm 	0.33	}$ & 	20.13$_{\pm 	0.20	}$ & 	22.94$_{\pm 	0.10 	}$ & 	18.95		\\
            & + SRKL & 25.05$_{\pm 	0.40	}$ & 	11.46$_{\pm 	0.26	}$ & 	15.71$_{\pm 	0.38	}$ & 	19.67$_{\pm 	0.23	}$ & 	23.49$_{\pm 	0.14 	}$ & 	19.08		\\
            & + AKL & 24.76$_{\pm 	0.24	}$ & 	10.72$_{\pm 	0.45	}$ & 	15.62$_{\pm 	0.40	}$ & 	21.54$_{\pm 	0.23	}$ & 	24.33$_{\pm 	0.15 	}$ & 	19.39		\\
            \hline
            % \textit{\textbf{On-Policy}} \\
            \multirow{5}{*}{\makecell{\textbf{On-Policy} DSKD-ETA (ours)}} & + KL &	24.98$_{\pm 	0.14	}$ & 	11.67$_{\pm 	0.46	}$ & 	16.82$_{\pm 	0.52	}$ & 	20.47$_{\pm 	0.26	}$ & 	22.01$_{\pm 	0.16 	}$ & 	19.19		\\
            & + RKL &	24.84$_{\pm 	0.31	}$ & 	12.64$_{\pm 	0.31	}$ & 	17.42$_{\pm 	0.48	}$ & 	19.43$_{\pm 	0.35	}$ & 	23.52$_{\pm 	0.08 	}$ & 	19.57		\\
            & + SKL &	25.29$_{\pm 	0.53	}$ & 	11.86$_{\pm 	0.39	}$ & 	17.03$_{\pm 	0.59	}$ & 	21.29$_{\pm 	0.30	}$ & 	25.58$_{\pm 	0.14 	}$ & 	20.21	\\
            & + SRKL &	26.12$_{\pm 	0.19 }$ & 	12.20$_{\pm 	0.36	}$ & 	16.69$_{\pm 	0.50	}$ & 	21.96$_{\pm 	0.34	}$ & 	25.61$_{\pm 	0.13 	}$ & \textbf{20.52} 	\\
            & + AKL &	25.62$_{\pm 	0.37	}$ & 	12.23$_{\pm 	0.61	}$ & 	16.64$_{\pm 	0.52	}$ & 	22.73$_{\pm 	0.30	}$ & 	24.39$_{\pm 	0.14 	}$ & 	20.32	\\
            \hline

            \multicolumn{8}{c}{\textbf{Mistral-7B $\rightarrow$ TinyLLaMA-1.1B}} \\
            \hline
            \rowcolor{lightgray}
            \multicolumn{2}{l}{Teacher} & 31.56$_{\pm 0.19}$ & 25.10$_{\pm 0.36}$ & 20.50$_{\pm 0.32}$ & 36.07$_{\pm 0.24}$ & 36.27$_{\pm 0.15}$ & 29.90 \\
            \hline
            \multicolumn{2}{l}{SFT} & 23.20$_{\pm 0.13}$ & 14.88$_{\pm 0.54}$ & 16.42$_{\pm 0.35}$ & 27.79$_{\pm 0.27}$ & 26.12$_{\pm 0.11}$ & 21.68 \\
            \multicolumn{2}{l}{SeqKD \cite{kim16seqkd}} & 23.56$_{\pm 0.39}$ & 15.87$_{\pm 0.54}$ & 15.99$_{\pm 0.55}$ & 25.50$_{\pm 0.37}$ & 26.64$_{\pm 0.09}$ & 21.51 \\
            \multicolumn{2}{l}{MinED \cite{wan24fusellm}} & 20.96$_{\pm 0.51}$ & 14.49$_{\pm 0.35}$ & 15.98$_{\pm 0.45}$ & 27.21$_{\pm 0.13}$ & 26.47$_{\pm 0.11}$ & 21.77 \\
            \multicolumn{2}{l}{ULD \cite{boizard2024uld}} \qquad & 22.80$_{\pm 0.28}$ & 15.93$_{\pm 0.74}$ & 16.43$_{\pm 0.60}$ & 26.94$_{\pm 0.28}$ & 24.83$_{\pm 0.13}$ & 20.64 \\
            % DSKD + Align (ours) & & & & & \\
            \multicolumn{2}{l}{MultiLevelOT \cite{cui2024multilevelot}} & 23.84$_{\pm 0.58}$ & 17.64$_{\pm 0.59}$ & 15.74$_{\pm 0.35}$ & 29.04$_{\pm 0.35}$ & 29.26$_{\pm 0.16}$ & 23.10  \\
            \hline
            % \textit{\textbf{Off-Policy}} \\
            \multirow{5}{*}{\makecell{\textbf{Off-Policy} DSKD-ETA (ours)}} & + KL	&	26.17$_{\pm 0.25}$ & 18.86$_{\pm 0.89}$ & 18.17$_{\pm 0.27}$ & 31.54$_{\pm 0.30}$ & 30.57$_{\pm 0.16}$ & 25.06  \\

            & + RKL & 26.91$_{\pm 0.17}$ & 20.02$_{\pm 0.36}$ & 18.75$_{\pm 0.38}$ & 33.71$_{\pm 0.20}$ & 34.41$_{\pm 0.06}$ & \textbf{26.76}$^{\dagger}$ \\

            & + SKL & 25.52$_{\pm 0.10}$ & 17.51$_{\pm 0.32}$ & 16.98$_{\pm 0.62}$ & 29.73$_{\pm 0.24}$ & 30.64$_{\pm 0.13}$ & 24.08 \\

            & + SRKL & 25.30$_{\pm 0.81}$ & 17.92$_{\pm 0.60}$ & 16.59$_{\pm 0.35}$ & 31.97$_{\pm 0.15}$ & 31.65$_{\pm 0.06}$ & 24.69$^{\dagger}$   \\

            & + AKL  & 27.16$_{\pm 0.31}$ & 19.38$_{\pm 0.73}$ & 18.59$_{\pm 0.22}$ & 34.95$_{\pm 0.13}$ & 33.59$_{\pm 0.10}$ & 26.73$^{\dagger}$   \\

            \hline
            % \textit{\textbf{On-Policy}} \\
            \multirow{5}{*}{\makecell{\textbf{On-Policy} DSKD-ETA (ours)}} & + KL	&	28.31$_{\pm 0.15}$ & 20.08$_{\pm 0.57}$ & 19.22$_{\pm 0.31}$ & 34.79$_{\pm 0.17}$ & 33.76$_{\pm 0.16}$ & 27.23$^{\dagger}$ \\
            % \hline
            & + RKL	&	27.87$_{\pm 0.30}$ & 18.29$_{\pm 0.65}$ & 18.72$_{\pm 0.21}$ & 34.41$_{\pm 0.11}$ & 36.16$_{\pm 0.07}$ & 27.09{\color{white}$^{\dagger}$} \\
            % \hline
            & + SKL &	27.95$_{\pm 0.43}$ & 19.57$_{\pm 0.31}$ & 18.59$_{\pm 0.36}$ & 33.89$_{\pm 0.24}$ & 34.86$_{\pm 0.15}$ & 26.97$^{\dagger}$ \\
            % \hline
            & + SRKL &	26.90$_{\pm 0.42}$ & 18.26$_{\pm 0.54}$ & 18.16$_{\pm 0.52}$ & 33.71$_{\pm 0.23}$ & 34.76$_{\pm 0.07}$ & 26.36$^{\dagger}$ \\
            % \hline
            & + AKL	&	28.10$_{\pm 0.28}$ & 18.72$_{\pm 0.24}$ & 19.46$_{\pm 0.40}$ & 35.82$_{\pm 0.19}$ & 36.26$_{\pm 0.17}$ & \textbf{27.67}$^{\dagger}$  \\
            \toprule
        \end{tabular}
    }
    \label{tab:main_results_dif_vocab}
\end{table*}

% \begin{figure}[ht]
%     \centering
%     \includegraphics[width=0.9\linewidth]{temperature.png}
%     \caption{Rouge-L scores (\%) on the validation set for different temperature coefficients in KL divergence and reverse KL divergence.}
%     \label{fig:temperature}
% \end{figure}

% \subsection{Effect of Temperature for KD} \label{sec:temperature}
% As an important hyper-parameter in KD, the temperature coefficient $\tau$ significantly affects the final performance of KD.
% As stated by the previous literature, a larger temperature ($>$1.0) will smooth the teacher's distribution and transfer more class relationship information to the student model. 
% Thus, we search for the best temperatures among [1.0, 1.5, 2.0, 3.0, 4.0] for two representative objectives (\emph{i.e.}, KL divergence and reverse KL divergence) on the validation set and report the results in Fig. \ref{fig:temperature}.
% The results show that both objectives perform best when the temperature is 2.0.
% Thus, we keep the temperature to 2.0 for all objectives in our experiments of Table \ref{tab:main_results_off_policy_same_vocab}, \ref{tab:main_results_on_policy_same_vocab} and \ref{tab:main_results_dif_vocab}.

\subsection{Main Results}

\textbf{Off-Policy KD with the same vocabulary.} 
The results of off-policy KD for models with the same vocabulary are reported in Table \ref{tab:main_results_off_policy_same_vocab}.
Firstly, it is shown that all white-box KD methods exhibit better performance than the black-box KD method SeqKD, which demonstrates that token-level distributions can transfer more knowledge than single target tokens.
Furthermore, our DSKD framework significantly and consistently outperforms the current white-box KD framework for both GPT2-120M and TinyLLaMA-1.1B on various divergences (please refer to the values of ``{\color{midgray2}(+*$\uparrow$)}'').
Particularly, our DSKD with RKL achieves the best improvement for both models.
On the one hand, it showcases the effectiveness of our DSKD framework that conducts KD in unified output spaces.
On the other hand, the improvements on all divergences also demonstrate that our framework is highly compatible with current divergences in white-box KD.

\textbf{On-Policy KD with the same vocabulary.} 
The results of on-policy KD for models with the same vocabulary are listed in Table \ref{tab:main_results_on_policy_same_vocab}.
Firstly, the on-policy DSKD consistently outperforms the off-policy DSKD (please refer to the values of ``{\color{newblue}(+*$\uparrow$)}''), demonstrating the superiority of the on-policy training strategy.
Additionally, our DSKD significantly exceeds the existing baseline methods, which further proves the potential of unifying the output space.
Particularly, the performance (marked by *) of our DSKD ``$+$ KL/RKL/SKL/AKL'' on the setting ``LLaMA2-7B $\rightarrow$ TinyLLaMA-1.1B'' even surpasses the performance of the teacher itself.

\textbf{KD with different vocabularies.} 
We list the results of off-policy and on-policy KD for LLMs with different vocabularies in Table \ref{tab:main_results_dif_vocab}.
As mentioned in Section \ref{sec:depend_same_vocab}, the key challenge in this setting is to deal with the unaligned distributions due to different vocabulary sizes and sequence tokenization.
Facing this challenge, existing KD methods only pre-define coarse alignment and thus yield limited performance, lagging behind KD methods for models with the same vocabulary.
In contrast, our ETA algorithm computes the exact token alignment, which facilitates our DSKD-ETA to perform better than existing baseline methods to a large extent.
Particularly, as the teacher models under this setting are stronger, our DSKD-ETA can sometimes achieve better performance than DSKD with the same vocabulary (please refer to the results marked by $^{\dagger}$), \emph{e.g.}, ``Off/On-Policy DSKD-ETA + SRKL/AKL''.
It suggests the potential of our method to train better students with stronger teachers even if they have different vocabularies, indicating the generalization and high applicability of our DSKD.
Moreover, on-policy DSKD-ETA further improves the performance of KD on LLMs with different vocabularies.

% \textbf{On-Policy KD.}
% The results of ``DSKD(-TA)-On-Policy'' present in Table \ref{tab:main_results_gpt2} and Table \ref{tab:main_results_tinyllama} show the effectiveness and superiority of on-policy knowledge distillation.
% Our DSKD-On-Policy outperforms other on-policy baselines (GKD, MiniLLM, and DistiLLM), which further proves the potential of unifying the output space.
% Particularly, the performance of ``KL/RKL/SKL/AKL \emph{w/} DSKD-On-Policy'' on the setting ``LLaMA2-7B $\rightarrow$ TinyLLaMA-1.1B'' even surpasses the performance of the teacher itself.

\begin{table*}[t]
\caption{The Ablation Study on ``GPT2-1.5B $\rightarrow$ GPT2-120M''. \textit{w/o} means without the corresponding settings. ``{\color{red}($-$*$\downarrow$)}'' represents the degraded performance compared with DSKD, and the bold denotes the largest decrease among three settings.}
    \centering
    \resizebox{0.9\linewidth}{!}{
        \begin{tabular}{lccccc|l}
            \bottomrule
            \textbf{Methods} & \textbf{Dolly} & \textbf{SelfInst} & \textbf{VicunaEval} & \textbf{S-NI} & \textbf{UnNI} & \quad  \quad  \textbf{Avg.} \\
            \hline
\rowcolor{lightgray}
KL	&	24.15$_{\pm 	0.32	}$ & 	10.69$_{\pm 	0.56	}$ & 	16.05$_{\pm 	0.50	}$ & 	19.29$_{\pm 	0.14	}$ & 	22.34$_{\pm 	0.07 	}$ & 	18.50				\\
\quad \emph{w/} DSKD	&	24.68$_{\pm 	0.58	}$ & 	10.98$_{\pm 	0.55	}$ & 	15.87$_{\pm 	0.28	}$ & 	21.73$_{\pm 	0.24	}$ & 	24.12$_{\pm 	0.11 	}$ & 	\textbf{19.48}				\\
\quad \quad \quad \emph{w/o} KD in Teacher Space		&	24.40$_{\pm 	0.20	}$ & 	11.32$_{\pm 	0.55	}$ & 	15.98$_{\pm 	0.31	}$ & 	18.96$_{\pm 	0.31	}$ & 	22.52$_{\pm 	0.12 	}$ & 	18.64	{\color{red}($-$\textbf{0.84}$\downarrow$)}	\\
\quad \quad \quad \emph{w/o} KD in Student Space	&	23.95$_{\pm 	0.10	}$ & 	11.18$_{\pm 	0.29	}$ & 	15.38$_{\pm 	0.29	}$ & 	21.20$_{\pm 	0.34	}$ & 	24.21$_{\pm 	0.10 	}$ & 	19.18	{\color{red}($-$0.29$\downarrow$)}	\\
\quad \quad \quad \emph{w/o} Initialize Projectors		&	24.81$_{\pm 	0.33	}$ & 	11.14$_{\pm 	0.28	}$ & 	15.87$_{\pm 	0.38	}$ & 	18.87$_{\pm 	0.29	}$ & 	23.41$_{\pm 	0.09 	}$ & 	18.82	{\color{red}($-$0.66$\downarrow$)}	\\
\hline																							\rowcolor{lightgray}		
RKL	&	24.41$_{\pm 	0.22	}$ & 	11.01$_{\pm 	0.40	}$ & 	15.16$_{\pm 	0.46	}$ & 	19.28$_{\pm 	0.32	}$ & 	22.56$_{\pm 	0.15 	}$ & 	18.48				\\
\quad \emph{w/} DSKD	&	25.19$_{\pm 	0.30	}$ & 	11.25$_{\pm 	0.36	}$ & 	15.90$_{\pm 	0.44	}$ & 	23.97$_{\pm 	0.20	}$ & 	25.17$_{\pm 	0.08 	}$ & 	\textbf{20.30}				\\
\quad \quad \quad \emph{w/o} KD in Teacher Space	&	24.87$_{\pm 	0.21	}$ & 	11.24$_{\pm 	0.52	}$ & 	15.24$_{\pm 	0.38	}$ & 	20.00$_{\pm 	0.14	}$ & 	22.53$_{\pm 	0.09 	}$ & 	18.78	{\color{red}($-$\textbf{1.52}$\downarrow$)}	\\
\quad \quad \quad \emph{w/o} KD in Student Space	&	23.77$_{\pm 	0.27	}$ & 	10.79$_{\pm 	0.32	}$ & 	15.95$_{\pm 	0.13	}$ & 	20.66$_{\pm 	0.09	}$ & 	23.20$_{\pm 	0.09 	}$ & 	18.87	{\color{red}($-$1.42$\downarrow$)}	\\
\quad \quad \quad \emph{w/o} Initialize Projectors	&	24.88$_{\pm 	0.42	}$ & 	10.70$_{\pm 	0.48	}$ & 	15.58$_{\pm 	0.29	}$ & 	21.27$_{\pm 	0.22	}$ & 	23.63$_{\pm 	0.05 	}$ & 	19.21	{\color{red}($-$1.08$\downarrow$)}	\\
\hline																							\rowcolor{lightgray}			
SKL	&	23.66$_{\pm 	0.24	}$ & 	11.62$_{\pm 	0.60	}$ & 	15.17$_{\pm 	0.19	}$ & 	20.37$_{\pm 	0.21	}$ & 	22.27$_{\pm 	0.23 	}$ & 	18.62				\\
\quad \emph{w/} DSKD	&	24.78$_{\pm 	0.26	}$ & 	12.10$_{\pm 	0.33	}$ & 	15.45$_{\pm 	0.36	}$ & 	21.25$_{\pm 	0.11	}$ & 	24.58$_{\pm 	0.11 	}$ & 	\textbf{19.63}				\\
\quad \quad \quad \emph{w/o} KD in Teacher Space	&	24.25$_{\pm 	0.29	}$ & 	11.38$_{\pm 	0.28	}$ & 	15.77$_{\pm 	0.38	}$ & 	18.95$_{\pm 	0.21	}$ & 	22.08$_{\pm 	0.13 	}$ & 	18.49	{\color{red}($-$\textbf{1.15}$\downarrow$)}	\\
\quad \quad \quad \emph{w/o} KD in Student Space	&	23.58$_{\pm 	0.17	}$ & 	10.67$_{\pm 	0.41	}$ & 	15.35$_{\pm 	0.41	}$ & 	20.32$_{\pm 	0.23	}$ & 	23.36$_{\pm 	0.16 	}$ & 	18.66	{\color{red}($-$0.98$\downarrow$)}	\\
\quad \quad \quad \emph{w/o} Initialize Projectors	&	24.17$_{\pm 	0.42	}$ & 	10.88$_{\pm 	0.32	}$ & 	15.39$_{\pm 	0.41	}$ & 	19.97$_{\pm 	0.30	}$ & 	22.76$_{\pm 	0.14 	}$ & 	18.63	{\color{red}($-$1.00$\downarrow$)}	\\
\hline																							\rowcolor{lightgray}			
SRKL	&	24.19$_{\pm 	0.43	}$ & 	10.71$_{\pm 	0.39	}$ & 	15.38$_{\pm 	0.43	}$ & 	19.18$_{\pm 	0.24	}$ & 	22.27$_{\pm 	0.04 	}$ & 	18.35				\\
\quad \emph{w/} DSKD	&	24.90$_{\pm 	0.45	}$ & 	11.13$_{\pm 	0.25	}$ & 	15.49$_{\pm 	0.17	}$ & 	22.15$_{\pm 	0.26	}$ & 	24.38$_{\pm 	0.06 	}$ & 	\textbf{19.61}				\\
\quad \quad \quad \emph{w/o} KD in Teacher Space	&	24.70$_{\pm 	0.32	}$ & 	10.66$_{\pm 	0.35	}$ & 	15.44$_{\pm 	0.50	}$ & 	19.39$_{\pm 	0.19	}$ & 	22.61$_{\pm 	0.09 	}$ & 	18.56	{\color{red}($-$\textbf{1.05}$\downarrow$)}	\\
\quad \quad \quad \emph{w/o} KD in Student Space	&	23.85$_{\pm 	0.38	}$ & 	11.29$_{\pm 	0.42	}$ & 	15.53$_{\pm 	0.56	}$ & 	21.59$_{\pm 	0.11	}$ & 	23.16$_{\pm 	0.07 	}$ & 	19.08	{\color{red}($-$0.53$\downarrow$)}	\\
\quad \quad \quad \emph{w/o} Initialize Projectors	&	24.88$_{\pm 	0.26	}$ & 	11.61$_{\pm 	0.32	}$ & 	15.54$_{\pm 	0.36	}$ & 	19.97$_{\pm 	0.33	}$ & 	22.90$_{\pm 	0.13 	}$ & 	18.98	{\color{red}($-$0.63$\downarrow$)}	\\
\hline																							\rowcolor{lightgray}		
AKL	&	24.29$_{\pm 	0.35	}$ & 	11.26$_{\pm 	0.41	}$ & 	15.82$_{\pm 	0.56	}$ & 	19.29$_{\pm 	0.29	}$ & 	22.36$_{\pm 	0.15 	}$ & 	18.60				\\
\quad \emph{w/} DSKD	&	25.30$_{\pm 	0.33	}$ & 	11.91$_{\pm 	0.23	}$ & 	16.06$_{\pm 	0.52	}$ & 	22.67$_{\pm 	0.23	}$ & 	25.44$_{\pm 	0.06 	}$ & 	\textbf{20.28}				\\
\quad \quad \quad \emph{w/o} KD in Teacher Space	&	24.36$_{\pm 	0.50	}$ & 	11.13$_{\pm 	0.31	}$ & 	15.37$_{\pm 	0.25	}$ & 	19.66$_{\pm 	0.31	}$ & 	22.93$_{\pm 	0.08 	}$ & 	18.69	{\color{red}($-$\textbf{1.59}$\downarrow$)}	\\
\quad \quad \quad \emph{w/o} KD in Student Space	&	23.79$_{\pm 	0.52	}$ & 	11.06$_{\pm 	0.31	}$ & 	15.85$_{\pm 	0.25	}$ & 	21.27$_{\pm 	0.22	}$ & 	23.20$_{\pm 	0.13 	}$ & 	19.03	{\color{red}($-$1.24$\downarrow$)}	\\
\quad \quad \quad \emph{w/o} Initialize Projectors	&	25.03$_{\pm 	0.25	}$ & 	11.19$_{\pm 	0.46	}$ & 	16.25$_{\pm 	0.31	}$ & 	22.07$_{\pm 	0.31	}$ & 	24.96$_{\pm 	0.11 	}$ & 	19.90	{\color{red}($-$0.38$\downarrow$)}	\\
            \toprule
        \end{tabular}
    }
    \label{tab:ablation_results_gpt2}
\end{table*}

\begin{figure}[t]
    \centering
    \includegraphics[width=\linewidth]{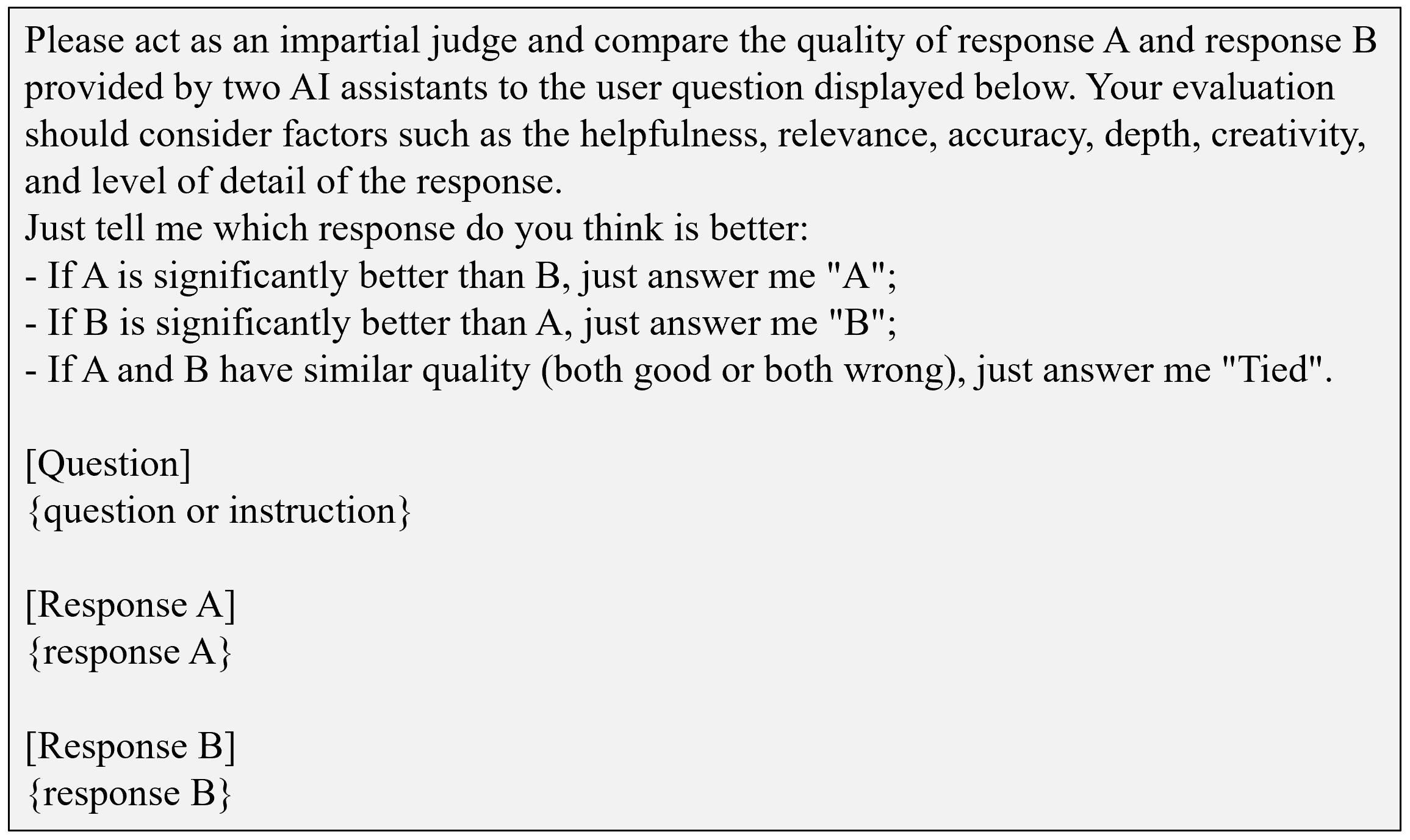}
    \caption{Prompt for GPT-4 Evaluation.}
    \label{fig:prompt}
\end{figure}

\begin{figure}[t]
    \centering
    \includegraphics[width=\linewidth]{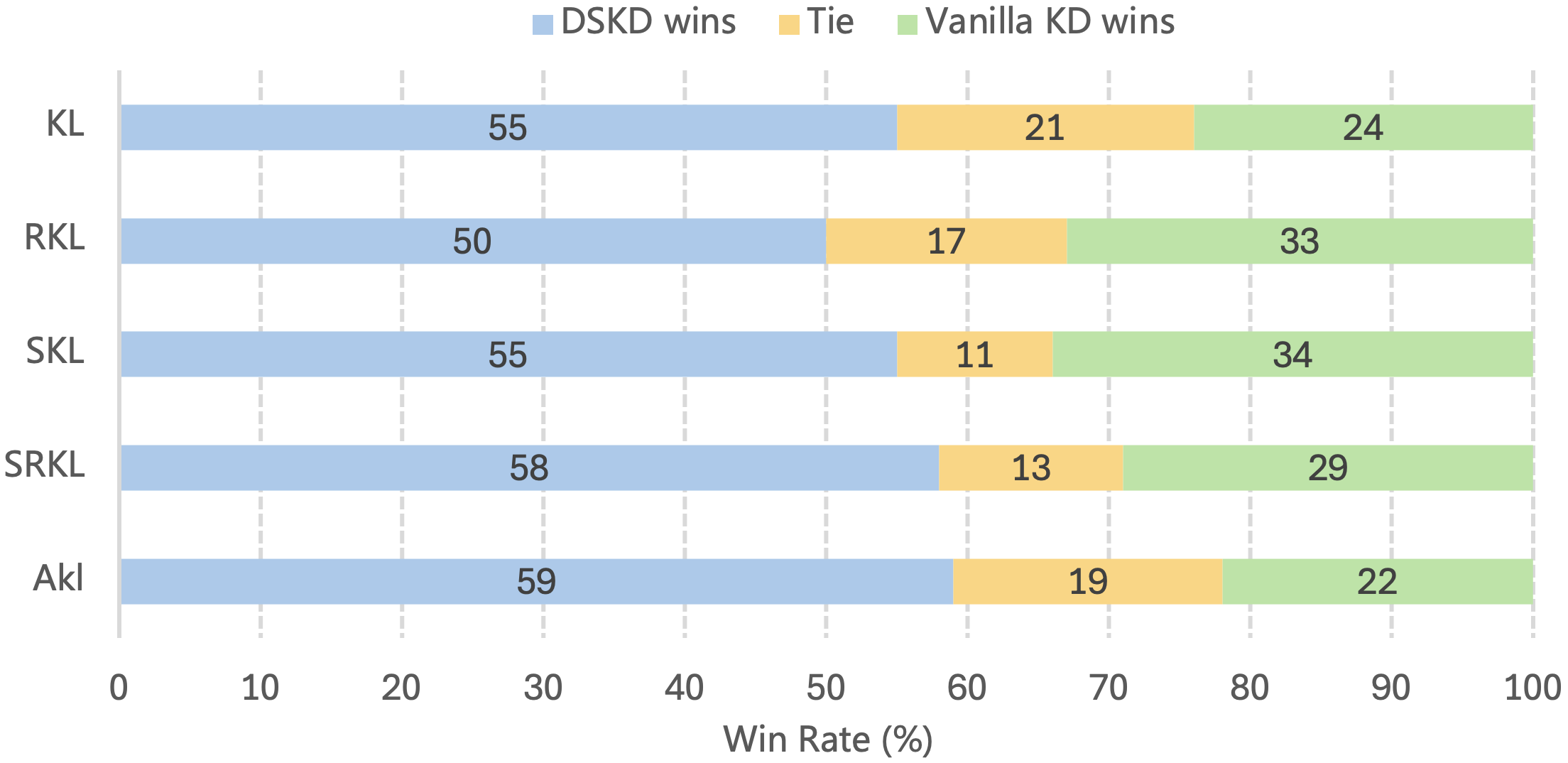}
    \caption{GPT-4 Evaluation Results for off-policy DSKD with all the divergences on the setting ``LLaMA2-7B $\rightarrow$ TinyLLaMA-1.1B''.}
    \label{fig:llm_eval_all}
\end{figure}

\subsection{Evaluation via GPT-4}
\label{sec:llm_eval}
We also employ GPT-4 evaluation to compare our DSKD with the vanilla white-box KD framework.
Specifically, we randomly pick 100 instructions in the test set of Dolly and generate responses with TinyLLaMA trained by DSKD and the vanilla framework.
Then we use the API of \texttt{gpt-4o-2024-08-06} to judge which responses are better with the evaluation prompt shown in Fig. \ref{fig:prompt}.
As we conduct a pairwise comparison between the responses from two models, to alleviate the position bias \cite{zheng2023judging} in the evaluation process of GPT-4, we randomly shuffle the two responses as Response A/B in the system prompts.
The win rates plotted in Fig. \ref{fig:llm_eval_all} show that the students trained by our DSKD always win more than the students trained by the vanilla white-box KD framework, indicating the consistent superiority of our DSKD framework on existing divergences.

\section{Analysis}
% \vspace{-10pt}

\subsection{Ablation Study of DSKD}\label{sec:ablation-study}
% \subsection{KD in Different Spaces \emph{vs.} Unified Space}
In this section, we conduct the ablation study to evaluate the individual effect of KD in the teacher/student space and the initialization of the projectors.
Specifically, we conduct experiments under the following three settings.
\textbf{1) \textit{w/o} KD in Teacher Space}: Only calculate the losses in Student Space, \textit{i.e.},  $\mathcal{L}_{dskd}=\mathcal{L}^{stu}_{kd} + \mathcal{L}^{t \rightarrow s}_{ce}$;
\textbf{2) \textit{w/o} KD in Student Space}: Only calculate the losses in Teacher Space, \textit{i.e.},  $\mathcal{L}_{dskd}=\mathcal{L}^{tea}_{kd}$; 
\textbf{3) \textit{w/o} Initialize Projectors}: Do not initialize the projectors $\bm{W}^{t \rightarrow s}$ and $\bm{W}^{s \rightarrow t}$ following $\bm{W}^{t \rightarrow s}=\bm{W}^t{\bm{W}^s}^{+}$ and $\bm{W}^{s \rightarrow t}=\bm{W}^{s}{\bm{W}^{t}}^{+}$, \textit{i.e.}, just randomly initialize the two projectors.
The results listed in Table \ref{tab:ablation_results_gpt2} show that all three settings lead to a performance decrease compared to DSKD, demonstrating the necessity of each part in our DSKD framework.
Particularly, ``\textit{w/o} KD in Teacher Space'' results in the largest performance drop since this process has less transformation error and benefits from the original teacher distribution.
% proving the importance of the teacher-side loss and further suggests that KD in the Teacher Space can be equivalent to optimizing the original distribution of the students (\textit{i.e.}, the proposition \ref{prop:tea=stu}).
``\textit{w/o} Initialize Projectors'' also leads to a substantial performance drop, which indicates the rationality and necessity of the initialization method.
% Nevertheless, even without projector initialization, our DSKD still exceeds vanilla KD under all divergences.
Additionally, most of the results in the three settings can still exceed the ones of the original baseline (\textit{e.g.}, KL, RKL, SRKL, and AKL), which further sufficiently demonstrates the superiority of unifying the output spaces of the distributions for KD.

% \subsection{Ablation Study of DSKD-ETA}
% In the above section, we have proved the effectiveness of each part in our DSKD framework.

\subsection{Representation Similarity between the Teacher and the Student}
In the simulation experiment, we find that the vanilla KD framework will lead to limited representation similarities between the student and the teacher (as shown in Fig. \ref{fig:kl_simulation}(b) and \ref{fig:kl_simulation}(f)). 
Thus, we evaluate whether this phenomenon also holds in the real KD scenario.
% Specifically, we use cosine similarity and the normalized inner product between output hidden states to represent a model's representation structure.
% The detailed calculation of the structure distance is as follows:
Since the student models and the teacher models generally have different dimensions on representations, it is difficult to directly measure the representation similarity between the student and the teacher.
To address this issue, we calculate the similarity of the representation structures given the same sentence between the student and the teacher.
Specifically, given a sentence with $n$ tokens, we calculate structure matrices with both the cosine similarity and normalized inner-product values between the output hidden states of this sentence:
\begin{equation}
    \mathcal{M}_{cosine}(i, j)=\frac{{h_i}^\top h_j}{|h_i| |h_j|} \in \mathbb{R}^{n \times n},
\end{equation}
\begin{equation}
    \mathcal{M}_{prod}(i, j)=\frac{{h_i}^{\top} h_j}{\sum_k {h_i}^{\top} h_k} \in \mathbb{R}^{n \times n},
\end{equation}
where $\mathcal{M}_{cosine}$ and $\mathcal{M}_{prod}$ are structure matrices calculated by cosine and normalized inner-product between output hidden states, respectively.
Then we calculate the L1 distance between the matrices of the student and the teacher:
\begin{equation}
    \mathcal{D}_{cosine}=\sum_i^n\sum_j^n |\mathcal{M}^t_{cosine}(i, j) - \mathcal{M}^s_{cosine}(i, j)|,
\end{equation}
\begin{equation}
    \mathcal{D}_{prod}=\sum_i^n\sum_j^n |\mathcal{M}^t_{prod}(i, j) - \mathcal{M}^s_{prod}(i, j)|.
\end{equation}
The smaller distance values denote that the representations of the student and the teacher are more similar.
% In Fig. \ref{fig:repr_sim}, we calculate and average the two distances $\mathcal{D}_{cosine}$ and $\mathcal{D}_{prod}$ on 1000 samples in the training set for GPT2 models that trained without KD (SFT), trained by the current white-box KD framework (Vanilla KD) and trained by our DSKD framework (DSKD).

\begin{figure}[t]
	\centering
	\subfigure[Cosine as Structure]{
		\begin{minipage}[t]{0.49\linewidth}
			\centering
			\includegraphics[width=\linewidth]{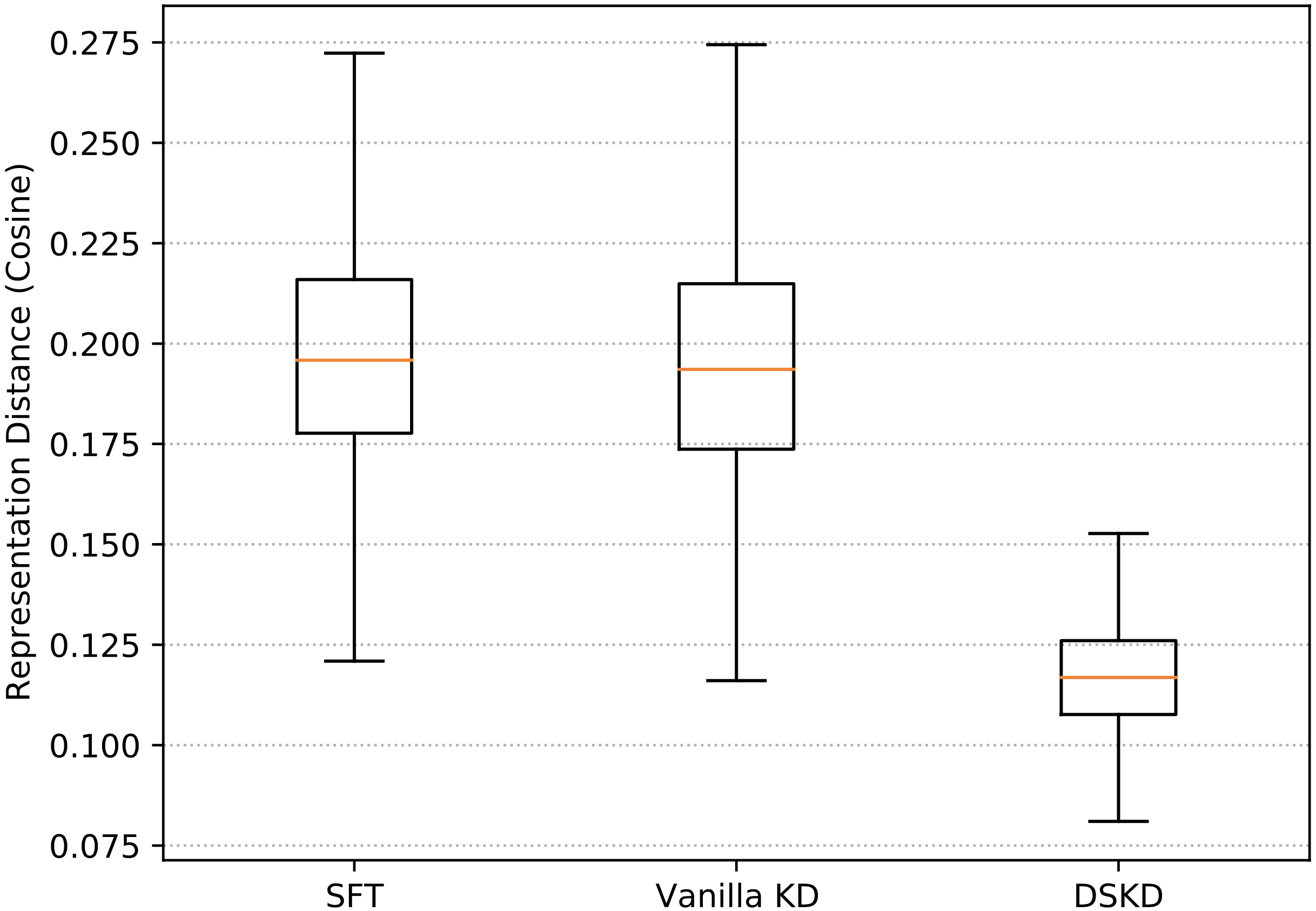}
		\end{minipage}
	}%
	\subfigure[Inner Product as Structure]{
		\begin{minipage}[t]{0.46\linewidth}
			\centering
			\includegraphics[width=\linewidth]{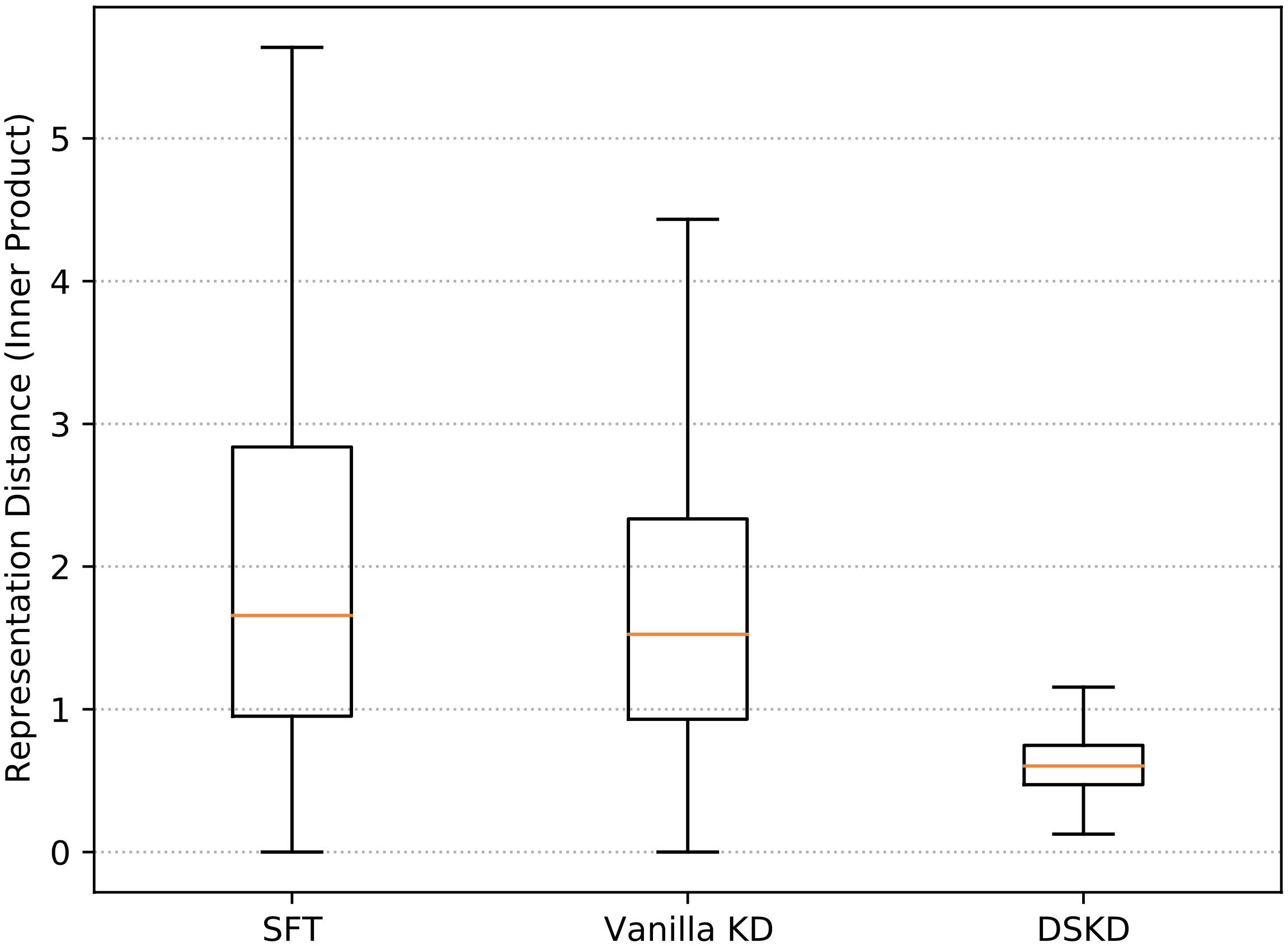}
		\end{minipage}
	}%
	\centering
	\caption{Distance between the representation structures of the teacher and the student.}
	\label{fig:repr_sim}
\end{figure}

We plot the average distance between the representation structures on 1000 training samples in Fig. \ref{fig:repr_sim}.
It shows that on both types of representation structures, the current KD framework (\textbf{Vanilla KD}) only reduces minor distances between the teacher and the student compared to fine-tuning without KD (\textbf{SFT}).
However, our \textbf{DSKD} achieves significantly lower distances between the teacher and the student, which demonstrates that DSKD can enhance the representation similarity between the student and the teacher.

\subsection{Computational efficiency}

% 加了多少参数 / 训练速度 / 推理速度 
In this section, we report the actual computational time for each epoch and the GPU memory during KD.
For GPT2-120M, we set batch size=2, gradient accumulation=4, and num-gpus=8. 
Under the settings GPT2-1.5B $\rightarrow$ GPT2-120M and Qwen2-1.5B $\rightarrow$ GPT2-120M, the proposed projectors add only 2.46M and 2.36M parameters than vanilla KD.
For TinyLLaMA-1.1B, we set batch size=4, gradient accumulation=1, and num-gpus=8 with gradient-checkpointing.
Under the settings LLaMA2-7B $\rightarrow$ TinyLLaMA-1.1B and Mistral-7B $\rightarrow$ TinyLLaMA-1.1B, the proposed projectors add only 16.78M  parameters than vanilla KD.
Then, the seconds per epoch and the GPU memory are listed in Table \ref{tab:gpu}.
It shows that as the language model grows larger, the additional training cost becomes more negligible.
Compared to DSKD, DSKD-ETA will cost more training time due to the ETA algorithm.
Nevertheless, the inference speed of student models after distillation is unchanged.

\begin{table}[ht]
\caption{The seconds per epoch and the GPU memory of different methods for GPT2-120M and TinyLLaMA-1.1B.}
    \centering
    \resizebox{\linewidth}{!}{
    \begin{tabular}{l|cc}
        \bottomrule
        Methods & Seconds / Epoch & GPU memory (MiB) \\
        
        \toprule
        \multicolumn{3}{c}{GPT2-120M} \\
        \bottomrule
        Vanilla KD & 42 & 14189 \\
        DSKD (ours) & 56 & 16673 \\
        DSKD-ETA (ours) & 130 & 22425 \\
        
        \toprule
        \multicolumn{3}{c}{TinyLLaMA-1.1B} \\
        \bottomrule
        Vanilla KD & 115 & 24549 \\
        DSKD (ours) & 122 & 26307 \\
        DSKD-ETA (ours) & 212 & 25943 \\
        
        \toprule
    \end{tabular}
    }
    \label{tab:gpu}
\end{table}

% vanilla KD
% DSKD
% DSKD-ETA

\subsection{Evaluation on Other Benchmarks}
In this section, we evaluate the effectiveness of our method on other typical capacities, including general instruction following, mathematical reasoning, and code generation.

\begin{table}[ht]
\caption{Performance of length-controlled win rate (LC) and win rate (WR) on instruction-following benchmarks AlpacaEval and Evol-Instruct. The baseline is \texttt{text-davinci-003} in AlpacaEval and \texttt{gpt3.5-turbo} in Evol-Instruct. ``KL'' is the divergence of Vanilla KD and DSKD. The \textbf{bold} denotes the best result.}
    \centering
    \resizebox{\linewidth}{!}{
    \begin{tabular}{l|cc|c|c}
        \toprule
        \multirow{2}{*}{\textbf{Methods}} & \multicolumn{2}{c|}{\textbf{AlpacaEval}} & \textbf{Evol-Instruct} & \multirow{2}{*}{\textbf{Avg.}} \\
        \cmidrule(lr){2-3}\cmidrule(lr){4-4}
        & LC (\%) & WR (\%) & WR (\%) & \\
        \midrule
        % \multicolumn{4}{c}{\textit{Teacher Model}} \\
        % \midrule
        
        Teacher & 92.55 & 97.83 & 88.30 & 92.89 \\
        
        \midrule
        \multicolumn{5}{c}{\textbf{Qwen2.5-7B-Instruct $\rightarrow$ Qwen2.5-1.5B}} \\
        \midrule
        SFT & 72.16 & 82.17 & 18.12 & 57.48 \\
        SeqKD & 80.14 & 89.13 & 54.73 & 74.67 \\
        \midrule
        % Vanilla KD & 79.22 & 89.07 & 48.17 & 72.15\\
        
        Vanilla KD & 82.69	& {89.75}	& 56.03 &	76.16 \\
        % Off-Policy DSKD & \textbf{84.45} & \textbf{89.50} & \textbf{54.82} & \textbf{76.26} \\
        Off-Policy DSKD & {85.29} & 89.63 & {62.84} & \textbf{79.25} \\
        \midrule
        \multicolumn{5}{c}{\textbf{Qwen2.5-7B-Instruct $\rightarrow$ Llama-3.2-1B}} \\
        \midrule
        SFT & 53.34 & 62.48 & 3.21 & 39.68 \\
        % SeqKD & 70.21 & 82.96 & 33.49 & 62.22 \\
        SeqKD & 69.32 & 78.48 & 30.49 & 59.43 \\
        
        \midrule
        Off-Policy DSKD-ETA & 72.52 & 78.79 & 29.36 & \textbf{60.22} \\
        \bottomrule
    \end{tabular}
    }
    \label{tab:instruction_evaluation}
\end{table}

\subsubsection{General Instruction Following}

\paragraph{Experimental Setup}
We randomly sample 50k examples from UltraChat200k \cite{ding-etal-2023-enhancing} as the training dataset.
We evaluate the performance on the common instruction following benchmarks AlpacaEval \cite{alpaca_eval} and Evol-Instruct \cite{xu2024wizardlm}.
For evaluation, we adopt LLM-as-a-Judge \cite{zheng2023judgingllmasajudgemtbenchchatbot} with Qwen2.5-72B-Instruct \cite{qwen2.5} as the judge model.
For distillation, we utilize Qwen2.5-7B-Instruct as the teacher and Qwen2.5-1.5B as the student with the same vocabulary, while Llama3.2-1B as the student with the different vocabulary.
For ``SeqKD'', the train data is generated by the teacher model with the temperature 0.7 and top-p 0.8.
Additionally, for ``Vanilla KD'', ``Off-Policy DSKD'', and ``Off-Policy DSKD-ETA'', we use the same data with ``SeqKD'' as the generated data by the teacher has higher quality than the original data for ``SFT''.
The experimental details and settings are introduced in Appendix \ref{sec:appendix-other-details}.

\paragraph{Results}
As shown in Table \ref{tab:instruction_evaluation}, our DSKD achieves the best performance on both settings.
Specifically, SeqKD significantly outperforms the SFT baseline, which indicates that the responses generated from the teacher model have a higher quality than the original training data.
Moreover, vanilla KD surpasses SeqKD since it transfers more information from the teacher model,
Notably, our DSKD further exceeds vanilla KD, demonstrating the effectiveness of our method.

\begin{table}[t]
\caption{Performance of Accuracy on Math-500 and GSM benchmarks. ``KL'' is the divergence of Vanilla KD and DSKD. The \textbf{bold} denotes the best average result.}
    \centering
    \resizebox{\linewidth}{!}{
    \begin{tabular}{l|c|c|c}
        \toprule
        \multirow{2}{*}{\textbf{Methods}} & \multicolumn{1}{c|}{\textbf{Math-500}} & \textbf{GSM} & \multirow{2}{*}{\textbf{Avg.}} \\
        \cmidrule(lr){2-2}\cmidrule(lr){3-3}
     & Accuracy (\%) & Accuracy (\%) & \\
        \midrule
        % \multicolumn{4}{c}{\textit{Teacher Model}} \\
        % \midrule
        Teacher & 55.40 & 86.20 & 70.80 \\
        
        \midrule
        \multicolumn{4}{c}{\textbf{Qwen2.5-Math-7B $\rightarrow$ Qwen2.5-1.5B}} \\
        \midrule
        SFT & 38.20 & 70.13 & 54.17 \\
        SeqKD & 36.60 & 70.58 & 53.59 \\
        \midrule

        Vanilla KD & 38.40 & 70.20 & 54.30 \\
        Off-Policy DSKD & {39.40} & {71.04} & \textbf{55.22} \\
        \midrule
        \multicolumn{4}{c}{\textbf{Qwen2.5-Math-7B $\rightarrow$ Llama-3.2-1B}} \\
        \midrule
        SFT & 7.80 & 42.38 & 25.09 \\
        SeqKD & 9.20 & 39.42 & 24.31 \\
        \midrule
        
        Off-Policy DSKD-ETA & {9.60} & {43.21} & \textbf{26.41} \\
        \bottomrule
    \end{tabular}
    }
    \label{tab:math_evaluation}
\end{table}
\subsubsection{Mathematical Reasoning}
\paragraph{Experimental Setup}
We randomly sample 50k examples from MetaMathQA \cite{yu2024metamathbootstrapmathematicalquestions} as the training dataset.
We evaluate the performance on the common mathematical reasoning benchmarks MATH-500 \cite{lightman2023lets} and GSM \cite{cobbe2021gsm8k}.
The evaluation tools are based on the Qwen2.5-Math project\footnote{\url{https://github.com/QwenLM/Qwen2.5-Math}}.
For experiments, we first fine-tune Qwen2.5-Math-7B as the teacher, Qwen2.5-1.5B as the student with the same vocabulary, and Llama3.2-1B as the student with the different vocabulary.
For ``SeqKD'', the training data is generated by the teacher model with the temperature 0.0.

\paragraph{Results}
The results of mathematical reasoning are listed in Table \ref{tab:math_evaluation}.
Unlike the observation in general instruction following, SeqKD exhibits inferior performance compared to SFT since there may be errors in teacher-generated data.
Besides, vanilla KD only brings little improvement over SFT.
By contrast, our DSKD showcases more significant improvement and outperforms other baselines for both settings.

% \textbf{Results.}
% The results listed in Table \ref{tab:math_evaluation} show that xxx.

\subsubsection{Code Generation}
\paragraph{Experimental Setup}
We randomly sample 10k examples from the Python subset of Magicoder \cite{wei2024magicoderempoweringcodegeneration} as the training dataset.
We evaluate the performance on the common code generation benchmarks HumanEval \cite{chen2021codex} and MBPP \cite{austin2021programsynthesislargelanguage}.
The evaluation tools are based on the Qwen2.5-coder project\footnote{\url{https://github.com/QwenLM/Qwen2.5-coder}}.
For experiments, we first fine-tune Qwen2.5-Coder-7B as the teacher, Qwen2.5-1.5B as the student with the same vocabulary, and Llama3.2-1B as the student with the different vocabulary.
For ``SeqKD'', the train data is generated by the teacher model with the temperature 0.0 and repetition-penalty 1.05.
Additionally, for ``Vanilla KD'', ``Off-Policy DSKD'', and ``Off-Policy DSKD-ETA'', we use the same data with ``SeqKD'' as the generated data by the teacher has higher quality than the original data for ``SFT''.

% \textbf{Results.}
% The results listed in Table \ref{tab:code_evaluation} show that xxx.
% \textbf{Results.}
\paragraph{Results}
The evaluation results are listed in Table \ref{tab:code_evaluation}.
For LLMs with the same vocabulary, our DSKD significantly exceeds SFT, Vanilla KD, and SeqKD.
And for LLMs with different vocabularies, our DSKD-ETA also surpasses SFT and SeqKD.

These evaluation results further prove the superiority and versatility of DSKD on various tasks.

\subsection{Case Study}
In Table \ref{tab:case_study}, we list the responses of `Vanilla KD`'' and our ``Off-Policy DSKD'' on one question after the KD setting ``Qwen2.5-Math-7B $\rightarrow$ Qwen2.5-1.5B'' and ``LLaMA2-7B $\rightarrow$ TinyLLaMA-1.1B''.
For ``Question-1'', after Vanilla KD, the student model does not understand the question well, leading to the wrong reasoning steps and final answer.
By contrast, the student trained by our DSKD can generate the correct reasoning steps and final answer.
For ``Question-2'', the answer of our DSKD is more detailed and comprehensive than Vanilla KD.
These two high-quality responses further showcase the superiority of DSKD over vanilla KD.

\begin{table}[t]
\caption{Performance of pass@1 on HumanEval and MBPP benchmarks. ``KL'' is the divergence of Vanilla KD and DSKD. The \textbf{bold} denotes the best average result.}
    \centering
    \resizebox{\linewidth}{!}{
    \begin{tabular}{l|c|c|c}
        \toprule
        \multirow{2}{*}{\textbf{Methods}} & \multicolumn{1}{c|}{\textbf{HumanEval}} & \textbf{MBPP} & \multirow{2}{*}{\textbf{Avg.}} \\
        \cmidrule(lr){2-2}\cmidrule(lr){3-3}
     & pass@1 (\%) & pass@1 (\%) & \\
        \midrule
        % \multicolumn{4}{c}{\textit{Teacher Model}} \\
        % \midrule
        Teacher & 62.20 & 59.65 & 60.93 \\
        \midrule
        \multicolumn{4}{c}{\textbf{Qwen2.5-Coder-7B $\rightarrow$ Qwen2.5-1.5B}} \\
        \midrule
        
        SFT & 43.29 & 33.33 & 38.31 \\
        SeqKD & 44.51 & 33.83 & 39.17 \\
        \midrule
        Vanilla KD & 43.29 & 36.59 & 39.94 \\
        
        Off-Policy DSKD & 45.73 & 37.59 & \textbf{41.86} \\
        \midrule
        \multicolumn{4}{c}{\textbf{Qwen2.5-Coder-7B $\rightarrow$ Llama-3.2-1B}} \\
        \midrule
        SFT & 14.63 & 15.54 & 15.09 \\
        SeqKD & 14.02 & 16.54 & 15.28 \\
        \midrule
        
        Off-Policy DSKD-ETA & 10.98 & 20.30 & \textbf{15.64} \\
        \bottomrule
    \end{tabular}
    }
    \label{tab:code_evaluation}
\end{table}

\section{Related Work}
\subsection{White-Box KD for Language Models}
The white-box KD framework for language models stems from the standard KD method proposed by \cite{hinton15kd}, which trains the student model to mimic the teacher model on the output probability distributions (\emph{a.k.a.}, logit-based KD).
In the field of NLP, numerous KD methods were designed following this framework to compress the excessive model sizes of pre-trained language models \cite{sun19patientkd,sanh19distilbert,sun20mobilebert,jiao20tinybert} or powerful yet cumbersome models in specific tasks \cite{tan19mnmtkd,chen19bertkd4textgen,zhang2023tiekd}.
Besides minimizing the distance between distributions, feature-based KD methods are also widely developed to transfer the fine-grained knowledge in intermediate hidden states and attention maps of the teacher model to the student model \cite{jiao20tinybert,wang20minilm,wang21minilmv2}.
% Also, white-box KD is widely used in text generation tasks, such as neural machine translation \cite{tan19mnmtkd}, text abstraction\cite{chen19bertkd4textgen}, and so on.
Recently, since LLMs are becoming predominant for different scenarios, several on-policy white-box KD techniques have also been proposed for LLMs \cite{lin20imitkd,wen23fdiv,gu23minillm,agarwal24gkd,ko24distillm,wu2024rethinking,zhang2024dskd,zhang2025aligndistil,xu24llmkdsurvey}.
Specifically, ImitKD \cite{lin20imitkd}, f-distil \cite{wen23fdiv}, and GKD \cite{agarwal24gkd} apply sequence-level KD for LM, where KD is conducted on the texts sampled from models rather than ground-truth texts.
This process helps mitigate the exposure bias caused by train-inference mismatch in traditional supervised KD methods and often achieves better performance \cite{agarwal24gkd}.
Furthermore, MiniLLM \cite{gu23minillm} converts the RKL minimization in sequence-level KD into a reinforcement learning problem and develops a policy gradient-based algorithm to optimize the sequence-level RKL.
Although effective, these on-policy KD algorithms involve sampling responses from the student model or the teacher model during training, which introduces a non-negligible computational cost.
To solve this, DistiLLM \cite{ko24distillm} designs an adaptive off-policy approach by reusing the model-generated text in the replay buffer.

However, existing KD methods for LLMs still follow the current white-box KD framework, which we reveal has inherent limitations on teacher-student similarity and the vocabulary requirement.
Thus, in this work, we directly challenge this framework and propose a more effective and general dual-space KD framework as the solution.

It is worth noting that \cite{chen2022reuse_classifier} also explores sharing the teacher's head during KD.
However, it keeps the teacher's head for the student model after KD and increases the inference cost of the distilled model.
On the contrary, our DSKD has a dual KD process and does not affect the inference speed of the distilled model.

\begin{table*}[t]
\caption{The responses of `Vanilla KD`'' and our ``Off-Policy DSKD'' on the setting ``Qwen2.5-Math-7B $\rightarrow$ Qwen2.5-1.5B'' and ``LLaMA2-7B $\rightarrow$ TinyLLaMA-1.1B''. The {\color{red}red} text means the wrong reasoning steps and final answer, while the {\color{midgray2}green} text means the correct reasoning steps and final answer.}
    \centering
    \resizebox{\linewidth}{!}{
    \begin{tabular}{l|l}
        \bottomrule
        \rowcolor{lightgray}
\textbf{Question-1} & \makecell[l]{
A 76-star flag has three rows of 8 stars, two rows of 6 stars and the rest are 5-star rows. How many rows of \\5 stars are there altogether on the flag?}  \\
\toprule
\bottomrule
\textbf{Methods} & \makecell[c]{\textbf{Responses \quad (Qwen2.5-Math-7B $\rightarrow$ Qwen2.5-1.5B)}} \\
\hline
Vanilla KD & \makecell[l]{
The total number of stars on the flag is 76.\\
\color{red}In the first row, there are 8 stars. In the second row, there are 6 stars. In the third row, there are 6 stars.\\\color{red}So far, we have accounted for 8 + 6 + 6 = 20 stars.\\
To find the number of rows of 5 stars, we subtract the number of stars accounted for from the total 
number\\ of stars: \color{red}76 - 20 = 56. \color{black}Since each row of 5 stars accounts for 5 stars, we divide the remaining\\
number of stars by 5: \color{red}56 / 5 = 11 remainder 1.\\
Therefore, \color{red}there are 11 rows of 5 stars on the flag. The answer is: 11} \\
\hline
Off-Policy DSKD & \makecell[l]{The total number of stars on the flag is 76.\\
\color{midgray2}The number of stars in the three rows of 8 stars is 3 x 8 = 24 stars. The number of stars in the two rows\\ \color{midgray2}of 6 stars is 2 x 6 = 12 stars. So, the total number of stars in these two types of rows is 24 + 12 = 36 stars.\\Therefore, the number of stars in the remaining rows is \color{midgray2}76 - 36 = 40 stars. \\Since each of these remaining rows has 5 stars, the number of rows is \color{midgray2}40 / 5 = 8 rows. The answer is: 8}\\   
        \toprule

\bottomrule
\rowcolor{lightgray}
\textbf{Question-2} & What is Linux?  \\
\toprule
\bottomrule
\textbf{Methods} & \makecell[c]{\textbf{Responses \quad (LLaMA2-7B $\rightarrow$ TinyLLaMA-1.1B)}} \\
\hline
Vanilla KD & \makecell[l]{Linux is a family of open source Unix-like operating systems. } \\
\hline
Off-Policy DSKD & \makecell[l]{
Linux is an open-source operating system based on the Linux kernel, which was first released in 1991 by\\ Linus Torvalds. It is typically packaged as Linux distributions, which include the kernel and supporting  sys-\\
tem software and libraries, many of which are provided by the GNU project.
}\\   
\toprule

    \end{tabular}
    }
    \label{tab:case_study}
\end{table*}

\subsection{Cross-Tokenizer KD for LLMs}
One of the main drawbacks of the white-box KD framework for current LLMs is the strict requirement that the teacher model and the student model must have the same tokenizer (vocabulary).
However, current LLMs from different model families usually have different vocabularies, which hinders the application of white-box KD methods.
To address this, \cite{fu23cotkd} first designs an exact match algorithm based on dynamic programming to solve the mismatch in tokenization and distribution from different LLMs.
Afterwards, \cite{wan24fusellm} refines this algorithm by replacing the exact match with the minimum edit distance match to align more tokens.
Additionally, \cite{boizard2024uld} proposes a universal logit distillation (ULD) loss from a closed-form solution of Wasserstein distance for cross-tokenizer distillation.
However, ULD ignores token alignment between differently-tokenized sequences and simply processes distributions from different LLMs by sorting and padding, which introduces noticeable noise into distillation.
To overcome the shortcomings of ULD, \cite{cui2024multilevelot} proposes multi-level optimal transport to achieve alignment on different levels for cross-tokenizer distillation.

Although effective, existing methods inevitably introduce noise or incomplete distributions from the teacher model. Differently, in this work, we decompose the cross-tokenizer KD problem into tokenization mismatch and distribution mismatch and separately address them: 1) the distribution mismatch is circumvented by the head sharing in the DSKD framework, which leverages the full information of the distributions; 2) the tokenization mismatch is solved by our exact token alignment algorithm to avoid the noisy supervision signals.

\section{Conclusion}
In this work, we first reveal two limitations in the current white-box KD framework for LLMs, \emph{i.e.}, low similarity between the student and the teacher and the requirement of the same vocabulary between two LLMs.
To address them, we propose a novel white-box KD framework, named dual-space knowledge distillation (DSKD), which unifies the output spaces of the student and the teacher for KD.
% Our framework is also compatible with various objectives for the current white-box KD framework.
On this basis, we further develop an exact token alignment algorithm to solve the vocabulary mismatch between different LLMs so that our DSKD framework can support KD between any two LLMs, regardless of their vocabulary.
Extensive experimental results on several benchmarks showcase that our DSKD significantly outperforms the vanilla white-box KD framework on various divergences and surpasses all existing cross-tokenizer KD methods.
Additionally, our on-policy DSKD exceeds existing on-policy KD methods.
All these results sufficiently demonstrate the effectiveness and the generalization of our DSKD framework.

\bibliographystyle{IEEEtran}
\bibliography{reference}

\onecolumn
\newpage
% \appendix
\appendices

% \begin{center}
%     \bfseries\Large Appendix
% \end{center}

% \newpage
\section{Difference Statement}\label{sec:appendix-difference}
This paper is an extended version of our prior work published at the conference EMNLP-2024 (\url{https://aclanthology.org/2024.emnlp-main.1010/}). The key differences include:
\begin{itemize}
    \item \textbf{Improvement of the primary DSKD method:} In the conference version, the two projectors for projecting teacher/student hidden states into the student/teacher space are initialized randomly, which requires substantial training steps till convergence. Therefore, we derive an optimal initialization (please refer to Equations (\ref{eq:init-stu1}), (\ref{eq:init-stu2}), and (\ref{eq:s2t_condition})) for the two projectors to achieve the logit invariance before/after projection. Empirically, we verify the effectiveness of this initialization method in the ablation study (Section \ref{sec:ablation-study}).
    \item \textbf{New token alignment algorithm for LLMs with different vocabularies:} In the conference version, we design a cross-model attention mechanism (CMA) to learn the alignment between tokens in two differently tokenized sequences. However, CMA further complicates the projectors and makes the convergence of the projectors slower. More importantly, we found that CMA does not directly support on-policy KD due to the reliance on the ground-truth target. To address these limitations, we propose the exact token alignment algorithm as an alternative to align the same tokens in two sequences, which is more easily to be incorporated into the DSKD framework (Section \ref{sec:method_cross_model}).
    \item \textbf{On-Policy training strategy:} We extend our DSKD framework from the original off-policy KD scenarios to the on-policy KD scenarios (Section \ref{sec:on-policy}). From our experiments, on-policy DSKD(-ETA) performs significantly better than the off-policy counterparts and outperforms existing on-policy KD methods. (Please refer to Tables \ref{tab:main_results_on_policy_same_vocab} and \ref{tab:main_results_dif_vocab}). 
    \item \textbf{More evaluation on three typical capacities:} To further prove the superiority and generalization of our DSKD and DSKD-ETA, we evaluate our method on three typical capacities, including general instruction following (AlpacaEval and Evol-Instruct), mathematical reasoning (MATH-500 and GSM), and code generation (HumanEval and MBPP). The results (reported in Tables \ref{tab:instruction_evaluation}, \ref{tab:math_evaluation}, and \ref{tab:code_evaluation}) show that our method achieves better performance than the baseline method.
\end{itemize}

\section{Pseudo Code for Simulation Experiments}
\label{sec:pseudo_code}

We provide the pseudo code for re-implementing the key parts of our simulation experiments in Algorithm \ref{alg:alg2}.
\begin{algorithm}[ht]
\caption{The Pseudo-code of the Simulation Experiment.}\label{alg:alg2}
\begin{algorithmic}[1]
\STATE \textbf{Class} Teacher(nn.Module):
\STATE \quad \textbf{def} \_\_init\_\_(self):
\STATE \quad \quad \quad \textbf{super}(Teacher, self).\_\_init\_\_()
% \STATE \quad \quad \quad \textit{\# Initialize teacher hiddens from Gaussian Distribution} $\mathcal{N}(0, 2)$
\STATE \quad \quad \quad self.hidden = torch.randn(100, 2) * 2 \quad \textit{\# Initialize teacher hiddens from Gaussian Distribution} $\mathcal{N}(0, 2)$.
% \STATE \quad \quad \textbf{\# Head with 10000 classes}
\STATE \quad \quad \quad self.head = torch.randn(10000, 2) \qquad \textit{\# The head contains 10000 classes.}
\STATE
\STATE \textbf{Class} Student(nn.Module):
\STATE \quad \textbf{def} \_\_init\_\_(self):
\STATE \quad \quad \quad \textbf{super}(Student, self).\_\_init\_\_()
\STATE \quad \quad \quad \textit{\# Initialize student hiddens from Gaussian Distribution} $\mathcal{N}(3, 1)$.
\STATE \quad \quad \quad self.hidden = nn.Parameter(torch.randn(100, 2) + 3) 
% \STATE \quad \quad \textbf{\# Head with 10000 classes}
\STATE \quad \quad \quad self.head = nn.Parameter(torch.randn(10000, 2))  \qquad \textit{\# The head contains 10000 classes.}
\STATE
\STATE \textbf{def} kd\_with\_different\_head(student, teacher):
\STATE \quad \quad \textit{\# Calculating logits with respective heads.}

\STATE \quad \quad stu\_logits = \textbf{student}.hidden.matmul(\textbf{student}.head.transpose(-1, -2))
\STATE \quad \quad tea\_logits = \textbf{teacher}.hidden.matmul(\textbf{teacher}.head.transpose(-1, -2))
\STATE \quad \quad kd\_loss = distance\_func(stu\_logits, tea\_logits)
\STATE \quad \quad \textbf{return} kd\_loss
\STATE
\STATE \textbf{def} kd\_with\_shared\_head(student, teacher):
\STATE \quad \quad \textit{\# Calculating logits with the same head (student's head).}

\STATE \quad \quad stu\_logits = \textbf{student}.hidden.matmul(\textbf{student}.head.transpose(-1, -2))
\STATE \quad \quad tea\_logits = \textbf{teacher}.hidden.matmul(\textbf{student}.head.transpose(-1, -2))
\STATE \quad \quad kd\_loss = distance\_func(stu\_logits, tea\_logits)
\STATE \quad \quad \textbf{return} kd\_loss
\end{algorithmic}
\end{algorithm}

% \begin{minted}{c}
%     stu_logits = self.h1.matmul(self.e1.transpose(-1, -2))
%     tea_logits = self.h3.matmul(self.e3.transpose(-1, -2))
% \end{minted}

% \lstset{
%     numbers=none
% }
% \begin{lstlisting}
% class Teacher(nn.Module):
%     def __init__(self):
%         super(Teacher, self).__init__()
%         # the initial teacher hiddens are sampled from Gaussian Distribution N(0, 2)
%         self.hidden = torch.randn(100, 2) * 2
%         # the head contains 10000 classes
%         self.head = torch.randn(10000, 2)  

% class Student(nn.Module):
%     def __init__(self):
%         super(Student, self).__init__()
%         # the initial student hiddens are sampled from Gaussian Distribution N(3, 1)
%         self.hidden = nn.Parameter(torch.randn(100, 2) + 3)
%         # the head contains 10000 classes
%         self.head = nn.Parameter(torch.randn(10000, 2))  

% def kd_with_different_head(student, teacher):
%     student_logits = student.hidden.matmul(student.head.transpose(-1, -2))
%     # calculating logits with the respective heads
%     teacher_logits = teacher.hidden.matmul(teacher.head.transpose(-1, -2))  
%     kd_loss = distance_func(student_logits, teacher_logits)
%     return kd_loss

% def kd_with_shared_head(student, teacher):
%     student_logits = student.hidden.matmul(student.head.transpose(-1, -2))
%     # calculating logits with the same head (student's head)
%     teacher_logits = teacher.hidden.matmul(student.head.transpose(-1, -2))  
%     kd_loss = distance_func(student_logits, teacher_logits)
%     return kd_loss 
% \end{lstlisting}
As shown in the code, we manually separate the hidden states of the student and teacher in initialization, so that the difference before and after KD will be clearer.
Besides, to unify the output spaces of the two models, we share the prediction head of the student with the teacher in ``kd\_with\_shared\_head''.
In this way, the output distributions of the student being optimized are as same as the ones in ``kd\_with\_different\_head'' and thus the results will be more comparable with the ones in ``kd\_with\_different\_head''.
The student models are optimized by the SGD optimizer with appropriate learning rates in $[1.0, 40.0]$ for different distance functions.

\section{Simulation Results for Other Distance Functions}
\label{sec:other_simulation}
\begin{figure*}[t]
	\centering
	\subfigure[SKL: Before KD]{
		\begin{minipage}[t]{0.24\linewidth}
			\centering
			\includegraphics[width=\linewidth]{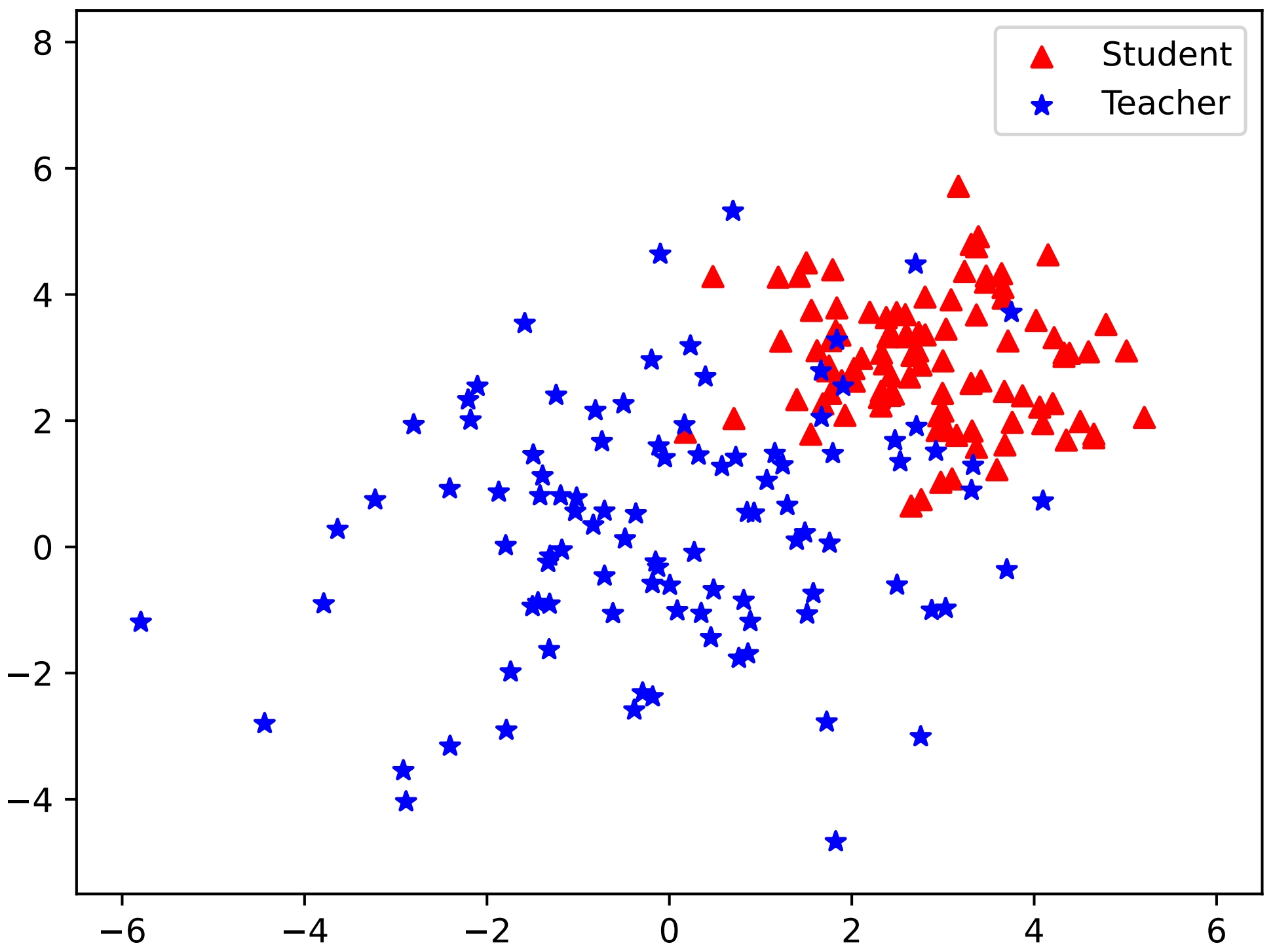}
		\end{minipage}
	}%
	\subfigure[SKL: Different heads]{
		\begin{minipage}[t]{0.24\linewidth}
			\centering
			\includegraphics[width=\linewidth]{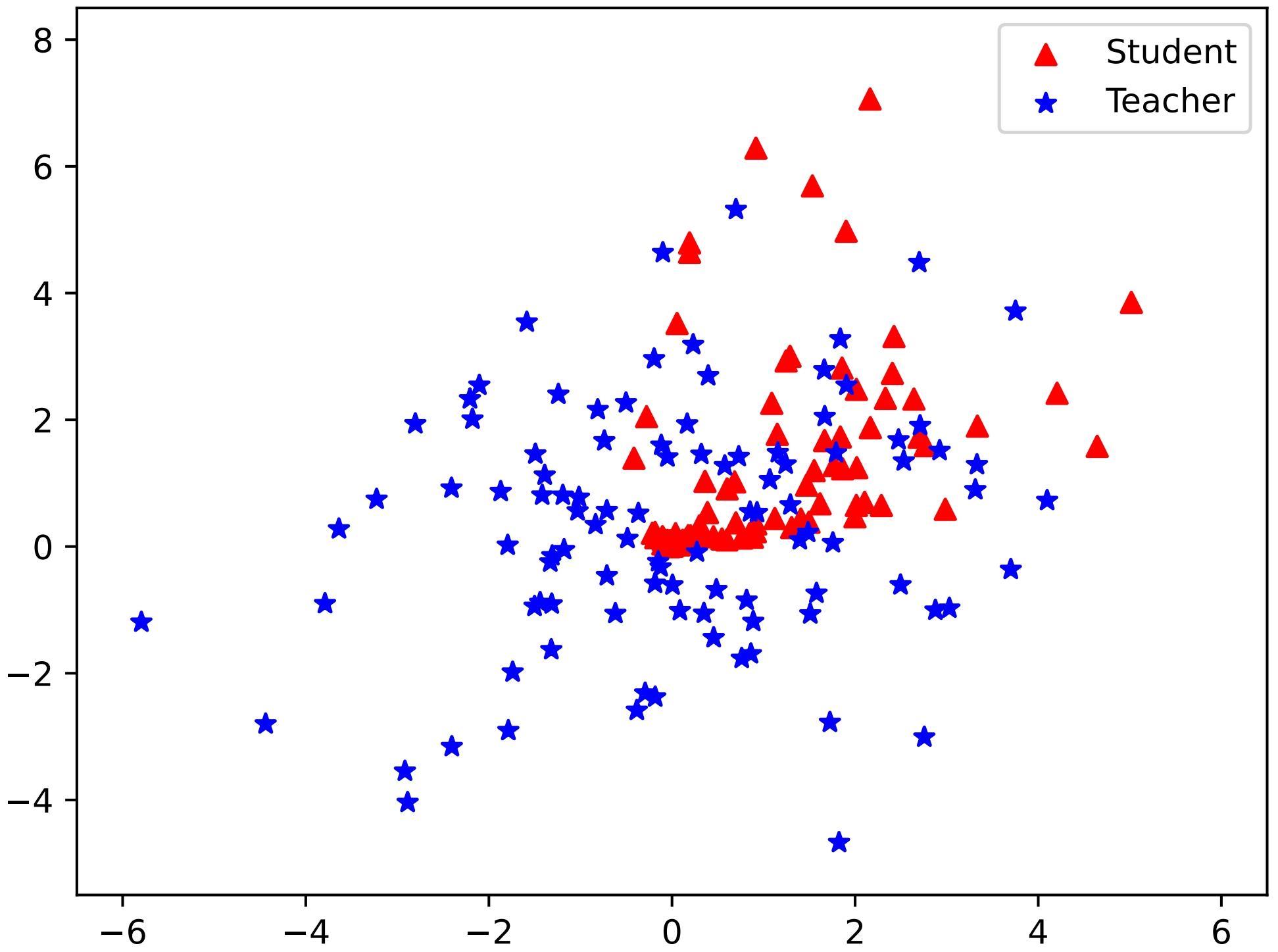}
		\end{minipage}
	}% 
	\subfigure[SKL: Shared head]{
		\begin{minipage}[t]{0.24\linewidth}
			\centering
			\includegraphics[width=\linewidth]{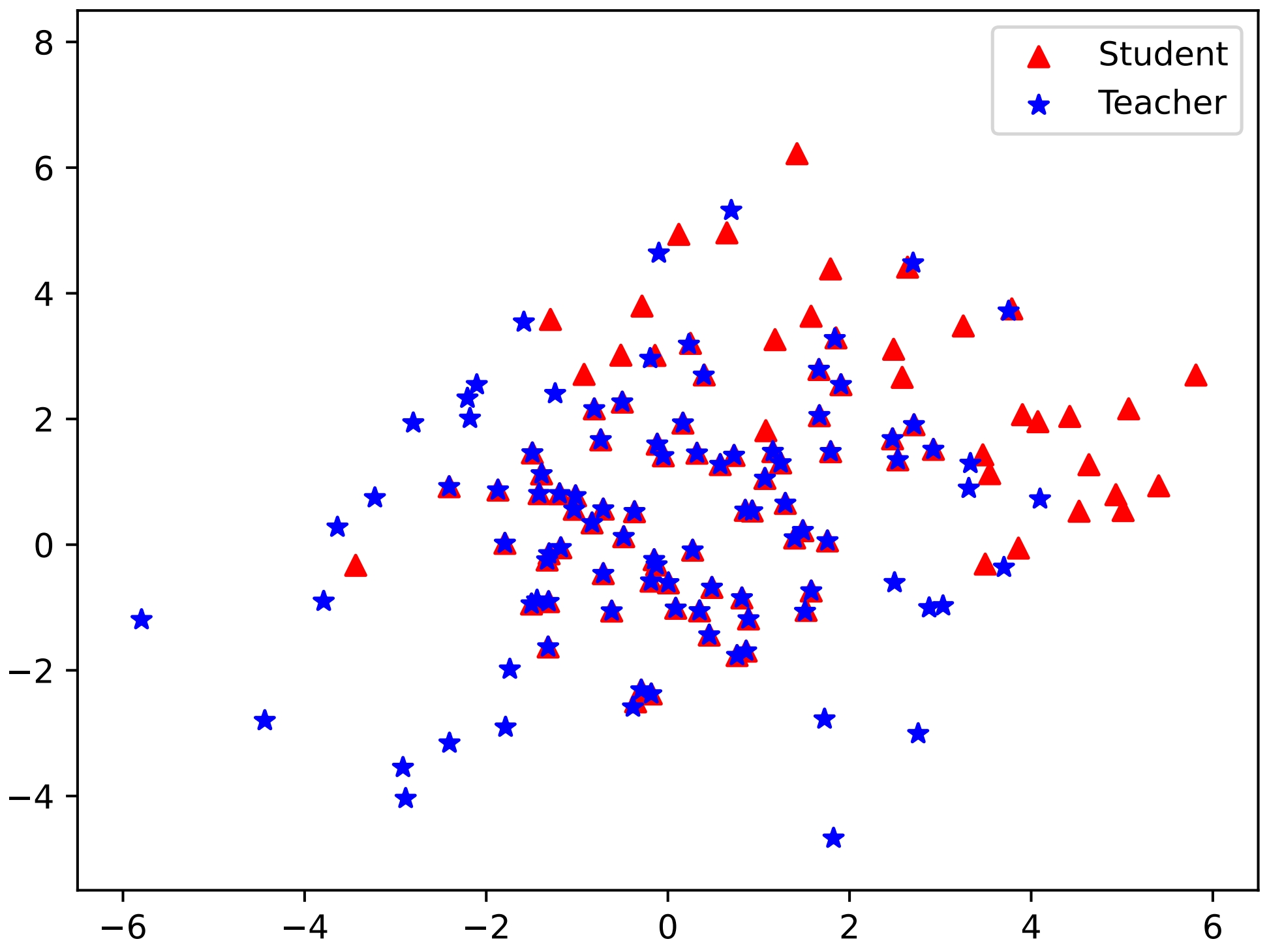}
		\end{minipage}
	}%
	\subfigure[SKL: Loss curves of KD]{
		\begin{minipage}[t]{0.24\linewidth}
			\centering
			\includegraphics[width=\linewidth]{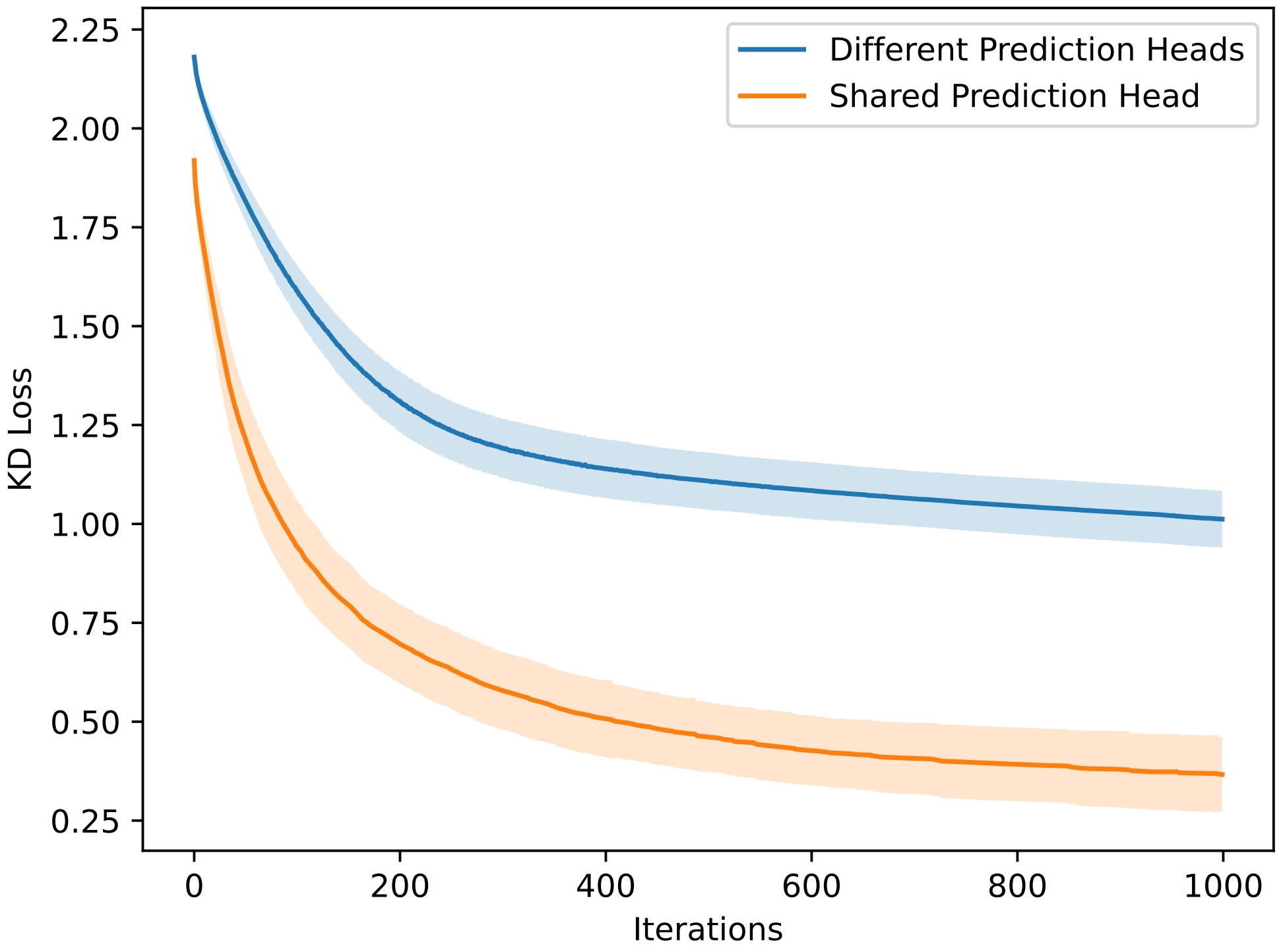}
		\end{minipage}
	}%

        \subfigure[SRKL: Before KD]{
		\begin{minipage}[t]{0.24\linewidth}
			\centering
			\includegraphics[width=\linewidth]{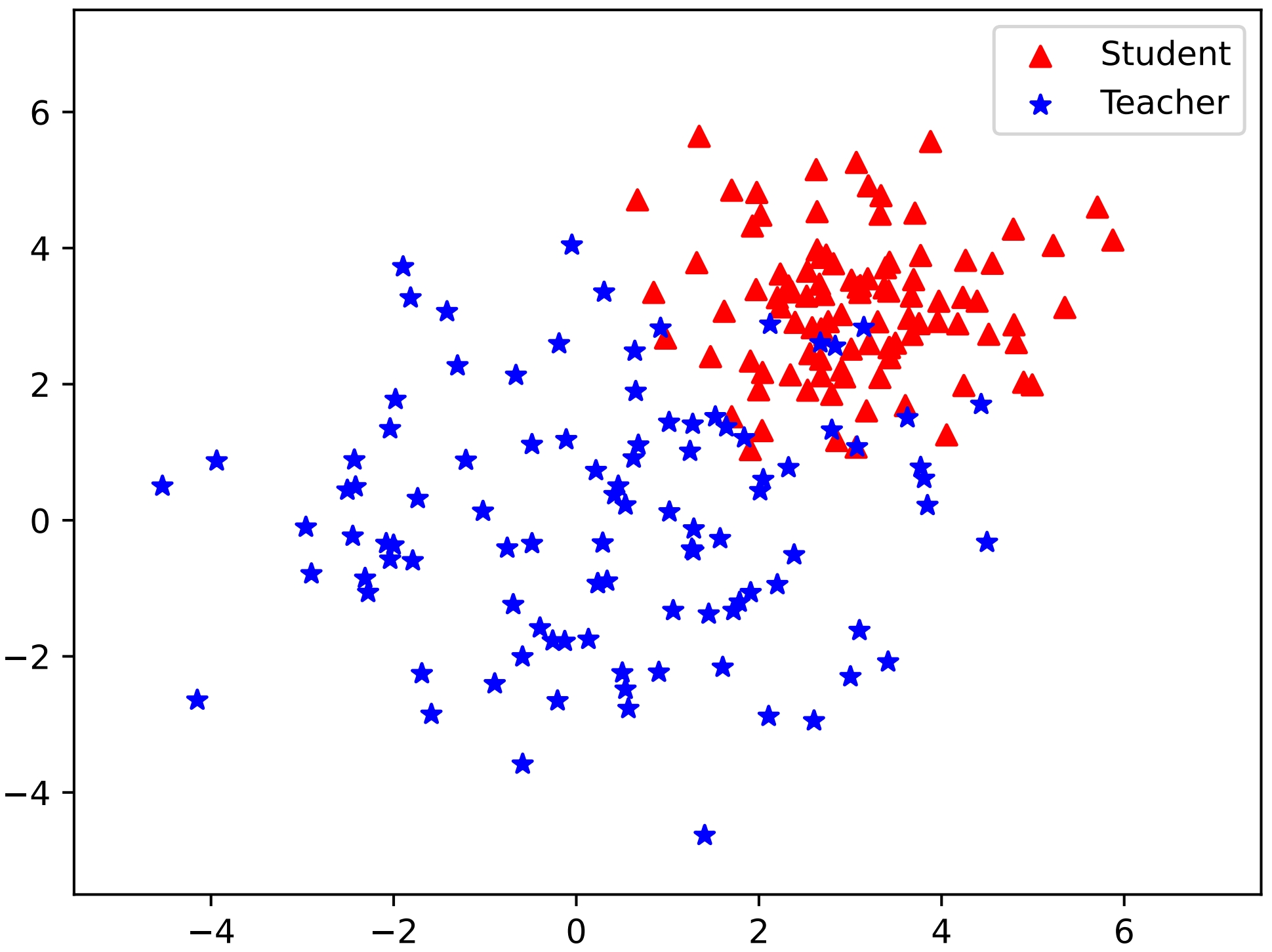}
		\end{minipage}
	}%
	\subfigure[SRKL: Different heads]{
		\begin{minipage}[t]{0.24\linewidth}
			\centering
			\includegraphics[width=\linewidth]{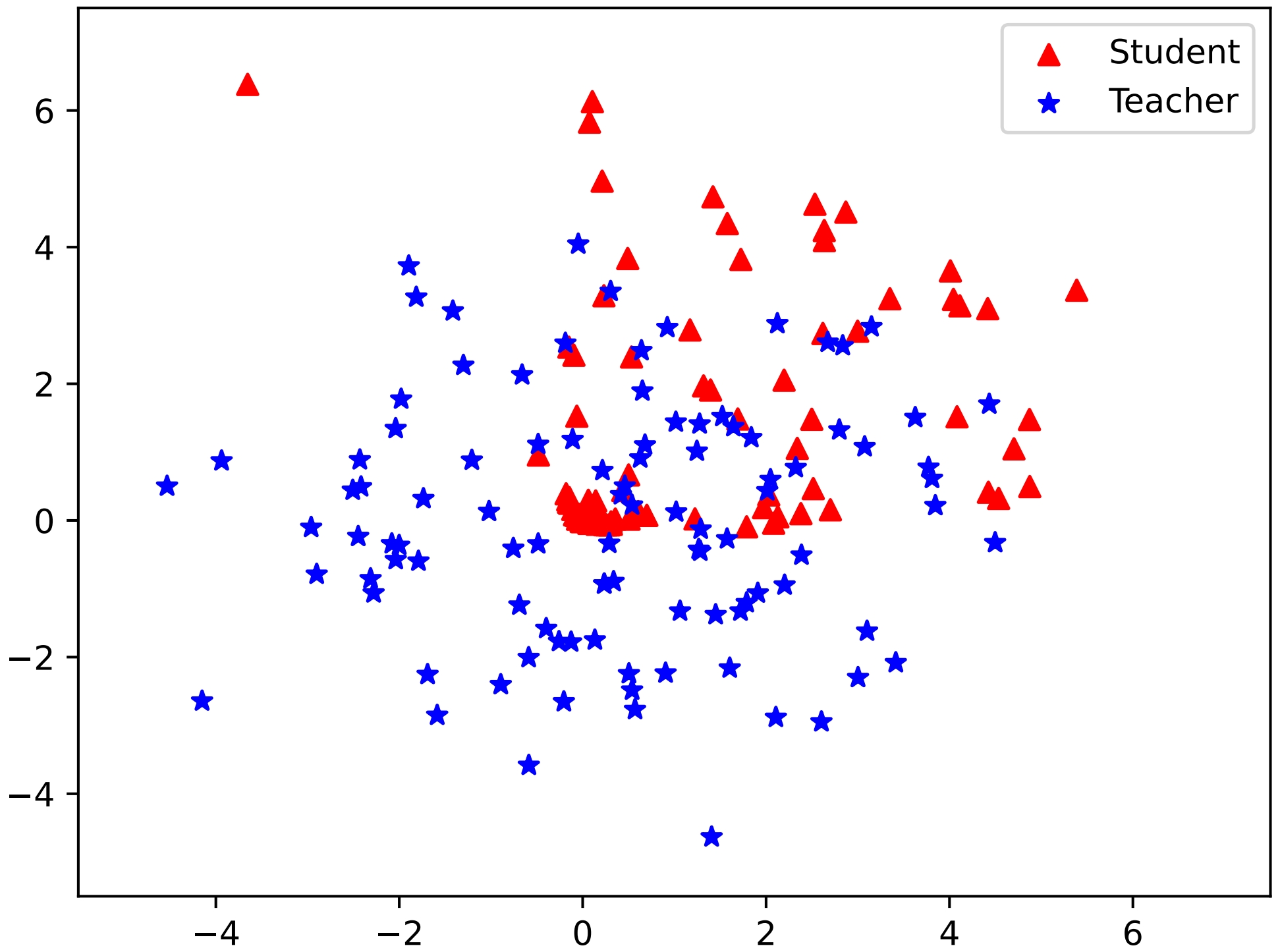}
		\end{minipage}
	}% 
	\subfigure[SRKL: Shared head]{
		\begin{minipage}[t]{0.24\linewidth}
			\centering
			\includegraphics[width=\linewidth]{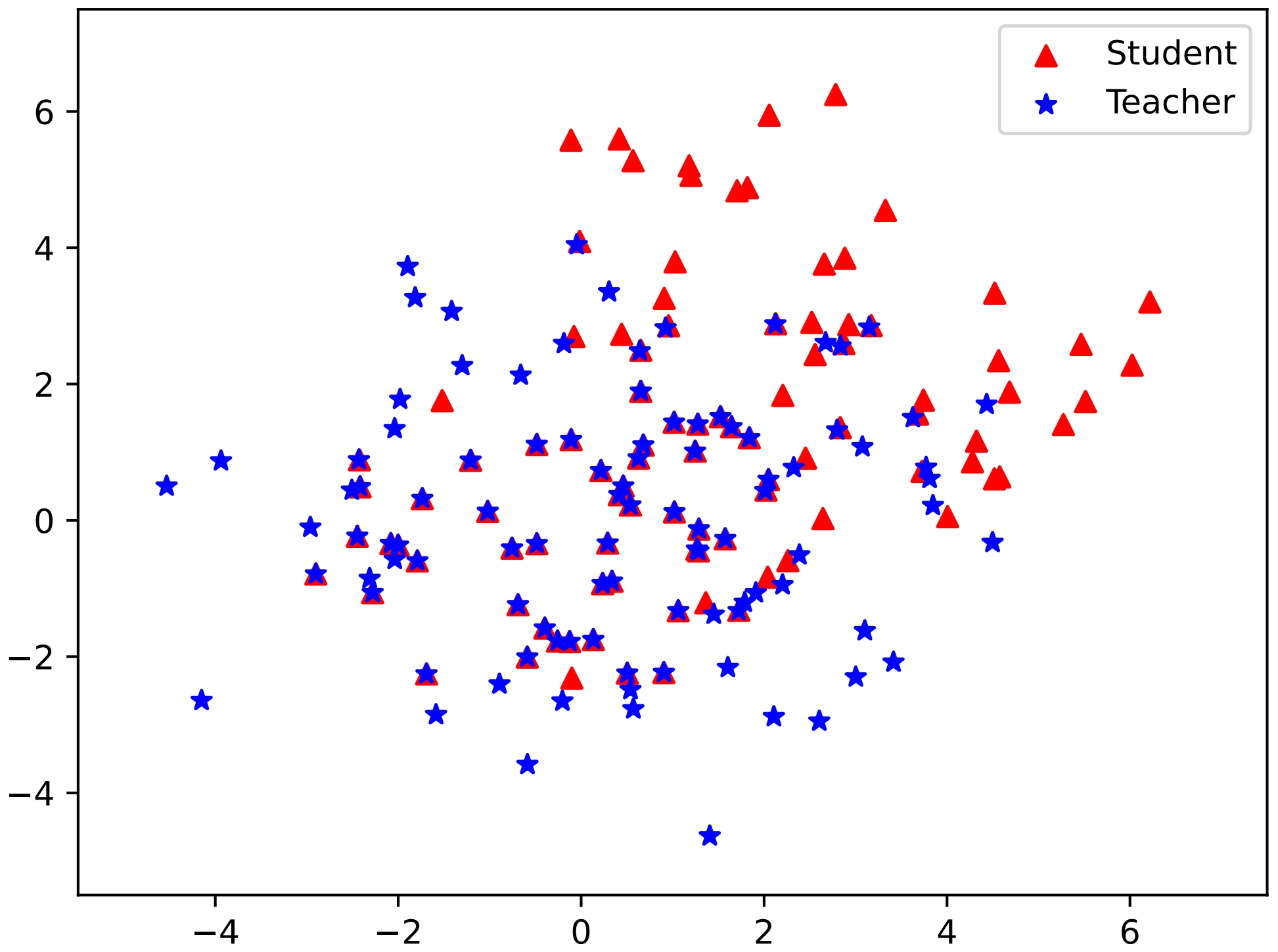}
		\end{minipage}
	}%
	\subfigure[SRKL: Loss curves of KD]{
		\begin{minipage}[t]{0.24\linewidth}
			\centering
			\includegraphics[width=\linewidth]{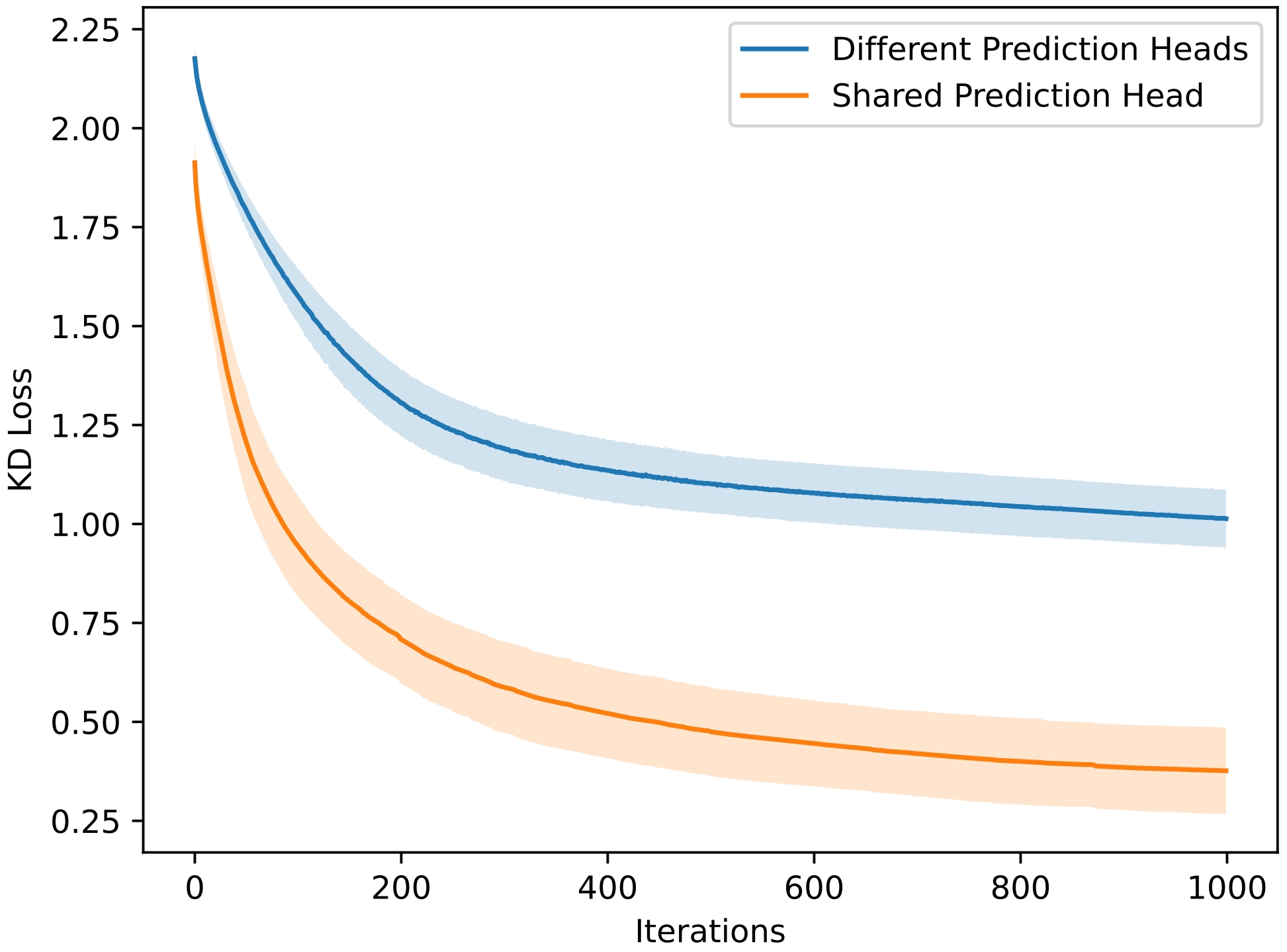}
		\end{minipage}
	}%

        \subfigure[AKL: Before KD]{
		\begin{minipage}[t]{0.24\linewidth}
			\centering
			\includegraphics[width=\linewidth]{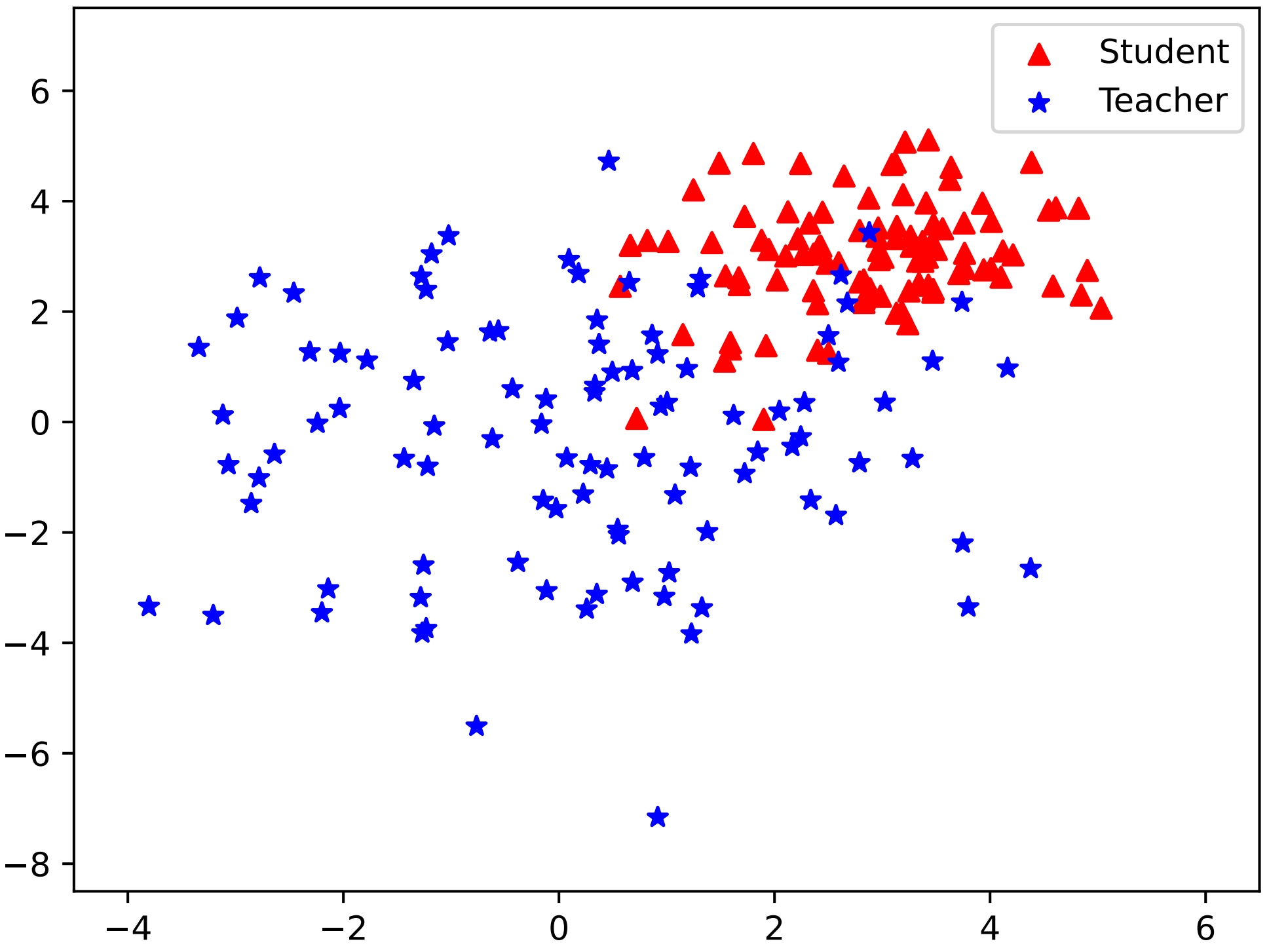}
		\end{minipage}
	}%
	\subfigure[AKL: Different heads]{
		\begin{minipage}[t]{0.24\linewidth}
			\centering
			\includegraphics[width=\linewidth]{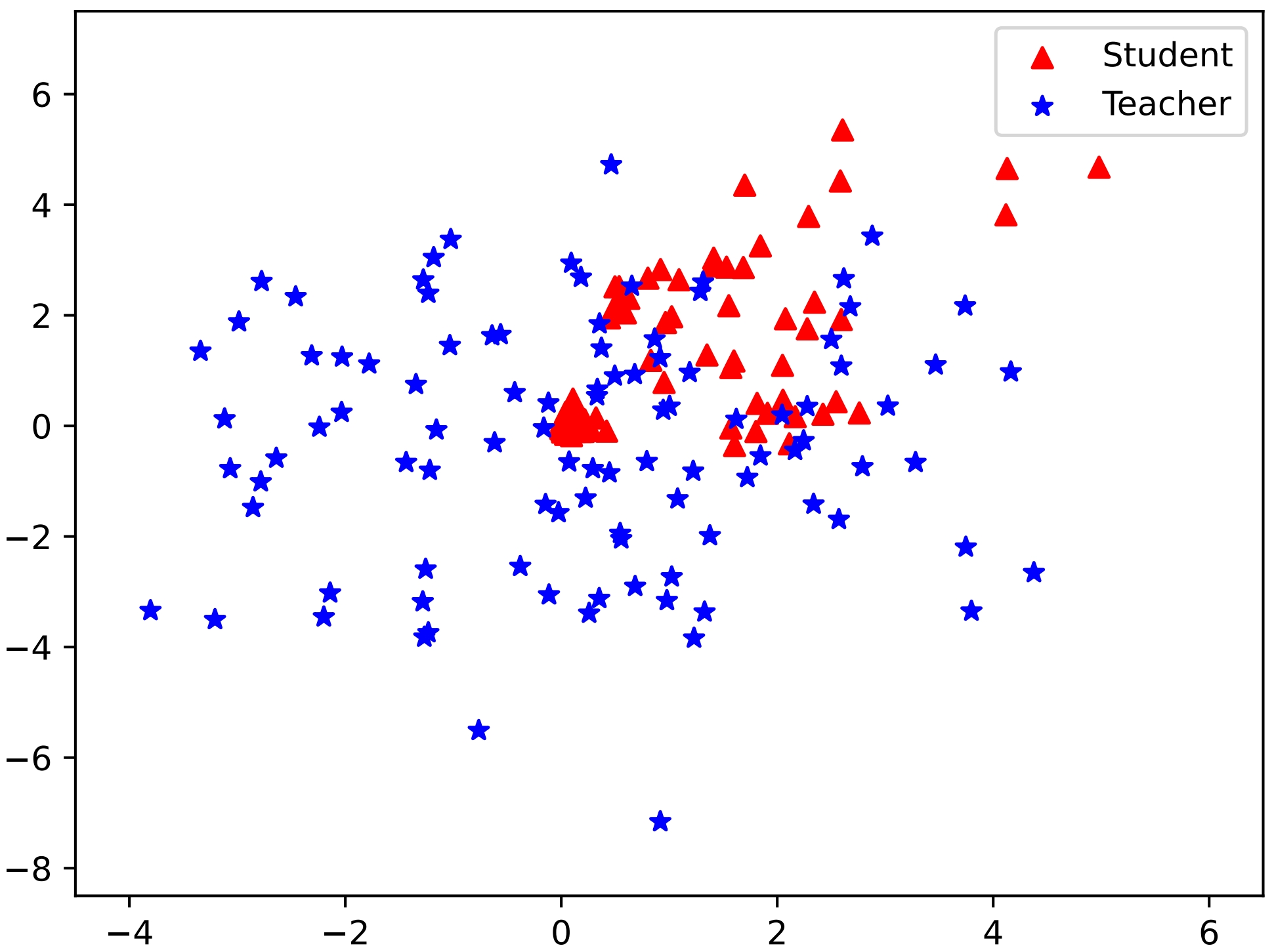}
		\end{minipage}
	}% 
	\subfigure[AKL: Shared head]{
		\begin{minipage}[t]{0.24\linewidth}
			\centering
			\includegraphics[width=\linewidth]{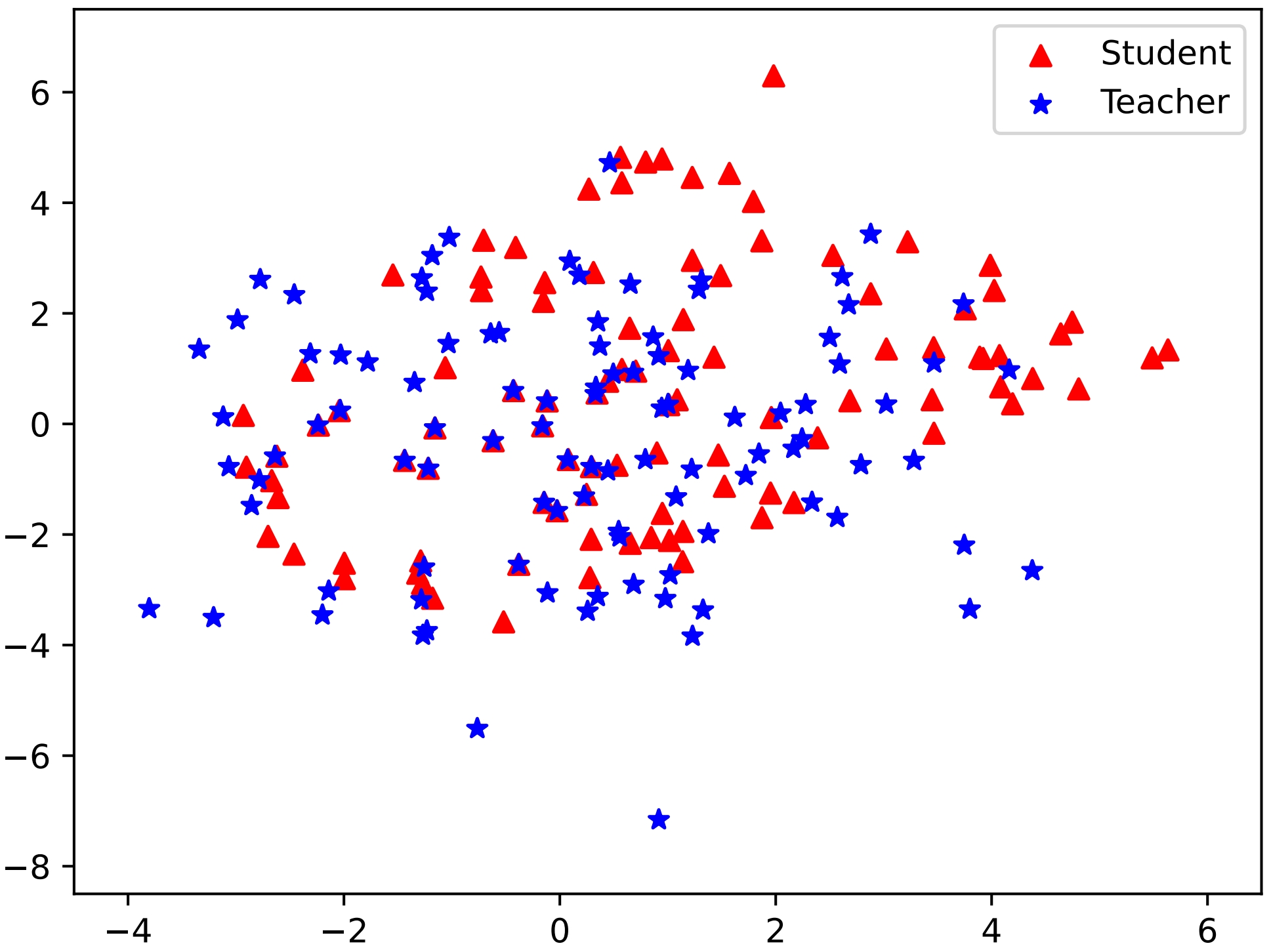}
		\end{minipage}
	}%
	\subfigure[AKL: Loss curves of KD]{
		\begin{minipage}[t]{0.24\linewidth}
			\centering
			\includegraphics[width=\linewidth]{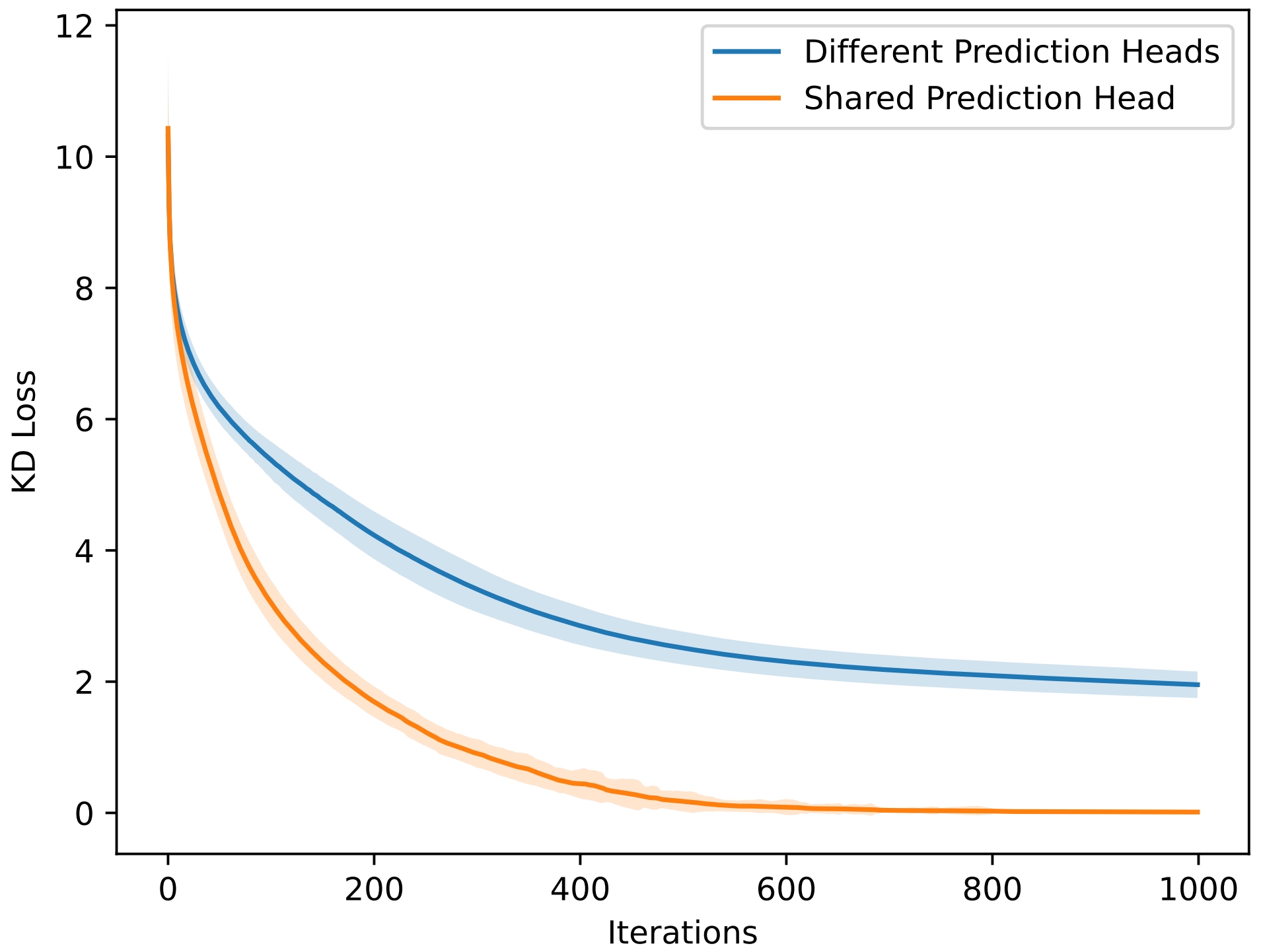}
		\end{minipage}
	}%
    
	\centering
	\caption{Simulation results with SKL/SRKL/AKL divergence as the distance function $\mathcal{D}(\cdot||\cdot)$. (a), (b), (c), (e), (f), (g), (i), (j), and (k) plot the {\color{red} student's hidden states} and the {\color{blue} teacher's hidden states} before and after the two KD processes. ``Different heads'' means using the teacher and student head respectively during KD, while ``Shared head'' means only using the student head as the shared head to obtain the distributions during KD. (d), (h), and (l) show the convergence curves of $\mathcal{L}_{kd}$ in the two KD processes.}
	\label{fig:srakl_simulation}
\end{figure*}

We complement the remaining results of simulation experiments for the following objectives: skewed KL divergence, skewed RKL divergence, and adaptive KL divergence.
The results are plotted in Figure \ref{fig:srakl_simulation}.
It is shown that no matter which distance function is used, the student after KD will have low representation similarity with the teacher and leave a large margin to the minimum distance between the two distributions when using different prediction heads.
Thus, all these results lead to the consistent conclusion in Section \ref{sec:low_sim}, and also suggest that the current KD framework may have inherent flaws on enhancing the similarity between the student model and the teacher model.
As a solution, unifying the output spaces by sharing the prediction head for teacher and student may achieve a more effective KD process. 

\section{Effect of Temperature for KD} \label{sec:temperature}
\begin{figure*}[ht]
    \centering
    \includegraphics[width=0.4\linewidth]{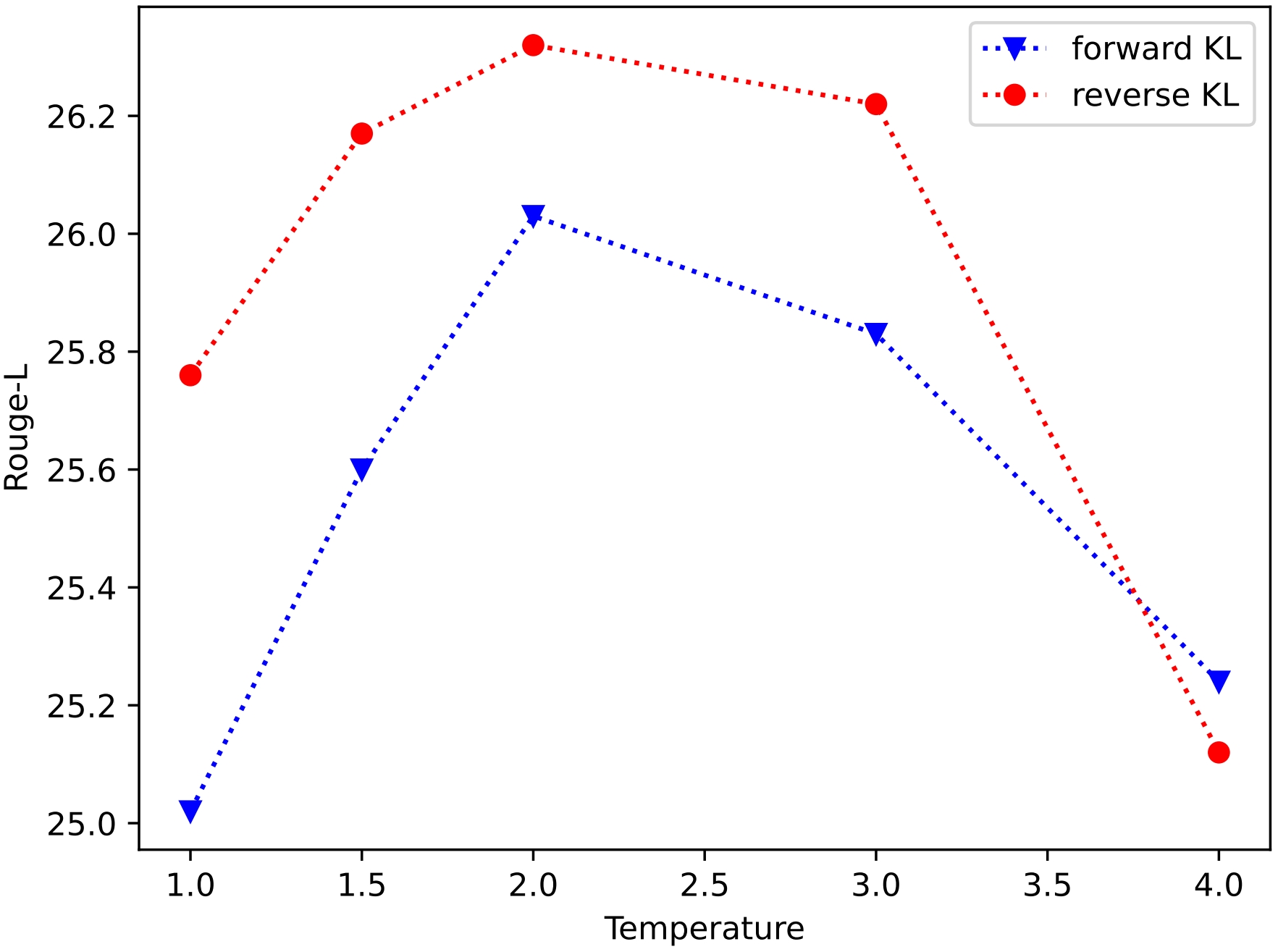}
    \caption{Rouge-L scores (\%) on the validation set for different temperature coefficients in KL divergence and reverse KL divergence.}
    \label{fig:temperature}
\end{figure*}

As an important hyper-parameter in KD, the temperature coefficient $\tau$ significantly affects the final performance of KD.
As stated by the previous literature, a larger temperature ($>$ 1.0) will smooth the teacher's distribution and transfer more class relationship information to the student model. 
Thus, we search for the best temperatures among [1.0, 1.5, 2.0, 3.0, 4.0] for two representative objectives (\emph{i.e.}, KL divergence and reverse KL divergence) on the validation set and report the results in Figure \ref{fig:temperature}.
The results show that both objectives perform best when the temperature is 2.0.
Thus, we keep the temperature to 2.0 for all objectives in our experiments.

\begin{table}[t]
\caption{Detailed training configurations of DSKD and DSKD-ETA for Qwen2.5-1.5B and Llama-3.2-1B.}
    \centering
    \resizebox{0.9\linewidth}{!}{
        \begin{tabular}{c|cc|cc|cc}
            \bottomrule
            \multirow{2}{*}{\textbf{Settings}} & \multicolumn{2}{c|}{\textbf{Instruct Following}} & \multicolumn{2}{c}{\textbf{Mathematical Reasoning}} & \multicolumn{2}{c}{\textbf{Code Generation}}\\
            \cline{2-7}
            & Qwen2.5-1.5B & Llama-3.2-1B & Qwen2.5-1.5B & Llama-3.2-1B & Qwen2.5-1.5B & Llama-3.2-1B \\ 
            \hline
            Epoch & 3 & 3 & 3 & 3 & 1 & 1 \\
            Learning Rate & 1e-5 & 1e-5 & 1e-5 & 1e-5 & 1e-5 & 1e-5 \\
            Projector Learning Rate & 1e-3 & 1e-3 & 1e-3 & 1e-3 & 1e-3  & 1e-3 \\
            Batch Size & 128 & 128 & 128 & 128 & 128  & 32\\
            LR Scheduler & Cosine & Cosine & Cosine & Cosine & Cosine & Cosine \\
            Fine-Tuning Method & Full & Full & Full & Full & Full & Full \\
            KD Rate & 0.5 & 0.5 & 0.5 & 0.5 & 0.5 & 0.5 \\
            KD Temperature & 1.0 & 1.0 & 1.0 & 1.0 & 1.0 & 1.0 \\
            % TOP-K Vocab & 20000 & All & 20000 & 20000 & 20000 & All \\
            \toprule
        \end{tabular}
    }
    
    \label{tab:train_config_other}
\end{table}

\section{Details of Evaluation on Other Benchmarks}\label{sec:appendix-other-details}
To further prove the superiority and generalization of our DSKD and DSKD-ETA, we evaluate our method on three typical capacities, including general instruction following, mathematical reasoning, and code generation.
The detailed introduction of the used dataset and benchmarks is as follows:

\begin{itemize}
    \item \textbf{UltraChat200k}\footnote{\url{https://huggingface.co/datasets/HuggingFaceH4/ultrachat_200k}} \cite{ding-etal-2023-enhancing}: This is a heavily filtered version of the UltraChat\footnote{\url{https://github.com/thunlp/UltraChat}} dataset, which consists of 1.4M dialogues generated by ChatGPT and spanning a wide range of topics. We randomly sample 50K as the train set.
    \item \textbf{AlpacaEval}\footnote{\url{https://github.com/tatsu-lab/alpaca_eval}} \cite{alpaca_eval}: AlpacaEval contains 805 challenging questions. And we use the \texttt{text-davinci-003} as the baseline.
    \item \textbf{Evol-Instruct}\footnote{\url{https://github.com/nlpxucan/WizardLM/blob/main/WizardLM/data/WizardLM_testset.jsonl}} \cite{xu2024wizardlm}: This benchmark contains 218 questions, spanning multiple topics generated using the Evol-Instruct procedure. It covers a diverse list of user-oriented instructions including difficult Coding Generation, Debugging, Math, Reasoning, Complex Formats, Academic Writing, Extensive Disciplines, and so on.
    
    \item \textbf{MetaMathQA}\footnote{\url{https://huggingface.co/datasets/meta-math/MetaMathQA}} \cite{yu2024metamathbootstrapmathematicalquestions}: All data of MetaMathQA are augmented from the training sets of GSM8K \cite{cobbe2021gsm8k} and MATH \cite{lightman2023lets}. (None of the augmented data is from the testing set.) We randomly sample 50K examples as the train set.
    \item \textbf{MATH-500}\footnote{\url{https://huggingface.co/datasets/HuggingFaceH4/MATH-500}} \cite{lightman2023lets}:  This benchmark contains a subset of 500 problems from the MATH benchmark.
    \item \textbf{GSM}\footnote{\url{https://huggingface.co/datasets/openai/gsm8k}} \cite{cobbe2021gsm8k}: This benchmark is the test set of GSM8K, including 1319 examples.
    GSM8K (Grade School Math 8K) is a dataset of 8.5K high-quality quality linguistically diverse grade school math word problems. The dataset was created to support the task of question answering on basic mathematical problems that require multi-step reasoning.

    \item \textbf{Magicoder}\footnote{\url{https://huggingface.co/datasets/mesolitica/mixtral-magicoder}} \cite{wei2024magicoderempoweringcodegeneration}: This dataset contains 10925 Python examples, including tasks and code functions. We randomly sample 10K examples as the training set.
    \item \textbf{HumanEval}\footnote{\url{https://huggingface.co/datasets/openai/openai_humaneval}}\cite{chen2021codex}: The HumanEval dataset released by OpenAI includes 164 programming problems with a function signature, docstring, body, and several unit tests. The programming problems are written in Python and contain English natural text in comments and docstrings.
    \item \textbf{MBPP}\footnote{\url{https://huggingface.co/datasets/google-research-datasets/mbpp}} \cite{austin2021programsynthesislargelanguage}: This benchmark consists of around 1000 crowd-sourced Python programming problems, designed to be solvable by entry-level programmers, covering programming fundamentals, standard library functionality, and so on. Each problem consists of a task description, code solution, and 3 automated test cases. We use the hand-verified subset of the data with 399 examples and evaluate the performance based on the Qwen2.5-coder project\footnote{\url{https://github.com/QwenLM/Qwen2.5-coder}}.
    
\end{itemize}

Additionally, the detailed training configurations are listed in Table \ref{tab:train_config_other}. 
% Since Qwen and Llama3 have a relatively larger vocabulary and may restrict the precision of projector initialization, we only choose the head for the Top-K tokens in the vocabulary for some tasks.

\end{document}